\newif\ifarxiv
\arxivtrue % compiles arxiv version 
\ifarxiv
    \documentclass[]{fairmeta} % Option "twocolumn" available, but please prioritize single-column
    \usepackage{algorithm}
    \usepackage{lmodern} % K: moved it here from preamble.tex
    \usepackage{amsmath}
    \usepackage{amssymb}
    \usepackage{subcaption}
    \usepackage{adjustbox}

\else
    \documentclass{article}

    \usepackage{neurips_2025}

    \usepackage[utf8]{inputenc} % allow utf-8 input
    \usepackage[T1]{fontenc}    % use 8-bit T1 fonts
    \usepackage{hyperref}       % hyperlinks
    \usepackage{url}            % simple URL typesetting
    \usepackage{booktabs}       % professional-quality tables
    \usepackage{amsfonts}       % blackboard math symbols
    \usepackage{nicefrac}       % compact symbols for 1/2, etc.
    \usepackage{microtype}      % microtypography
    \usepackage[dvipsnames]{xcolor}
    
    \hypersetup{
        colorlinks,
        linkcolor={red!80!black},
        citecolor={blue!80!black},
        urlcolor={blue!80!black}
    }
    \usepackage{adjustbox}
    \usepackage{graphicx}
    \usepackage{subfigure}
    \usepackage{nicematrix} 
    \usepackage{tikz} 
    \usepackage{sidecap} 
    \usepackage{arydshln}
    \usepackage{makecell}
    \usepackage{multirow}
    \usepackage{amsmath}
    \usepackage{amssymb}
    \usepackage{mathtools}
    \usepackage{amsthm}
    \usepackage[capitalize,noabbrev]{cleveref}

\fi

\usepackage{amsmath,amsfonts,bm}
\usepackage{commath}
\usepackage{physics}
\usepackage{xcolor}

\newcolumntype{H}{>{\setbox0=\hbox\bgroup}c<{\egroup}@{}}
\def\mypar#1{\vspace{2mm}\noindent\textbf{#1}\hspace{1mm}}

\newcommand{\m}[1]{\vec{\mu}_#1}
\makeatletter
\DeclareRobustCommand{\pdot}{\mathbin{\mathpalette\pdot@\relax}}
\newcommand{\pdot@}[2]{%
  \ooalign{%
    $\m@th#1\circ$\cr
    \hidewidth$\m@th#1\cdot$\hidewidth\cr
  }%
}

\makeatother
\def\eqref#1{equation~\ref{#1}}
\def\1{\bm{1}}

\DeclareMathAlphabet{\mathsfit}{\encodingdefault}{\sfdefault}{m}{sl}
\SetMathAlphabet{\mathsfit}{bold}{\encodingdefault}{\sfdefault}{bx}{n}

\DeclareMathOperator{\sign}{sign}

\usepackage{hyperref}
\usepackage{url}
\usepackage{lipsum}
\usepackage{graphicx}
\usepackage{wrapfig}

\usepackage{silence}
\usepackage{cleveref}
\crefname{section}{Sec.}{Secs.}
\Crefname{section}{Section}{Sections}
\crefname{table}{Tab.}{Tabs.}
\Crefname{table}{Table}{Tables}
\crefname{figure}{Fig.}{Figs.}
\Crefname{figure}{Figure}{Figures}
\crefname{appendix}{App.}{App.}
\crefname{equation}{Eq.}{Eqs.}

\usepackage{xspace}
\makeatletter %Had to add for ICML compatibility
\DeclareRobustCommand\onedot{\futurelet\@let@token\@onedot}
\def\@onedot{\ifx\@let@token.\else.\null\fi\xspace}
\makeatother %Had to add for ICML compatibility

\def\eg{\emph{e.g}\onedot} 

\def\ie{\emph{i.e}\onedot}

\def\wrt{w.r.t\onedot}

\title{Classifier-Free Guidance:  
From High-Dimensional Analysis to Generalized Guidance Forms}

\ifarxiv
    \author[1,2]{Krunoslav Lehman Pavasovic}
    \author[1]{Jakob Verbeek}
    \author[2,\dagger]{Giulio Biroli}
    \author[3,\dagger]{Marc Mezard}
    
    \affiliation[1]{FAIR at Meta}
    \affiliation[2]{École Normale Supérieure, Paris}
    \affiliation[3]{Bocconi University, Milan}

    \contribution[\dagger]{Joint last author}
    
    \abstract{
    \begin{abstract}
Classifier-Free Guidance (CFG) is a widely adopted technique in diffusion and flow-based generative models, enabling high-quality conditional generation. A key theoretical challenge is characterizing the distribution induced by CFG, particularly in high-dimensional settings relevant to real-world data. Previous works have shown that CFG modifies the target distribution, steering it towards 
a distribution sharper than the target one, more shifted towards the boundary of the class. In this work, we provide a high-dimensional analysis of CFG, showing that these distortions vanish as the data dimension grows.
We present a ``blessing-of-dimensionality'' result demonstrating that in sufficiently high and infinite dimensions, CFG accurately reproduces the target distribution.
Using our high-dimensional theory, we show that there is a large family of guidances enjoying this property, in particular non-linear CFG generalizations. We study a simple non-linear ``power-law'' version, for which we demonstrate improved robustness, sample fidelity and diversity. 
Our findings are validated with experiments on class-conditional and text-to-image generation using state-of-the-art diffusion and flow-matching models.

\end{abstract}
    }
    
    \date{\today}
    \correspondence{Krunoslav Lehman Pavasovic at \email{krunolp@meta.com}}
    
\else
    \author{%
      David S.~Hippocampus\thanks{Use footnote for providing further information
        about author (webpage, alternative address)---\emph{not} for acknowledging
        funding agencies.} \\
      Department of Computer Science\\
      Cranberry-Lemon University\\
      Pittsburgh, PA 15213 \\
      \texttt{hippo@cs.cranberry-lemon.edu} \\
    }
\fi

\begin{document}

\setlength{\parskip}{.45em}

\maketitle

\ifarxiv
\else
    
\fi

\begin{figure}[ht]
    \vspace{-0.3cm}
    \centering
    \includegraphics[width=.8\linewidth]{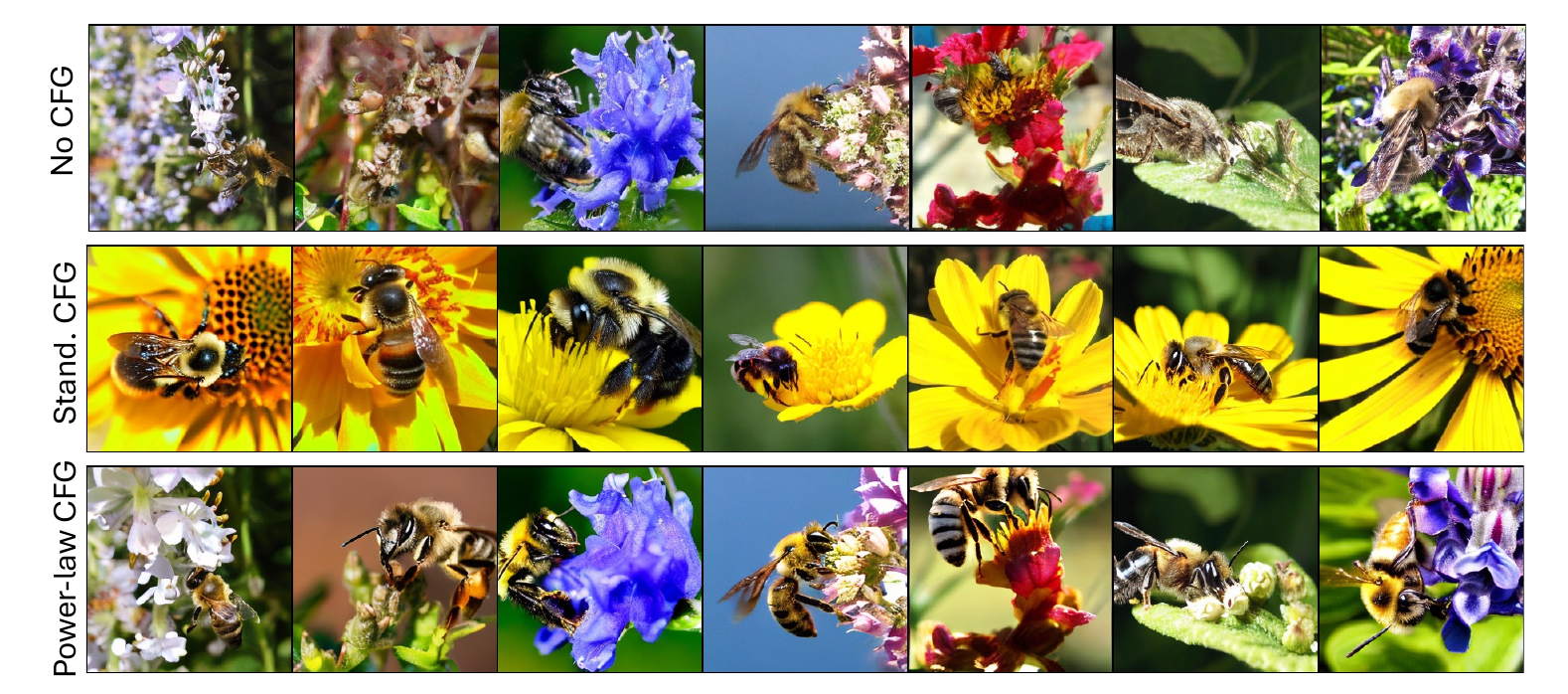}
    \caption{\small\textbf{Qualitative comparison of unguided sampling, standard Classifier-Free Guidance (CFG), and our proposed non-linear power-law version} (DiT/XL-2 on ImageNet-1K $256\times256$). Standard CFG increases fidelity at a substantial expense to diversity and semantic meaning compared to unguided CFG. 
    Our power-law guidance improves fidelity at no cost to semantics or diversity. 
    Samples in each column start from the same seed.}
    \label{fig:intro_img}
    \vspace{-0.2cm}
\end{figure}

\section{Introduction}

\looseness=-1Diffusion \citep{sohldickstein2015deepunsupervisedlearningusing, song2020generativemodelingestimatinggradients, ho2020denoisingdiffusionprobabilisticmodels} and flow-based methods \citep{lipman2022flow, albergo2023stochastic, liu2022flow} have emerged as the de facto state-of-the-art  for generating high-dimensional signals such as images, video, audio and molecular structures. Diffusion relies on Orstein-Uhlenbeck Langevin dynamics, where noise is progressively added to the data until it becomes completely random. New samples are generated by reversing this process through a time-reversed Langevin equation. This backward evolution is steered by a force, the \textit{score}, estimated from the data. In contrast, flow matching circumvents the diffusion construction by directly specifying the probability paths between noise and data. This is done by regressing onto a target vector field which in turn generates the desired probability paths. 
An important task for both paradigms is generating data conditioned on a class  label or  textual description of the image content. This can be achieved through conditioning mechanisms in the model architecture, as well as  guidance techniques  \citep{dhariwal2021diffusion, ho2022classifier} that steer the generation  process towards samples  aligned with user intentions or desired properties.

The notion of guidance was first introduced in  classifier guidance \citep{ song2020score, dhariwal2021diffusion}, where a %time-dependent 
pre-trained classifier is leveraged to induce class conditioning of the sampling. 
Although beneficial, relying on a pre-trained classifier can be computationally expensive and may introduce biases inherent to the classifier itself.
Classifier-free guidance (CFG) \citep{ho2022classifier} was developed as an alternative, and was quickly adopted as a standard technique in 
state-of-the-art generative models \citep{nichol2021glide, betker2023improving, saharia2022photorealistic, esser2024scaling}.
CFG does not rely on an auxiliary classifier, instead, the model is trained to generate unconditional and conditional samples, and at inference extrapolates the denoising path towards the conditional one.
Using  CFG, however, {\it it is no longer guaranteed} to sample the original conditional distribution. 
Indeed, CFG modifies it by  steering it towards a ``mode''  of high-quality and input-consistent samples, while reducing sample diversity in the process~\citep{astolfi2024pareto}.  % and  the force that better guides the generation process toward the desired data.

\begin{figure}
  \centering
    \begin{NiceTabular}{cc}
        \raisebox{.225cm}{
        \includegraphics[width=0.49\textwidth]{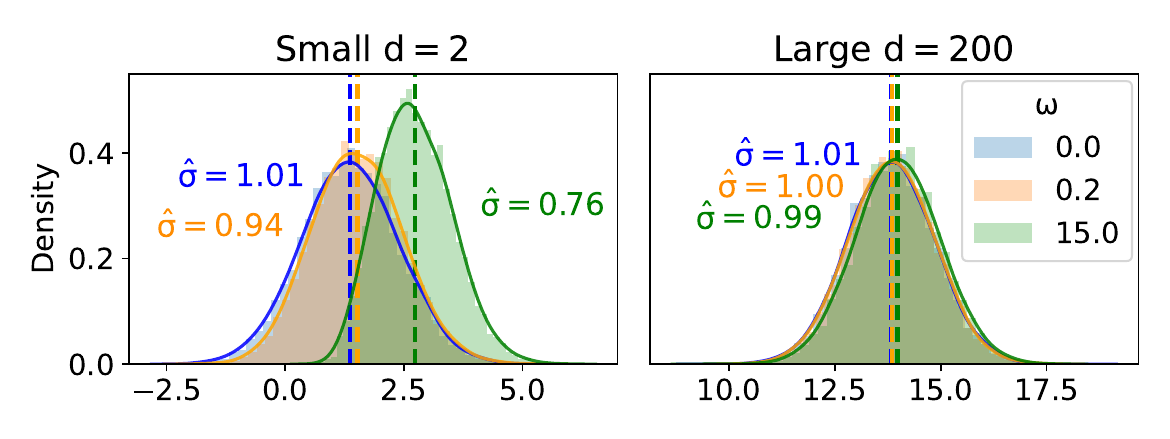}} &
        \hspace{-0.5cm}\includegraphics[width=0.49\textwidth]{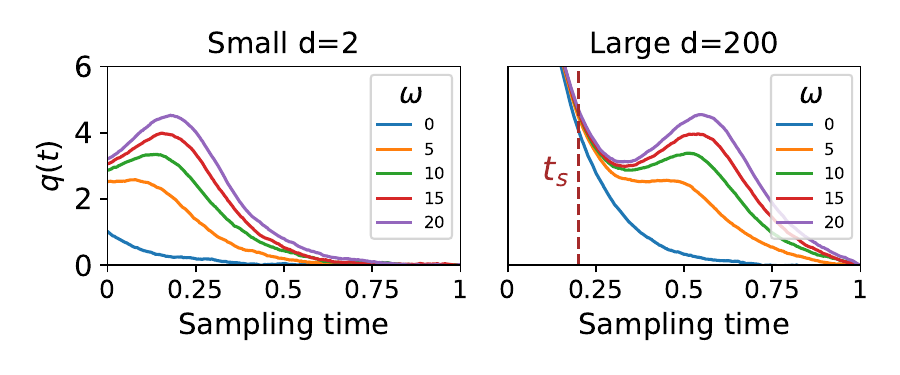} 
    \end{NiceTabular}
    \vspace{-0.3cm}
  \caption{\small\textbf{Left: CFG produces the exact target distribution in high dimensions.} We simulate the backward process using a two Gaussian mixture. We project and plot the generated samples onto the target mean $+\vec{m}$: $q(t=0)=\vec{x}\cdot\vec{m}/|\vec{m}|$. For small $d=2$, CFG generates a distribution with larger magnitude mean (dashed line) and smaller variance than the target one (for $\omega=0.$). This effect diminishes as the dimension increases: for $d=200$ it is practically absent. {\bf Right: High-dimensionality of the data allows CFG trajectories 
  to align.} We plot the evolution of the mean of trajectories $q(t)$: starting at large forward times denoted with $t=1$ (noise), for small $d=2$, CFG trajectories do not align with the unconditional trajectories at $t=0$ (data) causing the CFG overshoot. For large dimension $d=200$, the high-dimensionality of the data allows trajectories to realign with the unguided one at speciation time $t_s$, resulting in the correct target distribution. 
  }
  
  %Model:  mixture of two Gaussians (only one component shown) with dimension $d=2$ (left) and $d=200$ (right), centered in $\pm \vec m$ and variance $\sigma^2=1$, guidance parameter $\omega\in\{0, 0.2, 15\}$ ($\omega=0$ samples without CFG), see \Cref{eqn:gm_0}. We show histograms of $q(t=0)=\vec{x}\cdot\vec{m}/|\vec{m}|$, the generated samples projected onto the normalized mean vector $\vec{m}$ obtained with backward diffusion with 10,000 trajectories. For $d=2$, CFG generates a distribution with larger magnitude mean (dashed vertical line) and smaller variance. This effect diminishes when the dimension increases: for $d=200$ it is practically absent. 
    %{\bf Evolution of the mean  of $\bm{q(t)}$.} It starts from $0$ at large forward time $t$, and is pushed towards positive values (cluster centered in $+\vec m$) when $t$ decreases. It can be observed that CFG is effective when $t\gtrsim t_\textrm{s}$ (vertical line) and stronger for larger values of $\omega$. In Regime \rom{2}, the trajectories of $q$ for various values of $\omega$ merge. }
        \label{fig:intro_motiv}
        \vspace{-0.5cm}
\end{figure}

\looseness=-1The effectiveness of CFG remains surprising in many ways, and a main theoretical question is to characterize the distributions generated by CFG and how they compare to the target distribution. Recent theoretical works on CFG formally showed that in case of Gaussian mixtures in one and finite dimensions, it results in a sharper distribution than the target one, and more shifted towards the boundary of the class \citep{chidambaram2024does, xia2024rectified, wu2024theoretical, bradley2024classifier}. This effect, which  is exemplified in \Cref{fig:intro_motiv} for a two-dimensional Gaussian mixture, is similar to what found by practicioners, see \eg, \citet{saharia2022photorealistic}. 
From the theoretical point of view, it is important to analyze cases in which data is very high-dimensional, as in real applications, to assess also in this context the properties of the distributions generated by CFG, and how they compare to the target one. 
In this study, we address these questions developing a high-dimensional analysis of CFG, and use our high-dimensional results as guidelines to further enhance CFG's practical application.

%\paragraph{Our Contributions.} We show that  high-dimensionality of the data  results in distinct dynamical regimes, allowing CFG to reproduce the target distribution (exemplified in Fig. \ref{fig:1_histograms}). Our findings not only extend previous results to higher and infinite dimensions but also situate CFG within a broader family of guidance strategies. These encompass more effective guidance schemes that enhance performance in terms of image quality and diversity across class-conditional and text-to-image diffusion and flow-matching models. 
% Our contributions are therefore two-fold:

In summary, our contributions are  two-fold:

\looseness=-1\textbf{(1)}. We theoretically describe CFG's behavior in high and infinite dimensions. We precisely characterize how increasing dimension affects mean overshoot and variance shrinkage. By linking CFG to the emergence of dynamical regimes \citep{biroli2024dynamical}, we show that in sufficiently high dimensions, CFG-guided paths realign with those of the unguided conditional path that generates the unmodified distribution. Therefore, CFG can indeed generate the target distribution, and its role is to accelerate sample convergence to the desired class. This path alignment coincides precisely with the symmetry-breaking and class formation \citep{biroli2023generative, raya2024spontaneous}. We demonstrate our theory aligns with numerical simulations and state-of-the-art diffusion and flow-matching experiments~\citep{sadat2023cads, sehwag2022generating}.

\textbf{(2)}. Using our developed theory, we put forward a family of guidance strategies generalizing CFG. We experimentally demonstrate their desirable properties: reduced overshoot, dampened variance shrinkage and faster convergence to the target distribution. We apply these 
to state-of-the-art diffusion and flow matching models, showcasing improved sample quality, consistency, and diversity. %class-conditional models (DiT  \citet{peebles2023scalable} and EDM2  \citet{karras2024analyzing}) trained on ImageNet-1K \citep{deng2009imagenet}, as well as text-to-image models (MMDiT,  \citet{esser2024scaling} and MDTv2, \citet{gao2023masked}) trained on CC12M \citep{changpinyo2021conceptual}, YFCC100M \citep{thomee2016yfcc100m} and a large internal dataset consisting of 320M Shutterstock images. 

\section{Related work}

 Introducing CFG, \citet{ho2022classifier} highlighted the trade-off between image quality, measured by Fréchet inception distance (FID, \citet{heusel2017gans}), and diversity, measured by inception score \citep{salimans2016improved} when adjusting the guidance strength parameter $\omega$. Since then, a significant body of research has examined CFG from various perspectives.

\looseness=-1\paragraph{Theoretical works on CFG.} Several works employed Gaussian mixture models (GMMs) to analyze diffusion and guidance, including \citet{shah2023learning, liang2024unraveling, cui2023analysis, bai2024expectation, song2020score}. In contrast, \citet{du2023reduce} explored alternative conditioning, while \citet{bradley2024classifier} characterized CFG as a predictor-corrector \citep{song2020score}. Most relevant to this work, \citet{chidambaram2024does} demonstrated CFG's mean overshoot and variance shrinkage in one-dimensional settings, while \citet{wu2024theoretical} extended the findings to multi-dimensions using GMMs. We expand on these by developing a high-dimensional statistical analysis and precisely characterizing how these effects diminish as dimensionality increases, ultimately demonstrating that the CFG-generated distribution in fact aligns with the target one for $d\rightarrow \infty$.

\looseness=-1\paragraph{CFG variants and experimental analyses.} Among experimental analyses of CFG, \citet{karras2024guiding} propose guiding generation using a less-trained version of the model, \citet{kynkaanniemi2024applying} apply CFG during a limited interval, and \citet{wang2024analysis} use weight schedulers for the classifier strength parameter. 
Several other CFG alternatives have been proposed, such as rectified guidance \citep{xia2024rectified}, projected score guidance \citep{kadkhodaie2024feature}, characteristic guidance \citep{zheng2023characteristic}, second-order CFG \citep{sun2023inner}, CADS \citep{sadat2023cads}, CFG++ \citep{chung2024cfg++}, REG \citep{xia2024rectified} and APG \citep{sadat2024eliminating}. In later sections, we demonstrate our framework generalizes to these variants, consistently enhancing performance.

\looseness=-1\paragraph{Dynamical regimes, statistical physics and high-dimensional settings.} Statistical physics methods have shown particularly useful in analyzing high-dimensional generative models, \eg, data from Curie-Weiss models \citep{biroli2023generative}, high-dimensional Gaussian mixtures \citep{biroli2024dynamical}, and hierarchical models \citep{sclocchi2024phase}. Furthermore, several recent works studied dynamical regimes diffusion models \citep{biroli2023generative,raya2024spontaneous,biroli2024dynamical,sclocchi2024phase,yu2024nonequilbrium,li2024critical,aranguri2025optimizing}, however none of them analyzed the effects brought by classifier-free guidance.
%Other relevant studies include \citet{ghio2024sampling}, who provided a comprehensive theoretical comparison between flow, diffusion, and autoregressive models from a spin glass perspective; \citet{achilli2024losing}, who extended the theory of memorization in generative diffusion to manifold-supported data; \citet{cui2025precise,cui2023analysis}, who analyzed sample complexity for high-dimensional Gaussian mixtures. A rigorous formulation of diffusion models in infinite dimensional setting was developed by \citet{pidstrigach2023infinitedimensionaldiffusionmodels}.  
 
%%%%%%%%%%%%%%%%%%%%%%%%%%%%%%

\section{Background and high-level discussion}
\label{sec:2}

We begin by providing an overview of the standard framework for generative diffusion, serving as the foundation for our analysis\footnote{For clarity of presentation, our exposition focuses on diffusion, though our findings directly extend to flow-matching with Gaussian paths, as discussed in Sec.\ 4.10.2 of \citet{lipman2024flow}.}. We let $\{\vec{a}_i\}_{i=1}^n \in \mathbb{R}^d$ represent $n$ independent data points sampled from the \textit{true} underlying data distribution $P_0(\vec{a})$ that we aim to model. 

\subsection{General setup} 

The forward diffusion process, starting from the data points $\{\vec{a}_i\}_{i=1}^n$, is modeled by an Ornstein-Uhlenbeck process, described by the following stochastic differential equation (SDE):
\begin{equation}
    d \vec{x}(t) = -\vec{x}(t) \, dt + \sqrt{2} \, d \vec{B}(t),
\label{eqn:original_ou_process}
\end{equation}
where $d \vec{B}(t)$ denotes the standard Brownian motion in $\mathbb{R}^d$. At any given time $t$, the state $\vec{x}(t)$ is distributed according to a Gaussian with mean $\vec{a} e^{-t}$ and variance $\Delta_t = 1 - e^{-2t}$. The forward process is terminated at time $t_f \gg 1$, when $\vec{x}(t_f)$ is effectively pure Gaussian noise, distributed as $\mathcal{N}(0, \mathcal{I}_d)$, with $\mathcal{I}_d$ being the identity matrix in $\mathbb{R}^d$. %Note that for $\{\vec{a}_i\}_{i=1}^n$ drawn from $P_0$, the distribution $P_t(\vec{x})$ at time $t$ is the convolution of the original distribution $P_0$ with a Gaussian kernel.

The backward diffusion process operates in reverse time $\tau = t_f - t$, described by the following SDE:
\begin{align}
    d \vec{x}(\tau) = \vec{x}(\tau) \, d\tau + 2\vec S(\vec{x}, \tau) \, d\tau + \sqrt{2} \, d \vec{B}(\tau),
\label{eqn:backw_dyn}
\end{align}
where $\vec S(\vec x,t)=\vec\nabla \log P_t(\vec x)$ denotes the score function. The backward diffusion process generates points $\vec{x}$ sampled from the distribution $P_t(\vec{x})$ for every time step $\tau$. At the end of the backward process, \ie, when $\tau = 0$, the process generates points drawn from the original distribution $P_0$.

In this work, we focus on generating data that can be categorized into distinct classes. We begin by assuming that the underlying data distribution is a $d$-dimensional probability distribution $P_0(\vec{x}, c)$, where $c$ represents a discrete class index and $\vec{x}$ a $d$-dimensional vector. The aim is to generate data conditioned on $c$, the class label.  The procedure that is mathematically guaranteed to generate the exact conditional  target distribution consists of using the true conditional score, $\vec S_t(\vec x,c)=\vec\nabla \log P_t(\vec x|c)$ in  \Cref{eqn:backw_dyn}. CFG, however,  does not do that; it instead further directs diffusion in a manner proportional to the difference between conditional and unconditional scores:
\begin{align}
S_t^{\text{CFG}}(\vec{x}, c) = S_t(\vec{x}, c) + \omega [S_t(\vec{x}, c)-S_t(\vec{x})].
\label{eqn:score_CFG_linear}
\end{align}
Although CFG may offer practical advantages, such as enhanced fidelity and classification confidence \citep{wu2024theoretical}, a key question remains: whether CFG is at all capable of accurately generating the desired target distribution. 
%In  \Cref{sec:gauss_mixt}, we will provide an affirmative answer to this question.

\begin{figure}[t]
        \centering
        \begin{NiceTabular}{cc}
            \hspace{-0.65cm}
            \includegraphics[width=0.562\columnwidth]{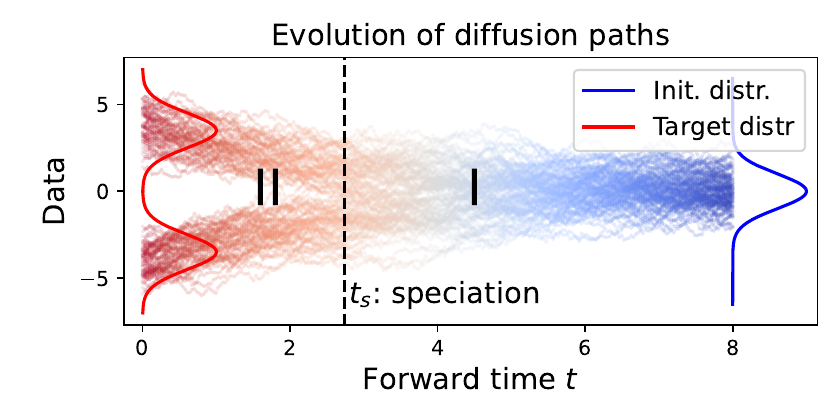} &
            \hspace{-0.6cm}
            \includegraphics[width=0.32\columnwidth]{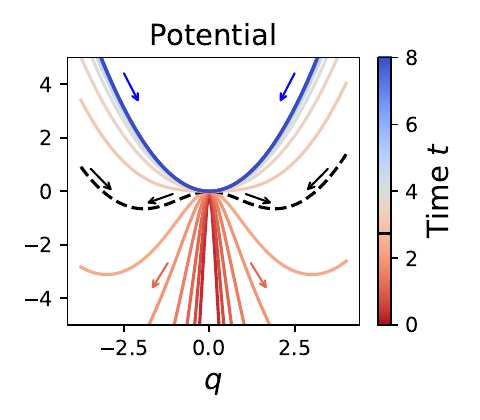} 
        \end{NiceTabular}
        \vspace{-0.25cm}
        \caption{\small\textbf{Dynamical regimes in diffusion.} \textbf{Left:} Illustration of the speciation phenomenon using a one-dimensional Gaussian mixture. Starting from pure Gaussian noise at large time $t$, the backward diffusion begins in Regime \rom{1}, where the class has not been decided yet. After speciation time $t_\textrm{s}$ (dashed line), the class membership is decided.
        %(speciation is a cross-over in $d=1$, it becomes sharp for large $d$). 
        \textbf{Right:} Evolution of the effective potential (conditional potential in Eq. (\ref{eqn:eff_pot})) over time for high-dimensional Gaussian mixture showcasing the symmetry breaking phenomenon.
        }
        \label{fig:2_diffusion_regimes}
        \vspace{-0.3cm}
\end{figure}

\subsection{Connecting dynamical regimes of diffusion to classifier-free guidance} 

\looseness=-1 Our analysis adopts the approach outlined by \citet{biroli2023generative} and \citet{biroli2024dynamical}, which identifies three distinct regimes: our exposition focuses on the first two, as the effect of CFG is the same in the second and third one. The first two regimes are distinguished by symmetry-breaking, characterized through the eigenvalue of the principal component of the data covariance matrix. %\footnote{We include the definition for clarity, ensuring that it is clear which definition of the dynamical regimes we are using.}

\paragraph{CFG and distinct dynamical regimes.} \looseness=-1\citet{biroli2024dynamical} analyze the dynamical regimes of the backward process in \Cref{eqn:backw_dyn} for two classes with $d\to\infty$. They identify \emph{speciation time} $t_\textrm{s}$ as the transition between the first and second regime. 
In Regime \rom{1}, the backward trajectories have not yet committed to a particular class of data.
In  Regime \rom{2}, the trajectories have committed to a class and generate the features necessary to produce samples from that class.
%Additionally, in cases with more structured data with several classes and subclasses, multiple speciation times can exist, e.g., see \citet{li2024critical,sclocchi2024phase}. 
As exemplified in \Cref{fig:2_diffusion_regimes}, in Regime \rom{2}, the probability $P_t(\vec x)$ consists of separated non-overlapping lumps corresponding to the target classes.

In our work, we show that CFG is beneficial in Regime \rom{1} as the class membership of the trajectories has not been decided yet, whereas in Regime \rom{2} the well-separated probability lumps corresponding to different classes make CFG redundant. This can be stated as three results: 
%In other words, we obtain the following three findings: 

\textbf{Result \rom{1}.} Before speciation time $t_\textrm{s}$, CFG is effective in aiding class selection and speeds up the convergence towards the target class.

\textbf{Result \rom{2}.} Just before speciation time $t_\textrm{s}$, CFG-guided paths realign with the unguided path that generates the correct, unmodified target distribution.

\textbf{Result \rom{3}.} After speciation time $t_\textrm{s}$, CFG has no effect on the generation process.

In \Cref{sec:gauss_mixt}, we substantiate these results with  theoretical arguments for the case of Gaussian mixtures, showing first that for $d\rightarrow \infty $ CFG reproduces the correct target distributions, and then characterizing the finite-$d$ corrections.
In \Cref{sec:non_lin_class_guid} we then demonstrate their applicability to enhance performance of real-world models.

% We first provide a general explanation of CFG in high dimensions, followed by an in-depth GMM analysis. Assuming $K$ classes, the unconditional score can be written in terms of conditional scores:
%  \begin{align}
%     \vec{S}_t(\vec{x})=\vec \nabla \log P_t(\vec x)= \frac{\sum_{c=1}^K P_t(\vec x|c)p(c)\; \vec{S}_t(\vec{x}, c)}{\sum_{c=1}^K P_t(\vec x|c)p(c)},
%      \label{eq:basicCFG}
%  \end{align}
%  where $p(c)$ is the probability of class $c$ in the original data. 

% In high-dimensional settings, the speciation phenomenon ensures that once the backward process enters Regime \rom{2}, any sample  $\vec{x}$ drawn from $P_t(\vec{x})$  will almost surely belong to a single, well-defined class  $c$.
%  If $\vec x$ is in the lump associated with class $c$, then $ \sum_{c'=1}^K P_t(\vec x|c')p(c')\rightarrow P_t(\vec x|c)p(c) $. \Cref{eq:basicCFG} then implies that the full score equals the one computed in the class $c$, i.e., $\vec{S}_t(\vec{x},c)=\vec{S}_t(\vec{x})$, 
%  and the additional guidance term introduced by CFG vanishes. Hence,
% CFG has no effect in Regime \rom{2}. In contrast, during Regime \rom{1}, before speciation occurs, CFG steers the dynamics toward the cluster $c$ faster. The crucial point lies in demonstrating that the CFG-guided paths converge with the unguided, class-conditional path by the end of Regime \rom{1}, as we will illustrate in the following section.

\section{CFG in the high-dimensional limit of Gaussian mixtures}
\label{sec:gauss_mixt}

Distinct dynamical regimes emerge in a wide range of generative models and across various data modalities \citep{ventura2024manifolds, george2025analysis, bae2024very}. To examine the dynamical regimes, we adopt the two-Gaussian mixture framework of \citet{biroli2023generative}, which has been tested on real data and shown to hold for models of data lying on manifolds \citep{biroli2024dynamical}\footnote{Our analysis can be extended to any number of Gaussians, with different variances, or data supported on manifolds, by following e.g., \citet{achilli2025memorization, george2025analysis}. We provide discussion in \Cref{sec:appx_C_extended_proof}.
}.
%We subsequently demonstrate that our GMM theory is not only supported by state-of-the-art models, but can also be readily applied to enhance their performance. 

\subsection{Theoretical framework} 
We examine the case where $P_0(\vec{a})$ is a superposition of two Gaussians with equal weight, means $\pm \vec{m}$ and isotropic variance $\sigma^2$. To ensure the two Gaussians are well separated, we take the large $d$ limit with fixed values of $|\vec{m}|^2 / d $ and  $\sigma$. We assume that the exact scores are available. % is correctly estimated. 
%The extension to more than two Gaussian is straightforward and discussed in App. \ref{sec:appx_4mg}.
% To simplify the notation, we focus on the SDE forward equation corresponding to $f(t)=-1$ and $g(t)=\sqrt{2}$ (see \Cref{eqn:original_ou_process}). All results can be easily translated for general choices of $f(t),g(t)$.  

In this setting, the speciation transition between Regimes \rom{1} and \rom{2} resembles a symmetry-breaking phenomenon occuring on timescales $t_\textrm{s} = \frac{1}{2} \log(d)$. \citet{biroli2024dynamical} show $t_\textrm{s}$ emerges as the time at which diffusion paths commit to a specific class by relating it to a change of the potential in the backward Langevin equation, as displayed in \Cref{fig:2_diffusion_regimes}. We find that \textit{the speciation time $t_\textrm{s}$ aligns precisely with the time until which CFG is effective in aiding class selection}. Beyond this point, as the trajectories have committed to a class, CFG no longer influences the generated outcome. We now spell out the required theoretical arguments, with detailed proofs presented in \Cref{sec:appx_B_basic_proof} and~\Cref{sec:appx_C_extended_proof}.

\subsection{Key findings: Infinite dimensional limit}
We first rewrite the distribution of $\vec{x}$ at time $t$ as $P_t(\vec{x})\propto[e^{-\left(\vec{x}-\vec{m} e^{-t}\right)^2 /\left(2 \Gamma_t\right)}+e^{-\left(\vec{x}+\vec{m} e^{-t}\right)^2 /\left(2 \Gamma_t\right)}]$, where $ \Gamma_t =1+(\sigma^2-1)e^{-2t}$. In this case, the CFG formula in \Cref{eqn:score_CFG_linear} can be rewritten as:
\begin{align}
        S_t^{\text{CFG}}(\vec{x},c) = -\frac{\vec{x}}{\Gamma_t}  + \frac{c \vec{m} e^{t}}{\Gamma_t}+ \omega\frac{\vec m e^{-t}}{\Gamma_t}\left\{c-\tanh{\left(\frac{\vec{x}\cdot\vec{m}e^{-t}}{\Gamma_t}\right)}\right\},
        \label{eqn:cfg_score_form}
\end{align}
with $c=\pm1$ and $\omega>0$.

% \subsubsection{In Regime \rom{1}, CFG provides an additional push toward the target class}

\mypar{Result \rom{1}: Before speciation time $t_\textrm{s}$, CFG is effective in aiding class selection and speeds up the convergence towards the target class.} \looseness=-1To obtain this result, we examine $S_t^{\text{CFG}}(\vec{x},c)$ (\ref{eqn:cfg_score_form}) in Regime \rom{1}, which lasts until speciation occurs at $t_\textrm{s}=(1/2)\log d$.

\mypar{Which directions of \Cref{eqn:cfg_score_form} are affected by CFG?}
 \Cref{eqn:cfg_score_form} shows that CFG only affects the $\vec m$ directions (as it is the direction multiplied by $\omega$), therefore CFG has no effect on orthogonal directions $\vec{v}\perp\vec{m}$.
This is formally shown by projecting the backward \Cref{eqn:backw_dyn} on a unit vector orthogonal to $\vec m$: the resulting equation $dp=p (1-2/\Gamma_{t_f-\tau})d\tau +\sqrt{2}dB$ equals the backward equation for an initial Gaussian $\mathcal{N}(0, \sigma^2)$, thus not depending on $\omega$ and therefore unaffected by CFG.

\mypar{What happens in the $\vec{m}$ directions?} As CFG only affects $\vec{m}$ directions in \Cref{eqn:cfg_score_form}, let us project onto $\vec{m}$ and observe how the CFG score $\vec{S}_t^{\text{CFG}}$ influences the backward process. Defining $q(t) := \frac{\vec{x} \cdot \vec{m}}{|\vec{m}|}$ where $|\vec{m}| = \sqrt{d}$, the evolution guided to class $c=1$ is given by the following:
\begin{equation}
    \begin{aligned}
        d q = 
        \Big(q + 2\Big[-q+ e^{-(t_f-t_\textrm{s}-\tau)} \Big((1+\omega) - \omega\tanh{\left(qe^{-(t_f-t_\textrm{s}-\tau)}\right)}\Big)\Big]\Big)d\tau+d\eta(\tau),
    \end{aligned}
    \label{cfg:eqn}
\end{equation}

where $\tau=t_{f}-t$, with $t_\textrm{s}=(1/2)\log d$.  Here, $\eta(\tau)$ denotes $\sqrt{2}$ times a Brownian motion, and we used the fact that in Regime \rom{1} we have $\Gamma_t\simeq 1$, see \citet{biroli2024dynamical}. 
To simplify notation, we omit the dependency $t(\tau)$ for backward time and use $t$ hereafter.
  
By rewriting \Cref{cfg:eqn} as: $d q  = -\pdv{V^\textrm{CFG}(q,\tau)}{q} d \tau+d \eta(\tau)$, we can analyze the effective potential:

\newcommand{\labeleffpot}{}

\begin{align}
    V^\textrm{CFG} = \underbrace{\left(\frac1{2}q^2-2e^{-(t-t_\textrm{s})}q\right)}_\text{Conditional potential} +\ \omega\ \underbrace{\left[-qe^{-(t-t_\textrm{s})}+\ln\cosh{\left(qe^{-(t-t_\textrm{s})}\right)}\right]}_\text{Extra CFG potential $V_{\text{extra}}$}. 
    \label{eqn:eff_pot}
\end{align}

Result \rom{1} follows from \Cref{eqn:eff_pot}: we observe that CFG-added-potential provides an additional push toward the positive values of $q$, corresponding to target class $c=1$. The effect of CFG is particularly strong for trajectories deviating from typical behavior: its effect is particularly prominent when correcting the trajectories going toward the wrong class (see \Cref{fig:4_potentials} in \Cref{sec:appx_B_basic_proof}).

\mypar{Result \rom{2}: CFG paths align before exiting Regime \rom{1}.} During late stage of Regime \rom{1}, $q$ becomes of order $\sqrt{d}$ \citep{biroli2023generative}, while the CFG-added-term in \Cref{eqn:cfg_score_form} gives exponentially small corrections to the SDE. % (of order $e^{-2q e^{-(t_f-t_\textrm{s}-\tau)}}$).
At these late times $\tau_i$ of Reg. \rom{1}, SDE \Cref{eqn:cfg_score_form} simplifies to: $d q = -q+  2 e^{-(t_f-t_\textrm{s}-\tau_i)} +d\eta(\tau_i)$. Although different values of $\omega$ have led to different values of $q(\tau_i)$ during the backward process, we show that the value $q(\tau_i)$ is  exponentially quickly forgotten when $\tau$ departs from $\tau_i$. Therefore, the backward evolution readjusts to the ``correct'' value without CFG. 

Result \rom{2} is therefore obtained by solving the SDE starting from $\tau_i$: $q(\tau)=q(\tau_i)e^{-(\tau-\tau_i)}+e^{-(t_f-t_\textrm{s})}\left(e^{\tau}-e^{-\tau+2\tau_i} \right) +\sqrt{1-e^{-(2(\tau \tau_i))}}z_{\tau}$,
% \begin{align*}
%     q(\tau)=q(\tau_i)e^{-(\tau-\tau_i)}&+e^{-(t_f-t_\textrm{s})}\left(e^{\tau}-e^{-\tau+2\tau_i} \right)\\
%     &+\sqrt{1-e^{-(2(\tau-\tau_i))}}z_{\tau},
% \end{align*}
where $z_\tau$ denotes a standard Gaussian variable. When $\tau \gg \tau_i$ but still in Regime \rom{1}, the solution of the SDE does not depend any longer on $q(\tau_i)$ and it coincides statistically with the one of the backward process of the single Gaussian corresponding to the target class $c=+1$.

\mypar{Result \rom{3}: After speciation time $t_\textrm{s}$, CFG has no effect on the generation process.} At the end of Regime \rom{1}, $q$ diverges so one has to focus on the rescaled variable $\vec{x}\cdot\vec{m}/d$.  
As $q$ has realigned with the value it would have had without CFG (for $\omega=0$), the initial condition ($\vec{x}\cdot\vec{m}/d= 0$) in Regime \rom{2} is independent of $\omega$. 
To see that CFG has no effect in this regime, all that is left to show is that the term CFG-added term (multiplied by $\omega$) in \Cref{eqn:cfg_score_form} is zero. From \citet{biroli2024dynamical} we know that in Regime \rom{2}, $ |\vec{x}\cdot\vec{m}| e^{-t}/\Gamma_t$ is of order $O(d) $ and $ \sign(\vec{x}\cdot\vec{m})=1$. Thus, Result \rom{3} follows by observing the extra CFG term in \Cref{eqn:cfg_score_form} thus equals zero as $1-\tanh{(\vec{x}\cdot\vec{m}e^{-t}/\Gamma_t)}\rightarrow 0$ for $d\rightarrow \infty$.

To summarize during Regime \rom{1}, CFG pushes faster towards the target distribution. Before speciation $t_\textrm{s}$ occurs, the paths realign with the correct, unguided path. Once Regime \rom{2} kicks in, CFG no longer affects the generation process. This can be observed in \Cref{fig:intro_motiv} (right), where \Cref{cfg:eqn} is simulated for $t_\textrm{s}=1000, \sigma=1$, averaging over 10,000 trajectories. This shows that, unlike in low-dimensions where the paths never realign (see, \eg, \citet{chidambaram2024does}), large dimensionality of the data allows CFG to indeed generate the correct target distribution, as seen in \Cref{fig:intro_motiv} (left).

\subsection{Key findings: Finite dimensional setting}
\label{sec:fin_dim_eff}

So far, we have shown that for any value of $\omega$ the target distribution is correctly reproduced in the infinite-$d$ limit. We now consider the changes brought by finite $d$. 

\paragraph{Consistent conclusions in large, finite $d$.}\looseness=-1 Within Regime \rom{1}, for large, yet finite dimension, the CFG-added-term in the score in \Cref{eqn:cfg_score_form} remains of the same order as the conditional score of the unguided path so CFG has the same effect as in the infinite limit. When exiting Regime \rom{1} and during Regime \rom{2}, the extra CFG term is zero for $d\rightarrow\infty$, and exponentially small in $d$ for finite $d$, so the remaining two results carry over as well.

\paragraph{Mean overshoot and variance shrinkage in small $d$.} 

In low dimensions, the paths will not realign when exiting Regime \rom{1}. The additional push introduced by CFG within Regime \rom{1} will have an effect on Regime \rom{2}, resulting in an overshoot of the target distribution of relative amplitude of order $1/\sqrt{d}$. The CFG-added-term also results in a larger second derivative of the potential $V^\textrm{CFG}(q,t)$. Thus, the resulting CFG Langevin equation is associated to a more confining potential, ultimately shrinking  the variance of the CFG-generated distribution. These are in line with previous empirical \citep{ho2022classifier} and theoretical findings \citep{chidambaram2024does, wu2024theoretical}.

\looseness=-1So far, we have shown that CFG correctly generates the target distribution in infinite dimensions and modifies in lower dimensions in a way that is consistent from low dimensions to large dimensions  \citep{chidambaram2024does, wu2024theoretical}. 
These theoretical results support the use of CFG in practical applications, by establishing general properties of the CFG-generated distributions.
This raises the question: can we use the high-dimensional results as a guideline to design guidance schemes that enjoy similar properties of standard CFG, and in particular preserve the same behavior in high and infinite dimensions? In the next section, we provide an affirmative answer and present new CFG-procedures.

\begin{figure}[t]
        \centering
        \begin{minipage}{0.66\columnwidth}
            \includegraphics[width=.5\textwidth]{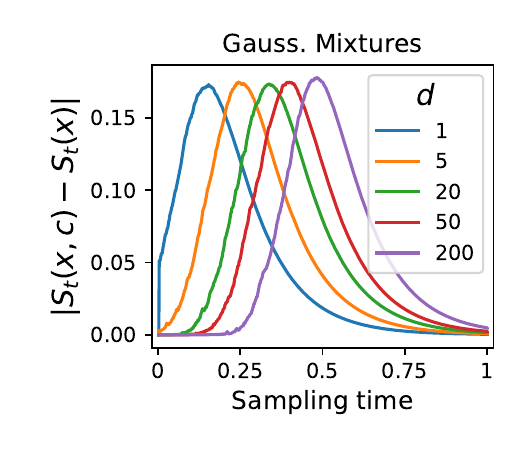}
            \includegraphics[width=.5\textwidth]{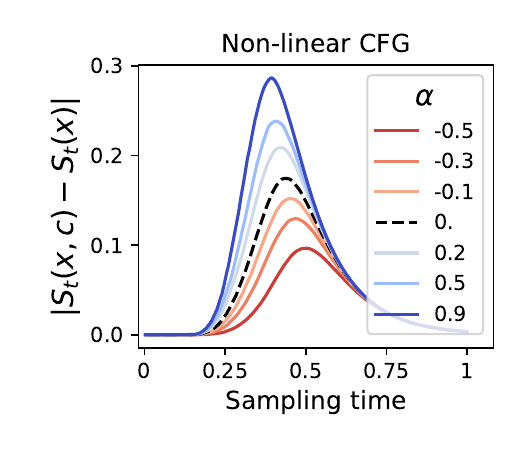}
        \end{minipage}
        \hfill \vline \hfill
        \begin{minipage}{0.33\columnwidth}
            \includegraphics[width=.95\textwidth]{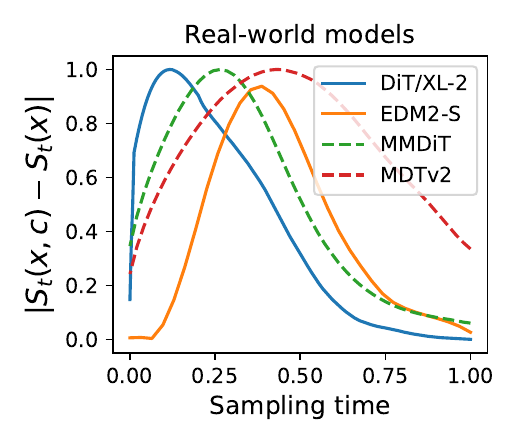}
        \end{minipage}
        \caption[Caption for LOF]{\small\looseness=-1\textbf{Evolution of the CFG score difference, from noise ($t=1$) to data ($t=0$).} \textbf{Left (stand. CFG):} 
        Numerically simulating mixture of two Gaussians: as $d$ increases, the score difference becomes substantial earlier (this happens during Regime \rom{1}). \textbf{Middle (non-lin. CFG, $d=200$):} Non-linear CFG parameter $\alpha$ allows more flexible behavior of the score difference. \textbf{Right (stand. CFG):} Real-world experiments using advanced models show consistent behavior with theory: monotonically increasing score difference followed by decay after a certain point. Experimental details are provided in App. \ref{sec:appx_E_gm} and \ref{sec:appx_F_real_world}.
        %Numerically simulating mixture of two, four, and eight Gaussians with equidistant means on a sphere ($r=\sqrt{d}$), with varying dimension $d$, with $\omega=4, \sigma^2=1$, averaged over 10,000 trajectories. As $d$ increases, the score difference is substantial on earlier backward times $\tau$ (in Regime \rom{1}). Additionally, as the number of classes increases, the magnitude of the score difference grows, as well as the duration of large difference between the scores. \textbf{Right:} We replicated the same experiment using  class-conditional (DiT/XL-2 and EDM2-S) and text-to-image (MMDiT and MDTv2) diffusion models. For presentation clarity, we normalized x-axis timesteps and y-axis score difference to range $[0,1]$. We observed a consistent pattern with theory: monotonically increasing score difference followed by decay after a certain point.
        }
        \label{fig:6_cfg_score_diff}
        \vspace{-0.35cm}
\end{figure}

\section{Generalized classifier-free guidance}
\label{sec:non_lin_class_guid}

The ``blessing of dimensionality'' that allows CFG to generate the target distribution in high-dimensions is due to two main properties: (1) CFG acts in Regime \rom{1} pushing stronger toward the desired class, (2) CFG does not play any role in Regime \rom{2} where the detailed properties of the data are generated. There is, however, a larger class of guidance schemes that also enjoy these properties. As a straightforward but effective extension, we introduce non-linear variants of CFG.

\subsection{Non-linear classifier-free guidance}
\label{sec:nlg}

We  consider non-linear versions of score-based guidance of the form:
\begin{flalign}
        \label{eqn:nonlin_cfg}
        S_t^{\textrm{CFG-NL}}&(\vec{x},c)= S_t(\vec{x}, c) + \left[S_t(\vec{x}, c) -  S_t(\vec{x})\right] \phi_t \left(\left|\vec{S}_t(\vec{x}, c) -  \vec{S}_t(\vec{x})\right|\right).
\end{flalign}
\looseness=-1For constant $\phi_t(s)=\omega$, \Cref{eqn:nonlin_cfg} reduces to standard CFG.
As long as the function  $\phi_t(s)$ satisfies $\lim_{s\to 0} \left[s \phi_t(s)\right]{}=0$, the arguments from Results \rom{2}-\rom{3} imply that in Regime \rom{2} the extra contribution to the score due to $\phi_t$ vanishes, thus leading to a correct target distribution in high-dimensions. 
The freedom in the choice of $\phi_t$ can be used to improve the effect of CFG in Regime \rom{1}, helping to push the system in the direction of class $c$, while reducing the unwanted finite-dimensional drawbacks. 
In the following, as a proof of principle, we propose a first example for $\phi_t$ (another is discussed in \Cref{sec:appx_G_nonlin_cfgs}). 
As we shall show, this choice already allows to improve state-of-the-art generative models. 
Ultimately,  the whole function $\phi_t$  may be optimized as a hyperparameter. 
 
\subsection{Power-law CFG}
 
We choose $\phi_t(s)=\omega s^{\alpha}$ with $\alpha>-1$ to obtain the following guidance scheme:
\begin{align}
    \vec S_{t}^\textrm{PL}(\vec{x},c)= S_t(\vec{x}, c)+ \omega \left[S_t(\vec{x}, c) -  S_t(\vec{x})\right] \left|\vec{S}_t(\vec{x}, c) -  \vec{S}_t(\vec{x})\right|^{\alpha}. 
    \label{eqn:power_law}
\end{align}
\looseness=-1One can understand the effect of non-linear guidance as follows. The $\ell_2$ distance between scores $\delta S_t =|\vec{S}_t(\vec{x}, c) -  \vec{S}_t(\vec{x})|$ is exponentially small both at the beginning of the backward process (as both cond. and uncond. distributions are standard Gaussian) and before exiting Regime \rom{1} (as shown in \Cref{sec:gauss_mixt}), after which it remains zero. The non-linear scheme, while automatically switching off in Regime \rom{2}, allows altering the shape of $\delta S_t$ during Regime \rom{1}.
Choosing $\alpha<0$ provides guidance which speeds up convergence to the target at early times, while $\alpha>0$ dampens the guidance for small $\delta S_t$ and strengthens it for large $\delta S_t$. 
In practice, we found positive values for $\alpha$ to perform best.
 
%  \begin{figure*}[t]
%     \centering
%       \begin{NiceTabular}{ccc}
%           \Block[borders={right}]{1-1}{}
%           \begin{minipage}[b]{0.46\textwidth}
%             \centering
%             \vspace{-0.5cm}
%             {\footnotesize Improved image quality}
%             \includegraphics[width=\textwidth]{assets/plots/plots_1a.pdf}
%           \end{minipage} &
%           \begin{minipage}[b]{0.46\textwidth}
%             \centering
%             {\small Higher sample diversity}
%             \includegraphics[width=\textwidth]{assets/plots/plots_1.pdf}
%           \end{minipage} &
%       \end{NiceTabular}
%     \caption{Qualitative comparison between Standard CFG ($\omega=4$) and Power-Law CFG ($\omega=4, \alpha=0.1$) on DiT/XL-2 trained on ImageNet-1K at  256$\times$256 resolution. \textbf{Left panel:} Each paired image starts from same initial noise with standard (left) and Power-Law CFG (right). Power-Law often exhibits improved details which might contribute to the improved FID. \textbf{Right panel:} For three classes, starting from same seeds (0, 1 and 2), Power-Law CFG generates images with higher diversity. Further qualitative examples are provided in App. \ref{appx:ex_by_dit}.}
%     \label{fig:qualitative_expls}
% \end{figure*}

\looseness=-1\paragraph{Additional Non-Linear CFG forms.}
We note that the framework in \Cref{eqn:nonlin_cfg} covers many concurrent works on alternative CFG forms: with $\phi_t(s) = \omega \cdot \mathbb{I}_{[t_1,t_2)}(t)$, we obtain limited-interval CFG by \citet{kynkaanniemi2024applying}. Using $\phi_t(s)=\omega_t$ yields CFG weight schedulers as in \citet{wang2024analysis, gao2023masked}. Guidance schemes by \citet{chung2024cfg++, xia2024rectified, ventura2024manifolds} also yield simple $\phi_t(s)$ expressions. However, all of the aforementioned works use a guidance term that is linear in the score  difference $\delta S_t$. In the experiments below, we find that non-linear power-law guidance $\vec S_{t}^\textrm{PL}$ improves over these existing linear methods.

\subsection{Generative image model experiments}
\label{sec:nonlin}

\looseness=-1\paragraph{Experimental details.} We examine power-law CFG (\ref{eqn:power_law}) GMM simulations, and four generative models: DiT \citep{peebles2023scalable} and EDM2 \citep{karras2024analyzing}, trained and evaluated on ImageNet-1K (resolutions 256 and 512). 
We also consider two text-to-image models: first is trained on ImageNet-1K and CC12M \citep{changpinyo2021conceptual}, evaluated on CC12M, using the diffusion DDPM training objective \citep{ho2020denoisingdiffusionprobabilisticmodels} with MMDiT architecture (\citet{esser2024scaling}, similar to SD3). 
The second model, using MMDiT scaled to 1.6B parameters, is trained with flow matching on YFCC100M \citep{thomee2016yfcc100m}, CC12M and a proprietary dataset of 320M Shutterstock images, evaluated on COCO dataset \citep{lin2014microsoft}. \Cref{sec:appx_F_real_world} contains a third text-to-image model trained with DDPM objective with the MDTv2 \citep{gao2023masked} architecture scaled to 800M parameters.

%
%\paragraph{Datasets used for evaluation.} We conducted a comprehensive evaluation on three datasets with varying scales and distributions: ImageNet (IMNET), widely used benchmark dataset for image classification tasks. Conceptual 12M (CC12M): A large-scale dataset containing images and captions. Shutterstock (SHST): An internal dataset comprising 320 million Shutterstock images. 
We blurred human faces in ImageNet-1K and CC12M, and utilized Florence-2~\citep{xiao2023florence} to recaption images for more accurate image content descriptions.

\looseness=-1\paragraph{Comparing GMM simulations to real-world experiments.} 
In \Cref{fig:6_cfg_score_diff} (first and third panel) we observe similar hump-shaped behavior of the difference between conditional and unconditional score $|S_t(\vec{x},c)-S_t(\vec{x})|$ for GMMs and real-world models, validating the applicability of our theoretical findings. Furthermore, using the parameter $\alpha$ in Power-law CFG, we can alter the shape of these curves,  obtaining a more flexible framework generalizing standard CFG (see \Cref{fig:6_cfg_score_diff} central panel). 
This enables faster convergence, and as we show in \Cref{fig:large} in \Cref{sec:appx_G_nonlin_cfgs} yielding paths with consistently smaller Jensen-Shannon divergence to the target distribution across all time $\tau$ and reducing the overshoot of the target distribution.

\begin{wrapfigure}{r}{0.57\columnwidth}
    \vspace{-0.6cm}
    \begin{minipage}[t]{0.25\columnwidth}
        \includegraphics[height=3cm]{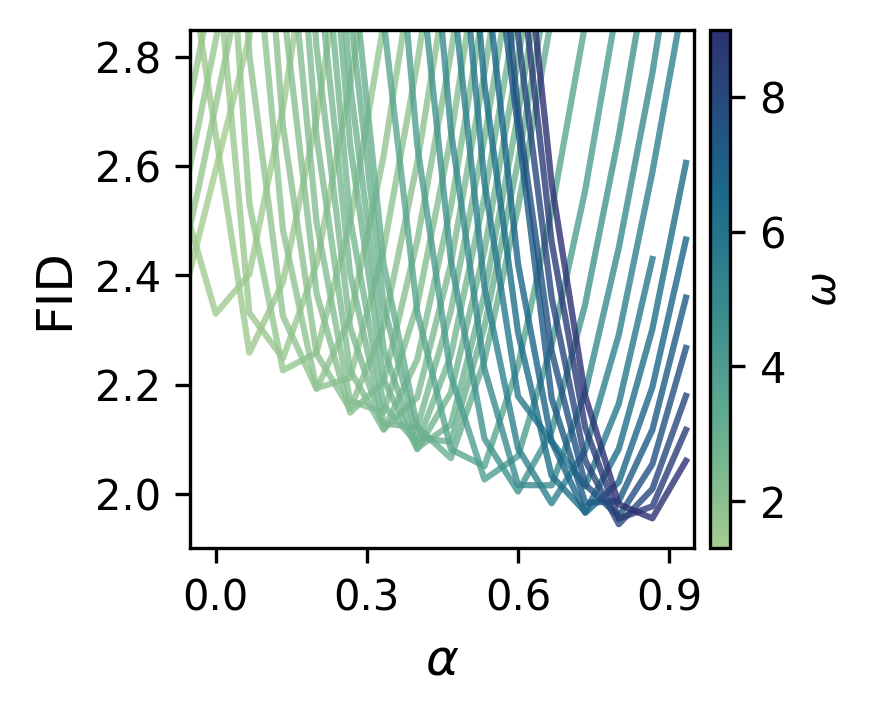}
    \end{minipage}%
    \hfill%
    \begin{minipage}[t]{0.3\columnwidth}
        \includegraphics[height=3cm]{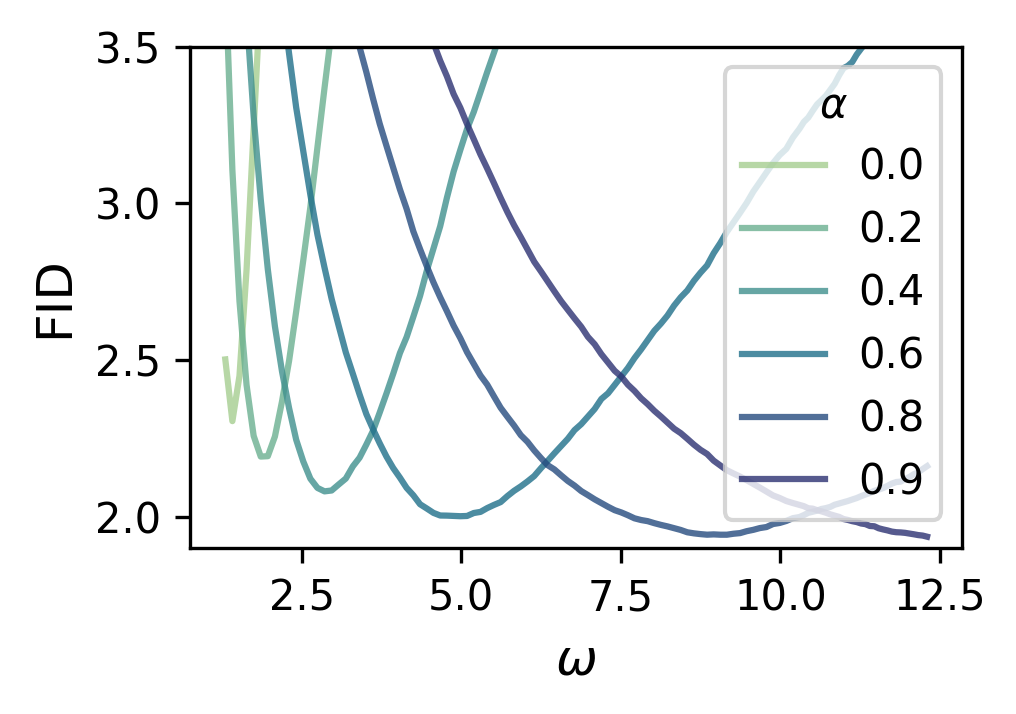}
    \end{minipage}
    \vspace{-0.4cm}
    \caption{\small\looseness=-1 \textbf{Sensitivity analysis} (EDM2-S, ImageNet-1K $512\times512$). \textbf{Left:} Increasing parameter $\alpha$ consistently improves FID to standard CFG ($\alpha=0.$). \textbf{Right:} Increasing $\alpha$ yields more stable  FID values across a larger range of $\omega$.}
    \vspace{-0.4cm}
    \label{fig:sens_analysis}
\end{wrapfigure}

\paragraph{Power-law CFG is robust.} We perform sensitivity analysis, showing that large values of $\alpha$ consistently yield improved performance, increasing robustness and stability when tuning for $\omega$.\footnote{Although power-law CFG introduces another hyerparameter, $\alpha$, we did not have to perform extensive hyperparameter search, and found large values, \eg, $\alpha=0.9$ to consistently perform well.} 
This is shown in  \Cref{fig:sens_analysis} for EDM2-S and in \Cref{sec:appx_F_real_world} for DiT/XL-2 and two T2IM models, together with further ablation studies showing that non-linear CFG consistently outperforms standard CFG when varying number of sampling steps.

\begin{table}
  \caption{\small\textbf{Power-law CFG often improves both fidelity and diversity metrics}. We applied power-law to standard CFG and limited and CADS variants, as the two were the strongest competitors. Applying power-law improved their performance further, achieving competitive results. Best results are \textbf{bolded}, second best \underline{underlined}. ($\textcolor{green}{\uparrow}$) indicates power-law CFG improves the guidance method compared to its version with stand.\ CFG, while ($\textcolor{red}{\downarrow}$) means the metric deteriorated. T2IM represents text-to-image models, CC class-conditional; FM is short for flow-matching objective and diff. stands diffusion. Experimental details are provided in  \Cref{sec:appx_F_real_world}.}
  \centering
  \resizebox{.99\textwidth}{!}{\begin{tabular}{l|cccHH|cccHH|cccHH|cccHH}
    \toprule
    & \multicolumn{5}{c}{EDM2-S (CC, IM-1K 512)} & \multicolumn{5}{c}{DiT/XL-2 (CC, IM-1K 256)} & \multicolumn{5}{c}{Diff. MMDiT (T2IM, CC12m)} & \multicolumn{5}{c}{FM MMDiT (T2IM, COCO)} \\
    \midrule
    Model & FID & Precision & Recall & Density & Coverage & FID & Precision & Recall & Density & Coverage & FID & Precision & Recall & Density & Coverage & FID & Precision & Recall & Density & Coverage \\
    \midrule
    Standard \citep{ho2022classifier} & 2.29 & 0.751 & 0.582 & 0.850 & 0.764 & 2.27 & 0.829 & 0.584 & &  & 8.58 & 0.661 & 0.569 & 1.091 & 0.840 & 5.20 & 0.629 & 0.594 & 0.902 & 0.772 \\
    Scheduler \citep{wang2024analysis} & 2.03 & 0.762 & 0.591 & 0.867 & 0.780 & 2.14 & 0.840 & 0.614 & & & 8.30 & 0.681 & 0.559 & 1.266 & 0.860 & 5.00 & 0.606 & 0.623 & 0.908 & 0.795 \\
    Limited \citep{kynkaanniemi2024applying} & 1.87 & 0.760 & 0.598 & 0.845 & 0.777 & 1.97 & 0.801 & 0.632 & & & 8.58 & 0.680 & 0.553 & 1.258 & 0.857 & 5.00 & 0.609 & 0.602 & 0.915 & \textbf{0.808} \\
    Cosine \citep{gao2023masked} & 2.15 & 0.770 & 0.619 & 0.850 & 0.769 & 2.30 & \textbf{0.861} & 0.520 & & & 8.29 & 0.659 & 0.564 & 1.106 & 0.840 & 5.14 & 0.630 & 0.616 & 0.920 & 0.802 \\
    CADS \citep{sadat2023cads} & \underline{1.60} & \textbf{0.792} & 0.619 & 0.854 & 0.765 & \underline{1.70} & 0.772 & 0.627 & & & 8.32 & \textbf{0.692} & 0.559 & 1.222 & 0.860 & 4.91 & \underline{0.633} & 0.613 & \underline{0.923} & 0.779 \\
    APG \citep{sadat2024eliminating} & 2.13 & 0.756 & \textbf{0.640} & 0.845 & 0.760 & 2.11 & 0.815 & 0.628 & & & 8.49 & 0.661 & \underline{0.571} & 1.095 & 0.858 & 5.23 & 0.614 & \textbf{0.631} & 0.915 & 0.797 \\
    REG \citep{xia2024rectified} & 1.99 & 0.761 & 0.608 & & & 1.76 & 0.799 & 0.601 & & & \underline{8.10} & 0.673 & 0.540 & 1.091 & 0.855 & 5.06 & 0.619 & 0.619 & 0.903 & 0.783 \\
    CFG++ \citep{chung2024cfg++} & N/A & N/A & N/A & N/A & N/A & N/A & N/A & N/A & N/A & N/A & 8.35 & 0.668 & 0.552 & 1.265 & 0.859 & 4.85 & 0.632 & 0.629 & 0.919 & 0.784 \\
    \midrule
    Power-law CFG (Ours) & 1.93 ($\textcolor{green}{\downarrow}$) & \underline{0.780} ($\textcolor{green}{\uparrow}$) & \underline{0.631} ($\textcolor{green}{\uparrow}$) & 0.845 ($\textcolor{red}{\downarrow}$) & 0.760 ($\textcolor{green}{\uparrow}$) & 2.05 ($\textcolor{green}{\downarrow}$) & 0.831 ($\textcolor{green}{\uparrow}$) & 0.595 ($\textcolor{green}{\uparrow}$) & & & 8.11 ($\textcolor{green}{\downarrow}$) & 0.670 ($\textcolor{green}{\uparrow}$) & 0.553 ($\textcolor{red}{\downarrow}$) & \textbf{1.310} ($\textcolor{green}{\uparrow}$) & 0.850 ($\textcolor{green}{\uparrow}$) & \underline{4.81} ($\textcolor{green}{\downarrow}$) & 0.621 ($\textcolor{red}{\downarrow}$) & 0.619 ($\textcolor{green}{\uparrow}$) & 0.918 ($\textcolor{green}{\uparrow}$) & 0.778 ($\textcolor{green}{\uparrow}$) \\
    Power-law CFG + Limited (Ours) & 1.73 ($\textcolor{green}{\downarrow}$) & 0.752 ($\textcolor{red}{\downarrow}$) & 0.600 ($\textcolor{green}{\uparrow}$) & 0.850 ($\textcolor{green}{\uparrow}$) & 0.778 ($\textcolor{green}{\uparrow}$) & 1.87 ($\textcolor{green}{\downarrow}$) & \underline{0.849} ($\textcolor{green}{\uparrow}$) & \textbf{0.642} ($\textcolor{green}{\uparrow}$) & & & 8.27 ($\textcolor{green}{\downarrow}$) & \textbf{0.692} ($\textcolor{green}{\uparrow}$) & 0.555 ($\textcolor{green}{\uparrow}$) & \underline{1.286} ($\textcolor{green}{\uparrow}$) & \underline{0.860} ($\textcolor{green}{\uparrow}$) & 4.84 ($\textcolor{green}{\downarrow}$) & 0.615 ($\textcolor{green}{\uparrow}$) & 0.622 ($\textcolor{green}{\uparrow}$) & 0.920 ($\textcolor{green}{\uparrow}$) & 0.795 ($\textcolor{red}{\downarrow}$) \\
    Power-law CFG + CADS (Ours) & \textbf{1.52} ($\textcolor{green}{\downarrow}$) & 0.770 ($\textcolor{red}{\downarrow}$) & 0.622 ($\textcolor{green}{\uparrow}$) & 0.862 ($\textcolor{green}{\uparrow}$) & 0.782 ($\textcolor{green}{\uparrow}$) & \textbf{1.63} ($\textcolor{green}{\downarrow}$) & 0.754 ($\textcolor{red}{\downarrow}$) & \underline{0.639} ($\textcolor{green}{\uparrow}$) & & & \textbf{7.98} ($\textcolor{green}{\downarrow}$) & \underline{0.690} ($\textcolor{red}{\downarrow}$) & \textbf{0.573} ($\textcolor{green}{\uparrow}$) & 1.279 ($\textcolor{green}{\uparrow}$) & \textbf{0.862} ($\textcolor{green}{\uparrow}$) & \textbf{4.71} ($\textcolor{green}{\downarrow}$) & \textbf{0.640} ($\textcolor{green}{\uparrow}$) & \underline{0.624} ($\textcolor{red}{\downarrow}$) & \textbf{0.924} ($\textcolor{green}{\uparrow}$) & \underline{0.804} ($\textcolor{green}{\uparrow}$) \\
    \bottomrule
  \end{tabular}}
  \label{tab:1}
  \vspace{-0.4cm}
\end{table}

\begin{figure}[t]
        \centering
        \begin{minipage}{\columnwidth}
            \includegraphics[width=\textwidth]{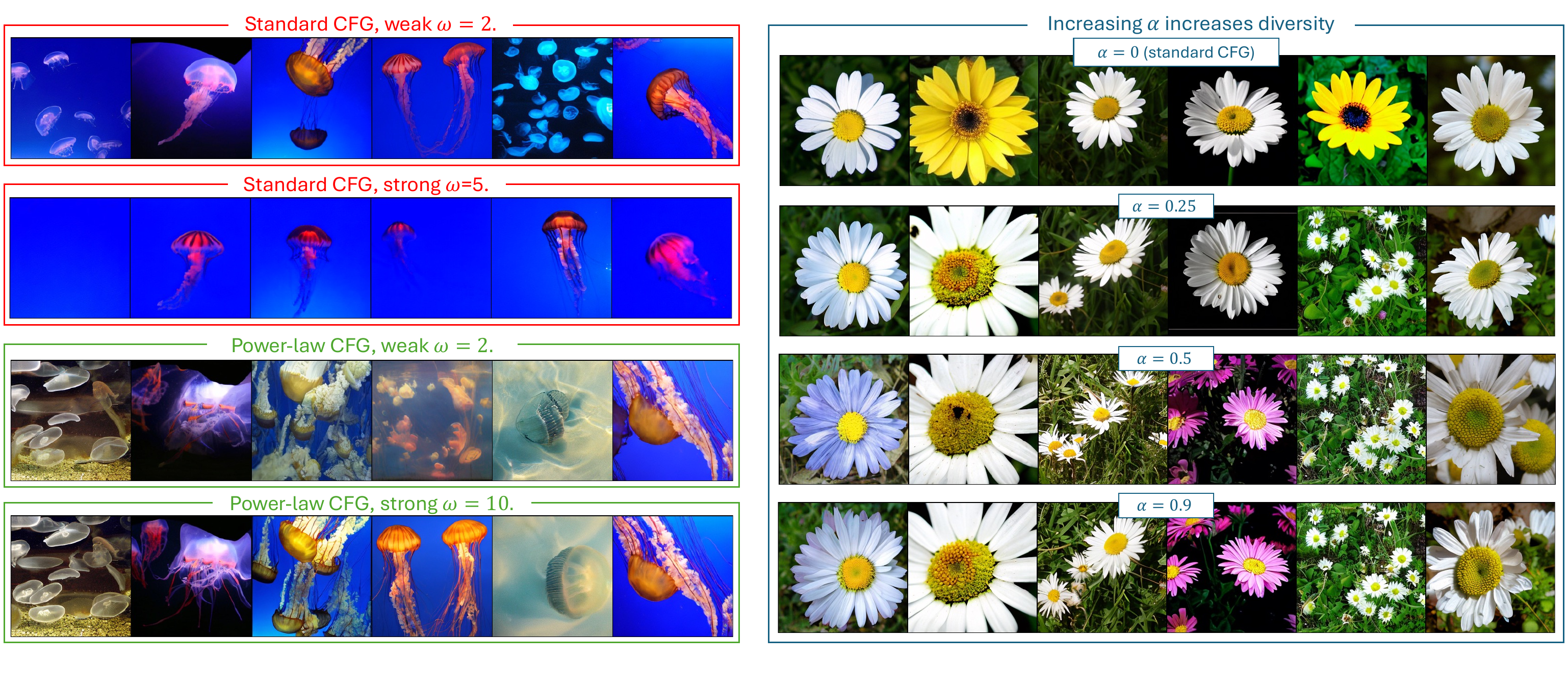}
        \end{minipage}        
        \vspace{-0.15cm}
        \caption{\small\textbf{Qualitative comparison of Standard and Power-Law CFG} on DiT/XL-2 trained on ImageNet-1K (256$\times$256). \textbf{Left:} while standard CFG results in diversity decrease or mode collapse (first image for $\omega=5.$), power-law CFG ($\alpha=0.9$) improves in diversity at no cost to fidelity, showing robustness to varying of $\omega$ (note very large $\omega=10.$). \textbf{Right:} Increasing non-linear parameter $\alpha$ yields larger diversity, while preserving image quality. Experimental details with further examples (as well as text-to-image) are provided in App. \ref{sec:appx_F_real_world}.
        }
        \vspace{-0.4cm}
        \label{fig:increasing_omega}
\end{figure}

\paragraph{Power-law CFG improves image quality and diversity.} 
We quantitatively evaluate our method using FID \citep{heusel2017gans} measuring image quality, and precision and recall \citep{sajjadi2018assessing} measuring diversity. 
In  \Cref{tab:1}, we compare power-law CFG to standard CFG and recent state-of-the-art guidance methods. 
As power-law guidance is easily combined with other guidance approaches, we also include results where we combine it with CADS~\citep{sadat2023cads} and limited-guidance~\citep{kynkaanniemi2024applying}, which we found to be the strongest competitors.
Power-law CFG improves over standard CFG in most cases (see arrows in table), and similarly it improves results of CADS and limited-interval guidance. 
Moreover, the latter combinations lead to results improving over existing approaches in many cases. 
We provide qualitative results in \Cref{fig:increasing_omega}, observing that power-law CFG improves both quality and diversity, while again being more robust to changing $\omega$. 

We provide additional  qualitative examples in \ref{sec:appx_F_real_world} for class-conditional and text-to-image models, as well as extend the  quantitative results using additional metrics.

%While Power-Law CFG demonstrates consistent benefits, its performance relative to other possible non-linear guidance strategies remains unexplored. There may be more effective and advantageous strategies, which are worth exploring in future work. 

\section{Conclusion}
\label{sec:conclusion}

\looseness=-1We studied the theoretical foundations of CFG, extending previous results to high and infinite-dimensional settings. Our research revealed that in sufficiently high dimension, CFG is in fact able to reproduce the correct target distribution, yielding a "blessing-of-dimensionality" result. 
Building on our theoretical analysis, we placed CFG in a larger family of guidance strategies, proposing a simple non-linear CFG extension. We confirmed its effectiveness through numerical and real-world experiments, applying it successfully to state-of-the-art text-to-image and class-conditional models. Our results demonstrate its consistent ability to improve sample quality and diversity.

\paragraph{Limitations and future work.} Our theory demonstrates that in high-dimensional settings, CFG generates the correct target distribution, extending  previous results showing CFG alters it in low-dimensions. 
In practice, CFG improves fidelity while reducing diversity: although our theory allows discovery of guidances that maintain strong fidelity while significantly boosting diversity, the reason why CFG-modified distribution is more effective in practice is not explained by our theory which relies on perfect score estimation. 
We hypothesize, therefore, that the practical benefits of (non-linear) CFG might be tied to the  imperfect score estimators used in practice. 
Investigating how score approximation errors impact guidance effectiveness is an important area for future research. 
Another key area of future study includes designing new non-linear CFG approaches.

\ifarxiv
\else
\clearpage
\newpage
\fi

\ifarxiv
\section*{Acknowledgements}
This work has received funding from the French government, managed by the National Research Agency (ANR), under the France 2030 program with the reference ANR-23-IACL-0008. Furthermore, this paper is supported by PNRR-PE-AI FAIR project funded by the NextGeneration EU program. We would like to thank Mathurin Videau, João Maria Janeiro, Kunhao Zheng, Tariq Berrada Ifriqi, Wes Bouaziz and Tony Bonnaire for fruitful discussions regarding the numerical experiments. We would further like to thank Levent Sagun, David Lopez-Paz, Brian Karrer, Ricky Chen, Arnaud Doucet, Yaron Lipman and Luke Zettlemoyer for feedback and support. Finally, we also thank Carolyn Krol and Carolina Braga for extensive consultation and support throughout this project.
\clearpage
\newpage
\else
\fi

\ifarxiv
    \bibliographystyle{assets/plainnat}
    \bibliography{paper}
    \clearpage
    \newpage
    \beginappendix
    \vspace{0.1cm}
\else
    \bibliography{paper}
    \bibliographystyle{assets/plainnat}
    
\newpage
\section*{NeurIPS Paper Checklist}

\begin{enumerate}

\item {\bf Claims}
    \item[] Question: Do the main claims made in the abstract and introduction accurately reflect the paper's contributions and scope?
    \item[] Answer: \answerYes{} %, \answerNo{}, or \answerNA{}.
    \item[] Justification: The abstract and introduction state our theoretical, methodological and experimental contributions, which are presented in detail in sections 3, 4 and 5.

\item {\bf Limitations}
    \item[] Question: Does the paper discuss the limitations of the work performed by the authors?
    \item[] Answer: \answerYes{} %, \answerNo{}, or \answerNA{}.
    \item[] Justification: We discuss the limitations of our work and directions for further investigation in a separate paragraph in \Cref{sec:conclusion} of the paper.

\item {\bf Theory assumptions and proofs}
    \item[] Question: For each theoretical result, does the paper provide the full set of assumptions and a complete (and correct) proof?
    \item[] Answer: \answerYes{} % Replace by \answerYes{}, \answerNo{}, or \answerNA{}.
    \item[] Justification: In the main manuscript, we specifically state the assumptions (e.g., well-separated means of the two Gaussians and perfect estimation of the score). The related material and prerequisite knowledge, full proofs and guidelines on how to extend the proofs to more general settings are provided in the supplementary material. 

    \item {\bf Experimental result reproducibility}
    \item[] Question: Does the paper fully disclose all the information needed to reproduce the main experimental results of the paper to the extent that it affects the main claims and/or conclusions of the paper (regardless of whether the code and data are provided or not)?
    \item[] Answer: \answerYes{} %, \answerNo{}, or \answerNA{}.
    \item[] Justification: 
We provide details on the Gaussian mixture simulations in \Cref{sec:appx_E_gm}, and details for the experiments with generative image models are provided in  \Cref{sec:appx_F_real_world}.


\item {\bf Open access to data and code}
    \item[] Question: Does the paper provide open access to the data and code, with sufficient instructions to faithfully reproduce the main experimental results, as described in supplemental material?
    \item[] Answer:  \answerYes{}  %, \answerNo{}, or \answerNA{}.
    \item[] Justification: We conducted experiments with open source pre-trained models, as well as one proprietary model that was trained on a mix of public and proprietary data. 
    We provide  details on the used models and data  in \Cref{sec:nonlin}.

\item {\bf Experimental setting/details}
    \item[] Question: Does the paper specify all the training and test details (e.g., data splits, hyperparameters, how they were chosen, type of optimizer, etc.) necessary to understand the results?
    \item[] Answer:  \answerYes{} %, \answerNo{}, or \answerNA{}.
    \item[] Justification: Our experiments do not involve training any models, but instead focus on the inference stage where we sample pre-trained generative image models. We speficy all the hyperaparameters used; moreover, the ones that are changed compared to the original models/codebases are specifically spelled out.
    
model size in Mparams and Gflops.
\item {\bf Experiment statistical significance}
    \item[] Question: Does the paper report error bars suitably and correctly defined or other appropriate information about the statistical significance of the experiments?
    \item[] Answer: \answerNo{} % Replace by \answerYes{}, \answerNo{}, or \answerNA{}.
    \item[] Justification: We do not provide error bars: we run each experiment once as the standard procedure in the field. Many benchmark results we were able to obtain from the original papers and we decided to avoid re-running them as it would require a significantly larger amount of compute resources.
    \item[] Guidelines:
    \begin{itemize}
        \item The answer NA means that the paper does not include experiments.
        \item The authors should answer "Yes" if the results are accompanied by error bars, confidence intervals, or statistical significance tests, at least for the experiments that support the main claims of the paper.
        \item The factors of variability that the error bars are capturing should be clearly stated (for example, train/test split, initialization, random drawing of some parameter, or overall run with given experimental conditions).
        \item The method for calculating the error bars should be explained (closed form formula, call to a library function, bootstrap, etc.)
        \item The assumptions made should be given (e.g., Normally distributed errors).
        \item It should be clear whether the error bar is the standard deviation or the standard error of the mean.
        \item It is OK to report 1-sigma error bars, but one should state it. The authors should preferably report a 2-sigma error bar than state that they have a 96\% CI, if the hypothesis of Normality of errors is not verified.
        \item For asymmetric distributions, the authors should be careful not to show in tables or figures symmetric error bars that would yield results that are out of range (e.g. negative error rates).
        \item If error bars are reported in tables or plots, The authors should explain in the text how they were calculated and reference tmodel size in Mparams and Gflops.he corresponding figures or tables in the text.
    \end{itemize}

\item {\bf Experiments compute resources}
    \item[] Question: For each experiment, does the paper provide sufficient information on the computer resources (type of compute workers, memory, time of execution) needed to reproduce the experiments?
    \item[] Answer: \answerYes{} % Replace by \answerYes{}, \answerNo{}, or \answerNA{}.
    \item[] Justification: In \ref{sec:appx_hyperparam_configs}, we give the hyperparameters over which the search has been conducted, as well as the type of GPUs used, together with the model size in Mparams and Gflops. All experiments performed are reported in the paper. Furthermore, our method proposes an alternative sampling algorithm for diffusion and flow-matching: during sampling, our method has exactly the same amount of NFEs to standard CFG: the main difference in the compute is when performing the hyperparameter search over $\alpha$, which we have shown above to yield consistently good results when selected around the value of $0.9$. Hence, the computational cost is not significantly different from standard CFG.
    
\item {\bf Code of ethics}
    \item[] Question: Does the research conducted in the paper conform, in every respect, with the NeurIPS Code of Ethics \url{https://neurips.cc/public/EthicsGuidelines}?
    \item[] Answer: \answerYes{} % Replace by \answerYes{}, \answerNo{}, or \answerNA{}.
    \item[] Justification: We do not involve human subjects or participants, nor we have data-related concerns: we have blurred faces in ImageNet-1K and CC12M. Potential impacts have, as mentioned below, been discussed in \Cref{sec:appx_H_impact}. 

\item {\bf Broader impacts}
    \item[] Question: Does the paper discuss both potential positive societal impacts and negative societal impacts of the work performed?
    \item[] Answer: \answerYes{}  %, \answerNo{}, or \answerNA{}.
    \item[] Justification: We provide a discussion of the broader impacts of our work in \Cref{sec:appx_H_impact}.
    
\item {\bf Safeguards}
    \item[] Question: Does the paper describe safeguards that have been put in place for responsible release of data or models that have a high risk for misuse (e.g., pretrained language models, image generators, or scraped datasets)?
    \item[] Answer:  \answerNA{}.
    \item[] Justification: We do not release any datasets or models. 

\item {\bf Licenses for existing assets}
    \item[] Question: Are the creators or original owners of assets (e.g., code, data, models), used in the paper, properly credited and are the license and terms of use explicitly mentioned and properly respected?
    \item[] Answer:  \answerYes{} %, \answerNo{}, or \answerNA{}.
    \item[] Justification: For all datasets and models used in our work we cite the corresponding papers in the main text, and provide links and license types in \Cref{tab:assets}.

\item {\bf New assets}
    \item[] Question: Are new assets introduced in the paper well documented and is the documentation provided alongside the assets?
    \item[] Answer:  \answerNA{}.
    \item[] Justification: We do not introduce any new assets in the paper. 

\item {\bf Crowdsourcing and research with human subjects}
    \item[] Question: For crowdsourcing experiments and research with human subjects, does the paper include the full text of instructions given to participants and screenshots, if applicable, as well as details about compensation (if any)? 
    \item[] Answer:   \answerNA{}.
    \item[] Justification: We do not report any experiments involving human subjects. 

\item {\bf Institutional review board (IRB) approvals or equivalent for research with human subjects}
    \item[] Question: Does the paper describe potential risks incurred by study participants, whether such risks were disclosed to the subjects, and whether Institutional Review Board (IRB) approvals (or an equivalent approval/review based on the requirements of your country or institution) were obtained?
    \item[] Answer:  \answerNA{}.
    \item[] Justification:  We do not report any experiments involving human subjects.

\item {\bf Declaration of LLM usage}
    \item[] Question: Does the paper describe the usage of LLMs if it is an important, original, or non-standard component of the core methods in this research? Note that if the LLM is used only for writing, editing, or formatting purposes and does not impact the core methodology, scientific rigorousness, or originality of the research, declaration is not required.
    \item[] Answer:  \answerNA{}.
    \item[] Justification: We do not use LLMs as a core component of our method. 

\end{enumerate}

    \newpage
    \appendix
    \onecolumn
    \section*{Supplementary Material}
\fi

The supplementary material is structured as follows:
\begin{itemize}
    \item In Section \ref{sec:appx_A_related_work}, we give a brief introduction to related work, focusing on \citet{biroli2024dynamical}.
    \item In Section \ref{sec:appx_B_basic_proof}, we give proofs for two equidistant, symmetric Gaussian mixtures.
    \item In Section \ref{sec:appx_C_extended_proof}, we present arguments how to extend the proofs to non-centered Gaussian mixtures (subsec. \ref{sec:appx_C1_extended_proof}) and multiple Gaussian mixtures (subsec. \ref{sec:appx_C2_extended_proof}).
    \item In Section \ref{sec:appx_D_fin_dim}, we present the theoretical and numerical findings for finite dimension (including low dimension $d$).
    \item In Section \ref{sec:appx_E_gm}, we present experimental details for Gaussian mixture numerical simulations.
    \item In Section \ref{sec:appx_F_real_world}, we provide experimental details involving real-world experiments.
    \item In Section \ref{sec:appx_G_nonlin_cfgs}, we propose another non-linear CFG alternative and provide num. experiments.
    \item In Section \ref{sec:appx_H_impact}, we discuss the broader societal impact of our work.
\end{itemize}

\section{Introduction to related work: Classifier-free Guidance (CFG) and Specification Time in the High-Dimensional Limit}
\label{sec:appx_A_related_work}
We start by briefly introducing the calculation required for estimating the speciation time $t_\textrm{s}$ for a case of two equally weighted Gaussians. This section is a direct adaptation of the framework introduced by \citet{biroli2024dynamical}. The diffusion process, consisting of $d$ independent Ornstein-Uhlenbeck Langevin equations, reads as follows (using $f(t)=-1$ and $g(t)=\sqrt{2}$ in \Cref{eqn:original_ou_process}):

\begin{align}
    d\vec{x}(t)=-\vec{x}dt+d\vec{B}(t),
    \label{eqn:ou}
\end{align}

where $d\vec{B}(t)$ equals the square root of two times the standard Brownian motion in $\mathbb{R}^d$. At time $t=0$, the process starts from the probability distribution $P_0(\vec{a})$, consisting of two Gaussian clusters that have means at $\pm \vec{m}$ and share the same variance $\sigma^2$. To guarantee that these Gaussians remain distinct in high-dimensional space, we assume that $|\vec{m}|^2 = d \tilde{\mu}^2$, where both $\sigma$ and $\tilde{\mu}$ are of order 1.

As the process evolves, the emergence of speciation resembles symmetry breaking observed during thermodynamic phase transitions. A common approach to analyzing this phenomenon is to construct a perturbative expansion of the free energy as a function of the field. Therefore, \citet{biroli2024dynamical} derive an expression for $\log P_t(\vec{x})$ using a perturbative expansion in terms of $e^{-t}$, which is valid for large time values. This method is justified since speciation occurs at large times.

One can rewrite the probability to be at $\vec{x}$ at time $t$ as
\begin{align*}
P_t(\vec{x}) & =\int d \vec{a} P_0(\vec{a}) \frac{1}{\sqrt{2 \pi \Delta_t^d}} \exp \left(-\frac{1}{2} \frac{\left(\vec{x}-\vec{a} e^{-t}\right)^2}{\Delta_t}\right) \\
& =\frac{1}{\sqrt{2 \pi \Delta_t}} \exp \left(-\frac{1}{2} \frac{\vec{x}^2}{\Delta_t}+g(\vec{x})\right),
\end{align*}

where the function $g(\vec{x})$, defined as
\begin{align*}
    g(\vec{x})=\log \int d \vec{a} P_0(\vec{a}) \exp \left(-\frac{1}{2} \frac{\vec{a}^2 e^{-2 t}}{\Delta_t}\right) \exp \left(\frac{e^{-t} \vec{x} \cdot \vec{a}}{\Delta_t}\right)
\end{align*}

can be viewed through a field-theoretic (or equivalently, a probabilistic) approach, where it serves as a generative function for connected correlations among the variables $\vec{a}$ \citep{zinn2021quantum}. By expanding this function at large times, one can show:
\begin{align*}
    g(\vec{x})=\frac{e^{-t}}{\Delta_t} \sum_{i=1}^d x_i\left\langle a_i\right\rangle+\frac{1}{2} \frac{e^{-2 t}}{\Delta_t^2} \sum_{i, j=1}^d x_i x_j\left[\left\langle a_i a_j\right\rangle-\left\langle a_i\right\rangle\left\langle a_j\right\rangle\right]+O\left(\left(x e^{-t}\right)^3\right),
\end{align*}

where we utilize the brackets $\langle\cdot\rangle$ to denote the expectation value with respect to the effective distribution $P_0(\vec{a}) e^{-\vec{a}^2 e^{-2 t} /\left(2 \Delta_t\right)}$. Therefore, the expansion can be used to show that at large times:

\begin{align*}
    \log P_t(\vec{x})=C+\frac{e^{-t}}{\Delta_t} \sum_{i=1}^d x_i\left\langle a_i\right\rangle-\frac{1}{2 \Delta_t} \sum_{i, j=1}^d x_i M_{i j} x_j+O\left(\left(x e^{-t}\right)^3\right),
\end{align*}

where $C$ is an $\vec{x}$-independent term and
\begin{align*}
    M_{i j}=\delta_{i j}-e^{-2 t}\left[\left\langle a_i a_j\right\rangle-\left\langle a_i\right\rangle\left\langle a_j\right\rangle\right].
\end{align*}

The curvature of $\log P_t(\vec{x})$ is closely linked to the spectral properties of the matrix $M$. In the large time regime, $M$ approaches the identity matrix, and consequently, all its eigenvalues are positive. However, a qualitative shift in shape occurs at the maximum time $t_\textrm{s}$, where the largest eigenvalue of $M$ transitions through zero. This marks the onset of the \textbf{speciation time}, distinguished by a change in curvature of the effective potential $-\log P_t(\vec{x})$. In this case, it can be easily computed: the matrix $M$ is given by $M_{i j}=\left(1-\sigma^2 e^{-2 t}\right) \delta_{i j}-e^{-2 t} m_i m_j$ and its largest eigenvalue is ( $1-\sigma^2 e^{-2 t}-d \tilde{\mu}^2 e^{-2 t}$ ). We get therefore in the large $d$ limit $t_\textrm{s}=\frac{1}{2} \log \left(d \tilde{\mu}^2\right)$ which up to subleading corrections identifies the speciation timescale as

\begin{align*}
    t_\textrm{s}=\frac{1}{2} \log (d).
\end{align*}

\section{Theoretical proofs: two equidistant, symmetric Gaussian mixtures}
\label{sec:appx_B_basic_proof}

\subsection*{Asymptotic stochastic process in Regime \rom{1} and symmetry breaking}

In the limit of large dimensions, a comprehensive analytical examination of the dynamics in Regime \rom{1}, taking place on time-scales  $t_\textrm{s}+O(1)=(1/2)\log d +O(1)$, can be provided, specifically at the beginning of the backward process. Assuming no collapse (for further details, refer to \citet{biroli2024dynamical}), an investigation into diffusion dynamics shows that the empirical distribution $P_t^e(\vec{x})$ at time $t$ can be approximated with high accuracy by $P_t(\vec{x})$. This approximation represents the convolution of the initial distribution $P_0$, comprising a mixture of Gaussians centered at $\pm \vec{m}$, and a diffusion kernel proportional to $e^{-\left(\vec{x}-\vec{a} e^{-t}\right)^2 / 2}$. Consequently, the explicit expression for this approximation is

\begin{align}
    P_0(\vec{x})=\frac{1}{2 \left(\sqrt{2 \pi \sigma^2}\right)^d}\left[e^{-\left(\vec{x}-\vec{m} \right)^2 /\left(2 \sigma^2\right)}+e^{-\left(\vec{x}+\vec{m} \right)^2 /\left(2 \sigma^2\right)}\right] \text{, and}
\label{eqn:gm_0}
\end{align}

\begin{align*}
    P_t(\vec{x})=\frac{1}{2 \left(\sqrt{2 \pi \Gamma_t}\right)^d}\left[e^{-\left(\vec{x}-\vec{m} e^{-t}\right)^2 /\left(2 \Gamma_t\right)}+e^{-\left(\vec{x}+\vec{m} e^{-t}\right)^2 /\left(2 \Gamma_t\right)}\right]
\end{align*}

where $\Gamma_t=\sigma^2 e^{-2 t}+\Delta_t$ goes to 1 at large times. The log of this probability is

\begin{align*}
    \log P_t(\vec{x})=-\frac{\vec{x}^2}{2 \Gamma_t}+\log \cosh \left(\vec{x} \cdot \vec{m} \frac{e^{-t}}{\Gamma_t}\right),
\end{align*}

and hence the score reads

\begin{align}
    S_t^i(\vec{x})=-\frac{x^i}{\Gamma_t}+m_i \frac{e^{-t}}{\Gamma_t} \tanh \left(\vec{x} \cdot \vec{m} \frac{e^{-t}}{\Gamma_t}\right).
    \label{eqn:uncond_score}
\end{align}

As there are two classes: $+\vec{m}$ and  $-\vec{m}$, the score conditioned to one class equals the score associated to a given Gaussian. Therefore, for the two classes we have:

\begin{align}
    \begin{split}
        &+\vec{m}: S_t^i(\vec{x}, +)=\frac{-x^i+m_i e^{-t}}{\Gamma_t}, \text{and} \\
        &-\vec{m}: S_t^i(\vec{x}, -)=\frac{-x^i-m_i e^{-t}}{\Gamma_t}. 
    \end{split}
    \label{eqn:cond_score}
\end{align}

\subsection{Result \rom{1}: What is the role of classifier-free guidance?}

Let us first analyze the ``transverse'' directions $\vec{v}\perp\vec{m}$. For these directions, for all $\omega$, the score is the same and equals $\vec{S}_t^{\text{CFG}} (\vec{x}, c) \cdot \vec{v} = -\frac{\vec{x}\cdot\vec{v}}{\Gamma_t}$. 
Let us project the backward \Cref{eqn:backw_dyn} on a unit vector $\vec v\perp\vec{m}$. 
We write $p=\vec x\cdot\vec v$, and the backward equation now reads $dp=p (1-2/\Gamma_{t_f-\tau})d\tau +\sqrt{2} dB $
which is the backward equation for a single Gaussian variable.  When $\tau\to t_f$ the projection $p=\vec x\cdot\vec v$ is thus distributed as $\mathcal{N}(0, \sigma^2)$, for all values of $\omega$.

Therefore, as all the components except the one in the $\vec{m}$ direction are not affected, we can consider only the component along $\vec{m}$:

\begin{align*}
    \vec{S}_{t_{CFG}}(\vec{x},c) \cdot \frac{\vec{m}}{\abs{\vec{m}}} = - \frac{\vec{x}\cdot \vec{m}/\abs{\vec{m}}}{\Gamma_t} + \omega \frac{\abs{\vec{m}}^2 e^{-t}}{\abs{\vec{m}} \Gamma_t} \cdot \left\{c-\tanh{\left(\frac{\vec{x}\cdot\vec{m} e^{-t}}{\Gamma_t}\right)}\right\} +\frac{\abs{\vec{m}} e^{-t} c}{\Gamma_t}.
\end{align*}

By denoting $\frac{\vec{x}\cdot\vec{m}}{\abs{\vec{m}}}=q(t)$, where $\abs{\vec{m}}=\sqrt{d}$, we can obtain the backward equation:

\begin{align*}
    d{x^i} = (x^i + 2 S_{\tau_{CFG}}^i)d\tau + d\eta_i(\tau),
\end{align*}

where $\tau=t_{f}-t$, i.e., the backward time. Therefore, we can obtain for Regime \rom{1} and by projecting onto the $\frac{\vec{m}}{\abs{\vec{m}}}$ direction, we have that:

\begin{align*}
    d q = dx^i\cdot\frac{\vec{m}}{\abs{\vec{m}}} = \Big(q + 2\Big[-q+ e^{-(t_f-t_\textrm{s}-\tau)} \Big((1+\omega) - \omega\tanh{\left(qe^{-(t_f-t_\textrm{s}-\tau)}\right)}\Big)\Big]\Big)d\tau+d\eta(\tau),
\end{align*}

as, in Regime \rom{1}, we have that $\Gamma_t\approx1$, and also $\sqrt{d}=e^{-t_\text{\textrm{s}}}$.

Again, from this point onward by $t(\tau)$ we denote the backward time for ease of notation. This is like having an effective potential:

\begin{align*}
    d q = -\pdv{V^\textrm{CFG}(q,\tau)}{q}d\tau+d\eta(\tau),
\end{align*}

where

\begin{align*}
    V^\textrm{CFG} &= \frac1{2}q^2 +2\left[-(1+\omega)cqe^{-(t-t_s)}+\omega \ln \cosh{\left(qe^{-(t-t_s)}\right)}\right] \notag \\
    &= \underbrace{(\frac1{2}q^2-2e^{-(t-t_s)}cq)}_\text{Classifier's potential}+\omega\underbrace{\left[-cqe^{-(t-t_s)}+\ln\cosh{\left(qe^{-(t-t_s)}\right)}\right]}_\text{Extra potential $V_{\text{extra}}$}.
\end{align*}

\begin{figure}[t]
        \centering
        \includegraphics[width=0.7\textwidth]{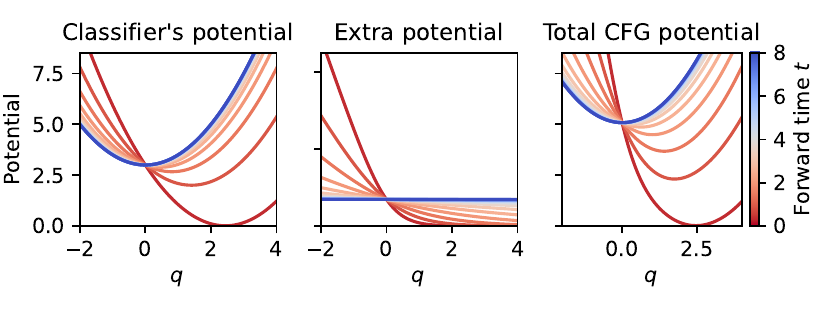}
        \vspace{-.5cm}
        \caption{\textbf{Effect of CFG on the guiding potential of a Gaussian mixture.} The backward diffusion for the variable $q$ giving the projection of $\vec x$ on the center $\vec m$ of the Gaussian where one wants to guide the backward diffusion. From left to right: Potential within the class, CFG-added-potential $V_\text{extra}$ with $\omega=2$, and their sum as in \Cref{eqn:eff_pot}. CFG exhibits faster convergence to the target ($t=0$), but results in narrower potential for small $t$ (with $t$ ranging from 0 to 8, as indicated on the right panel).
        } 
        \label{fig:4_potentials}
\end{figure}

Therefore, for class $c=+1$ (equivalently for $c=-1$), there is little effect for $qe^{-(t-t_s)}\gg1$, as then $-qe^{-(t-t_s)}+\ln\cosh{\left(qe^{-(t-t_s)}\right)} \approx 0$. Instead, for $qe^{-(t-t_s)}\ll-1$, we have that $-qe^{-(t-t_s)}+\ln\cosh{\left(qe^{-(t-t_s)}\right)}\approx-qe^{-(t-t_s)}\gg1$.  Therefore, we can conclude our first result:

\textbf{Result \rom{1}.} In Regime \rom{1}, before speciation time $t_\textrm{s}$, CFG is effective in aiding class selection and speeds up the convergence towards the target class $c$.

The utility of CFG is therefore to "push" in the right direction in Regime \rom{1} where arguably the class-based score/potential is likely not accurate in the rare region ($q>0$ for $c=-1$ and $q<0$ for $c=+1$). The behavior of the two potentials is displayed in Figure \ref{fig:4_potentials}.

\subsection{Result \rom{2}: Path alignment}

The role of CFG in Regime \rom{1} is to push the trajectories more in the direction of the selected class. We recall that the SDE verified by $q$ when pushed towards class $c=+1$ reads: 

\begin{equation}
    \begin{aligned}
        d q = 
        \Big(q + 2\Big[-q+ e^{-(t_f-t_\textrm{s}-\tau)} \Big((1+\omega) - \omega\tanh{\left(qe^{-(t_f-t_\textrm{s}-\tau)}\right)}\Big)\Big]\Big)d\tau+d\eta(\tau),
    \end{aligned}
\end{equation}

For large times but still during Regime \rom{1}, i.e. $t_f-t_\textrm{s} \ll \tau \ll \sqrt{d}$, $q$ is very large (positive or negative). In this regime the CFG term can be neglected as it leads to exponentially small corrections to the SDE (of order $e^{-2q e^{-(t_f-t_\textrm{s}-\tau)}}$) with $t_f-t_\textrm{s}-\tau \gg 1$.
In consequence, in Regime \rom{1} at large times, the SDE just reads:
\begin{align*}
    d q = -q+  2 e^{-(t_f-t_\textrm{s}-\tau)} +d\eta(\tau),   
\end{align*}
 
The effect of CFG is to lead to different values of $q$ when entering this late regime of Regime \rom{1}. We call these values $q(\tau_i)$ and denote $\tau_i$ the fixed time at which the CFG contribution can be neglected. The value $q(\tau_i)$  is quickly (exponentially) forgotten when $\tau$ departs from $\tau_i$, i.e., the evolution readjust to the correct value without CFG.
This can be shown by solving the SDE starting from a given $\tau_i$:
\begin{align*}
    q(\tau)=q(\tau_i)e^{-(\tau-\tau_i)}+e^{-(t_f-t_\textrm{s})}\left(e^{\tau}-e^{-\tau+2\tau_i} \right) +\sqrt{1-e^{-(2(\tau-\tau_i))}}z_{\tau}
\end{align*}

where $z_\tau$ is a Gaussian variable with mean zero and unit variance. 
When $\tau \gg \tau_i$ but still in Regime \rom{1} the solution of the SDE does not depend any longer on $q(\tau_i)$ and it coincides statistically with the one of the backward process of the single Gaussian corresponding to the class $c=+1$. This allows to conclude the second result:

\textbf{Result \rom{2}.} Just before speciation time $t_\textrm{s}$, CFG-guided paths realign with the unguided path that generates the correct, unmodified target distribution.

\subsection{Result \rom{3}: When does classifier-free guidance take effect?}

We can proceed to answer this question by examining the classifier-free guidance score, as defined in \citet{ho2022classifier}:

\begin{align}
S_{t_{CFG}}^i(\vec{x},c)= (1+\omega) S^i_t(\vec{x}, c) - \omega S_t^i(\vec{x}),
\label{appx:score_CFG_linear}
\end{align}

where $c=\pm1$ and $\omega>0$. By plugging in the cond. (\ref{eqn:cond_score}) and uncond. scores (\ref{eqn:uncond_score}), we can obtain:

\begin{align}
    S_{t_{CFG}}^i(\vec{x},c) &= -\frac{x^i}{\Gamma_t} + (1+\omega)\frac{c m_i e^{-t}}{\Gamma_t} - \omega \frac{m_i e^{-t}}{\Gamma_t} \tanh{\left(\frac{\vec{x} \cdot \vec{m} e^{-t}}{\Gamma_t}\right)}   \notag  \\
    &= -\frac{x^i}{\Gamma_t} + \omega\frac{m_i e^{-t}}{\Gamma_t}\left\{c-\tanh{\left(\frac{\vec{x}\cdot\vec{m}e^{-t}}{\Gamma_t}\right)}\right\} + \frac{cm_ie^{-t}}{\Gamma_t}.
    \label{eqn:cfg_score_appx_1}
\end{align}

Now, in Regime \rom{2}, when the trajectory has committed to a given class, $\vec{x}\cdot\vec{m}\sim O(d)$ and $ \sign(\vec{x}\cdot\vec{m})=c$. Therefore, $c-\tanh{\left(\frac{\vec{x}\cdot\vec{m}e^{-t}}{\Gamma_t}\right)}\approx0$, and one finds from (\ref{eqn:cfg_score_appx_1}), that  $S^i_{t_{CFG}} (\vec{x}, c)= S_t^i(\vec{x})$. This implies that, within this regime, classifier-free guidance equals the conditional score. Therefore, Classifier free-guidance only affects Regime \rom{1}, as $S^i_{t_{CFG}} (\vec{x}, c)= S_t^i(\vec{x})$ for $t>t_\textrm{s}=\frac1{2}\log(d)$. This allows us to conclude the third result:

\textbf{Result \rom{3}.} In Regime \rom{2}, after speciation time $t_\textrm{s}$, CFG has no effect on the generation process.

\section{Generalizations of the proof}
\label{sec:appx_C_extended_proof}

In this section, we present arguments for extending our proofs to more general cases. We start by discussing proof generalization for non-centered Gaussian mixtures (Section \ref{sec:appx_C1_extended_proof}) and then move on to a mixture of four Gaussians (\ref{sec:appx_C2_extended_proof}). Finally, we conclude with some remarks on how to further extend these results to more complex scenarios.

\subsection{Generalization to non-centered Gaussian mixtures}
\label{sec:appx_C1_extended_proof}
\subsubsection*{Asymptotic stochastic process in Regime \rom{1} and symmetry breaking}

Here we provide an example on how to generalize the study of Gaussian mixtures to the case where the two Gaussians are centered in $\vec m_1$ and $\vec m_2$. We take $\vec m_1, \vec m_2$ as two arbitrary vectors in $d$ dimensions, on the sphere $|\vec m_c|^2=d$ the case where they have different norms, both scaling proportionally to $d$, could be studied as well with the same formalism.

The initial probability density is

\begin{align}
    P_0(\vec{x})=\frac{1}{2 \left(\sqrt{2 \pi \sigma^2}\right)^d}\left[e^{-\left(\vec{x}-\vec{m}_1\right)^2 /\left(2 \sigma^2\right)}+e^{-\left(\vec{x}-\vec{m}_2 \right)^2 /\left(2 \sigma^2\right)}\right] \text{, and}
\label{eqn:gm_0_gen}
\end{align}

\begin{align*}
    P_t(\vec{x})=\frac{1}{2 \left(\sqrt{2 \pi \Gamma_t}\right)^d}\left[e^{-\left(\vec{x}-\vec{m}_1 e^{-t}\right)^2 /\left(2 \Gamma_t\right)}+e^{-\left(\vec{x}-\vec{m}_2 e^{-t}\right)^2 /\left(2 \Gamma_t\right)}\right]
\end{align*}

where $\Gamma_t=\sigma^2 e^{-2 t}+\Delta_t$ goes to 1 at large times. The log of this probability is

\begin{align*}
    \log P_t(\vec{x})=-\frac{\vec{x}^2}{2 \Gamma_t}+\log \left(
    e^{\vec{x} \cdot \vec{m}_1 \frac{e^{-t}}{\Gamma_t}}
    +e^{\vec{x} \cdot \vec{m}_2 \frac{e^{-t}}{\Gamma_t}}
    \right)+C,
\end{align*}
where $C$ is    a constant,
and hence the score reads

\begin{align}
    S_t^i(\vec{x})=-\frac{x^i}{\Gamma_t}+\frac{e^{-t}}{\Gamma_t}
    \frac{m_1^i\;   e^{\vec{x} \cdot \vec{m}_1 \frac{e^{-t}}{\Gamma_t}} + m_2^i \;  e^{\vec{x} \cdot \vec{m}_2 \frac{e^{-t}}{\Gamma_t}}}
    { e^{\vec{x} \cdot \vec{m}_1 \frac{e^{-t}}{\Gamma_t}} +  e^{\vec{x} \cdot \vec{m}_2 \frac{e^{-t}}{\Gamma_t}}}
    \label{eqn:uncond_score_gen}
\end{align}

As there are two classes: $\vec{m}_1$ and  $\vec{m}_2$, the score conditioned to one class equals the score associated to a given Gaussian. Therefore, for the two classes we have:

\begin{align}
    \begin{split}
        &\vec{m}_1: S_t^i(\vec{x}, +)=\frac{-x^i+m_1^i e^{-t}}{\Gamma_t}, \text{and} \\
        &\vec{m}_2: S_t^i(\vec{x}, -)=\frac{-x^i-m_2^i e^{-t}}{\Gamma_t}. 
    \end{split}
    \label{eqn:cond_score_gen}
\end{align}

\subsubsection*{What is the role of classifier-free guidance?}
We shall use as basis the vectors $\vec m_+=(\vec m_1+\vec m_2)/2$, 
$\vec m_-=(\vec m_1-\vec m_2)/2$, and  we shall denote by $\vec v$ the vectors orthogonal to the place generated by $\vec m_1,\vec m_2$. 

For these  ``transverse'' directions $\vec{v}\perp(\vec{m}_1,\vec{m}_2)$.  for all $\omega$, the score is the same and equals $\vec{S}_t^{\text{CFG}} (\vec{x}, c) \cdot \vec{v} = -\frac{\vec{x}\cdot\vec{v}}{\Gamma_t}$. 
Let us project the backward equation on a unit vector $\vec v$ in the transverse space.
We write $p=\vec x\cdot\vec v$, and the backward equation now reads $dp=p (1-2/\Gamma_{t_f-\tau})d\tau +\sqrt{2} dB $
which is the backward equation for a single Gaussian variable.  When $\tau\to t_f$ the projection $p=\vec x\cdot\vec v$ is thus distributed as $\mathcal{N}(0, \sigma^2)$, for all values of $\omega$.

Therefore, as all the components except the ones in the $\vec{m}_+$  and $\vec{m}_-$ directions are not affected.

We now project the score on $\vec m_+$ and $\vec m_-$, using $\vec m_+.\vec m_-=0$, $\vec m_+.\vec m_1=\vec m_+.\vec m_2=d^2/2 $ and 
$\vec m_-.\vec m_1=-\vec m_-.\vec m_2=d^2/2 $:

\begin{align*}
    \vec{S}_{t_{CFG}}(\vec{x},c) \cdot \vec{m}_+ &=  \frac{(\vec{m}_+ e^{-t}- \vec{x})\cdot \vec{m}_+}{\Gamma_t } \\ 
     \vec{S}_{t_{CFG}}(\vec{x},c) \cdot \vec{m}_-&=  \frac{(\vec{m}_- e^{-t}- \vec{x})\cdot \vec{m}_-}{\Gamma_t } 
      + 
     \omega \frac{\abs{\vec{m}_-}^2 e^{-t}}{ \Gamma_t} \cdot \left\{1-\tanh{\left(\frac{\vec{x}\cdot\vec{m}_- e^{-t}}{\Gamma_t}\right)}\right\} 
\end{align*}
Inserting these scores into the backward diffusion equation, one finds that:
\begin{itemize}
    \item $\vec x.\vec m_+/|\vec m_+|$ evolves as a Gaussian variable. At time $\tau\to t_f$ the distribution of this variable is $\mathcal{N}(|\vec m_+|,\sigma^2)$. 
    \item The variable $q_-(t) = \frac{\vec{x}\cdot\vec{m}_-} {|\vec{m}_-|}$ satisfies the same equation as the variable $q(t)$ which we analyzed in the 'centered' case where $\vec m_1=-\vec m_2=\vec m$
\end{itemize}

Therefore, we can conclude that in this case, CFG has the same effect: it is effective in aiding class selection, speeding up the convergence toward the correct target class $c$.

\subsubsection*{When does classifier-free guidance take effect?}

We can proceed to answer this question by examining the classifier-free guidance score:

\begin{align}
S_{t_{CFG}}^i(\vec{x},c)= (1+\omega) S^i_t(\vec{x}, c) - \omega S_t^i(\vec{x}),
\label{appx:score_CFG_linear_gen}
\end{align}
where $c\in\{1,2\}$ and $\omega>0$. The CFG score guiding to class $c=1$ is thus:
\begin{align}
    S_{t_{CFG}}^i(\vec{x},c) &= -\frac{x^i}{\Gamma_t} + (1+\omega)\frac{ m_1^i e^{-t}}{\Gamma_t} - \omega \frac{ e^{-t}}{\Gamma_t} 
    \frac{m_1^i\;   e^{\vec{x} \cdot \vec{m}_1 \frac{e^{-t}}{\Gamma_t}} + m_2^i \;  e^{\vec{x} \cdot \vec{m}_2 \frac{e^{-t}}{\Gamma_t}}}
    { e^{\vec{x} \cdot \vec{m}_1 \frac{e^{-t}}{\Gamma_t}} +  e^{\vec{x} \cdot \vec{m}_2 \frac{e^{-t}}{\Gamma_t}}}
    \label{eqn:cfg_score_appx_1_gen}
\end{align}

Now, in Regime \rom{2}, when the trajectory has committed to a given class say class $1*$, $\vec{x}\cdot\vec{m}_1-\vec{x}\cdot\vec{m}_2$ is positive and of order $  O(d)$. Therefore 
  $S^i_{t_{CFG}} (\vec{x}, c)= S_t^i(\vec{x},c)$. 
  %M NB There is a misprint in the mainpaper: Two lines after  (14) one should write "one finds from (14) that $S^i_{t_{CFG}} (\vec{x}, c)= S_t^i(\vec{x},c)$ " (now it is written incorrectly $S^i_{t_{CFG}} (\vec{x}, c)= S_t^i(\vec{x})$ " 
   This implies that, within this regime, classifier-free guidance equals the conditional score. Therefore, Classifier free-guidance only affects Regime \rom{1}, as $S^i_{t_{CFG}} (\vec{x}, c)= S_t^i(\vec{x})$ for $t>t_s=\frac1{2}\log(d)$. This allows us to conclude that in Regime \rom{2}, CFG is innocuous.

Therefore all the results obtained for the centered case $\vec m_1=-\vec m_2=\vec m$ also hold for the more general case when the two Gaussians are centered in $\vec m_1$ and $\vec m_2$.

\subsection{Extension to the mixture of four Gaussians}
\label{sec:appx_C2_extended_proof}

Here we present the computation for a mixture of four Gaussians, in order to analyze the behavior of the system for an increasing number of classes and emphasize the extendability of our framework. As before, assuming no collapse, we can approximate the empirical distribution $P_t^e(\vec{x})$ at time $t$ by $P_t(\vec{x})$ with high accuracy. In this case, the approximation represents the convolution of the initial distribution $P_0$, being a mixture of 4 Gaussians centered at $\pm\m1 \pm \m2$, s.t. $\m1\cdot\m2=0$, and a diffusion kernel proportional to $e^{-\left(\vec{x}-\vec{a} e^{-t}\right)^2 / 2}$. The explicit expression for the distribution is:

\begin{align*}
    P_0(\vec{x})=\frac{1}{4 \left(\sqrt{2 \pi \sigma^2}\right)^d}&\left[e^{-\left(\vec{x}-(\m1 - \m2) \right)^2 /\left(2 \sigma^2\right)}+e^{-\left(\vec{x}-(\m1 
    + \m2) \right)^2 /\left(2 \sigma^2\right)}\right.\\
    &\left.+e^{-\left(\vec{x}+(\m1 - \m2) \right)^2 /\left(2 \sigma^2\right)}+e^{-\left(\vec{x}+(\m1 + \m2) \right)^2 /\left(2 \sigma^2\right)}\right]
\end{align*}

and 
\begin{align*}
    P_t(\vec{x})=\frac{1}{4 \left(\sqrt{2 \pi \Gamma_t}\right)^d}&\left[e^{-\left(\vec{x}-(\m1 - \m2)e^{-t} \right)^2 /\left(2 \Gamma_t\right)}+e^{-\left(\vec{x}-(\m1 + \m2)e^{-t} \right)^2 /\left(2 \Gamma_t\right)}\right.\\
    &+ \left. e^{-\left(\vec{x}+(\m1 - \m2)e^{-t} \right)^2 /\left(2 \Gamma_t\right)}+e^{-\left(\vec{x}+(\m1 + \m2)e^{-t} \right)^2 /\left(2 \Gamma_t\right)}\right]
\end{align*}

where $\Gamma_t=\sigma^2 e^{-2 t}+\Delta_t$ goes to 1 at large times. This can be rewritten as:

\begin{align*}
    P_t(\vec{x})&= \frac{1}{2 \left(\sqrt{2 \pi \Gamma_t}\right)^d}e^{-(\vec{x}^2+\m1^2e^{-2t}+\m2^2e^{-2t})/(2\Gamma_t)}\left[e^{-\m1\cdot\m2e^{-2t}/\Gamma_t} \cosh (\vec{x}\cdot(\m1+\m2)\frac{e^{-t}}{\Gamma_t}) \right. \\
    & \left. + e^{\m1\cdot\m2e^{-2t}/\Gamma_t} \cosh (\vec{x}\cdot(\m1-\m2)\frac{e^{-t}}{\Gamma_t}) \right]
\end{align*}

The log of this probability is:

\begin{align*}
    \log P_t(\vec{x})=\frac{-\vec{x}^2}{2\Gamma_t}+\log\left(e^{-\m1\cdot\m2e^{-2t}/\Gamma_t} \cosh (\vec{x}\cdot(\m1+\m2)\frac{e^{-t}}{\Gamma_t}) + e^{\m1\cdot\m2e^{-2t}/\Gamma_t} \cosh (\vec{x}\cdot(\m1-\m2)\frac{e^{-t}}{\Gamma_t})\right)
\end{align*}

And the score reads:

\begin{align*}
    S^i_t(\vec{x})&=\frac{-x^i}{\Gamma_t}\\
    &+\frac{e^{-t}}{\Gamma_t}
    \frac
        {(\m1+\m2)_ie^{-\m1\cdot\m2e^{-2t}/\Gamma_t} \sinh (\vec{x}\cdot(\m1+\m2)\frac{e^{-t}}{\Gamma_t})+(\m1-\m2)_ie^{\m1\cdot\m2e^{-2t}/\Gamma_t} \sinh (\vec{x}\cdot(\m1-\m2)\frac{e^{-t}}{\Gamma_t})}
        {e^{-\m1\cdot\m2e^{-2t}/\Gamma_t} \cosh (\vec{x}\cdot(\m1+\m2)\frac{e^{-t}}{\Gamma_t}) + e^{\m1\cdot\m2e^{-2t}/\Gamma_t} \cosh (\vec{x}\cdot(\m1-\m2)\frac{e^{-t}}{\Gamma_t})}
\end{align*}

As $\vec{x}$ approaches one of the means $\pm \m1 \pm \m2$, the second summand reduces to $(\m1 \pm \m2) \tanh(x\cdot(\m1 \pm \m2)\frac{e^{-t}}{\Gamma_t})$\footnote{For large values of $x\cdot(\m1 \pm \m2)e^{-t}/\Gamma_t$, we utilized the \textit{log-sum-exp trick} to calculate the value of the fraction.}, resulting in an expression akin to the one for mixture of 2 Gaussians in (\ref{eqn:uncond_score}).

\paragraph{Conclusion.} The results above can be generalized to any finite number of Gaussians, centered around $\vec m_i$ where $\vec m_i$ is a vector of norm $\mu_i \sqrt{d}$. CFG will only have effect on space spanned by vectors $\vec m_i$ and only in regime \rom{1}. One can also consider non-isotropic Gaussians. As long as the covariance has eigenvalues not scaling with $d$, the backward process displays the two distinct regimes \rom{1} and \rom{2}, which is examined in detail for the mixture of two Gaussians. This result can be obtained by analyzing the forward process. The key point is that on all times of order one the noised Gaussian mixture still consists in non-overlapping Gaussian (regime\rom{2}). On times of order one close to the speciation time $1/2 \log d$ the Gaussians overlap and the center are of the same order of the noise (regime \rom{1}). 
Because of the existence of these two regimes, the general arguments presented at the beginning of the paper hold and CFG does reproduce the correct distribution in the large $d$ limit.

\section{Finite dimension}
\label{sec:appx_D_fin_dim}

In this section, we give exact analyses describing the effect of CFG in finite- (possibly low-) dimensional settings, outlined in Section \ref{sec:fin_dim_eff} in the main manuscript. We start the backward equation at a time $t_f$ large enough that the distribution of $x$ is a isotropic  Gaussian with variance one.
The backward equation for \( x(t) \) with the CFG score reads:

\begin{align}
    \frac{dx_i}{d\tau} =& x_i \left(1 - \frac{2}{\Gamma(t_f - \tau)}\right) + \frac{2m_i}{\Gamma(t_f - \tau)} e^{-(t_f - \tau)} 
    \nonumber\\
    &+2 \omega m_i\frac{e^{-(t_f-\tau)}}{\Gamma_{t_f-\tau}}\left\{1-\tanh\left(\frac{\vec x\cdot\vec m e^{-(t_f-\tau)}}{\Gamma_{t_f-\tau}}\right)\right\}+ 
    \eta_i(\tau)
\end{align}
where \( \tau = 0 \) at the beginning of the backward process and \( \tau = t_f (\gg 1) \) at the end.

This can be projected on the evolution of the single parameter $q(\tau)=\vec x\cdot\vec m /\sqrt{d}$.
We obtain
\begin{align}
    \frac{dq}{d\tau} =& q \left(1 - \frac{2}{\Gamma(t_f - \tau)}\right) + \frac{2\sqrt{d}}{\Gamma(t_f - \tau)} e^{-(t_f - \tau)} \nonumber\\
    &+2 \omega \sqrt{d}\frac{e^{-(t_f-\tau)}}{\Gamma_{t_f-\tau}}\left\{1-\tanh\left(\frac{q \sqrt{d} e^{-(t_f-\tau)}}{\Gamma_{t_f-\tau}}\right)\right\}+ 
    \eta(\tau).
    \label{basic_backward_eq_q}
\end{align}

Considering the right-hand side as a force due to a moving external potential $-\partial_q V(q,t)$, the effect of CFG is to add an extra term which has two main effects: (1) it adds a positive term to the force and, in consequence, it pushes $q$ faster away from zero, (2) it increases the value of the Hessian at any point in $q$ with respect to its $\omega=0$ counterpart, thus making the potential more confining.

The initial condition is \( q(\tau = 0) \sim \mathcal{N}(0, \sigma^2) \) and
\begin{equation}
    \Gamma(t_f - \tau) = \sigma^2 e^{-2(t_f - \tau)} + 1 - e^{-2(t_f - \tau)}.
\end{equation}

\subsection*{Case: $\omega=0$}
The solution of the backward equation is:

\begin{equation}\label{bsol}
    q(\tau) = q(0)e^{ \tau-2\int_0^\tau \frac{1}{\Gamma(t_f-\tau'')}d\tau''} + \int_{0}^{\tau} \left[\frac{2\sqrt{d} e^{-(t_f - \tau')}}{\Gamma(t_f-\tau')} + \eta_i(\tau')\right] e^{(\tau - \tau')-2\int_{\tau'}^\tau \frac{1}{\Gamma(t_f-\tau'')}d\tau''} d\tau'.
\end{equation}
Its probability distribution must coincide with the one of the solutions of the forward equation, which reads:
\begin{equation*}
    q(t) = \sqrt{d} e^{-t} + \sqrt{1 - e^{-2t}} z_i + e^{-t} \sigma \tilde{z}_i,
\end{equation*}
where \( z_i, \tilde{z}_i \sim \mathcal{N}(0,1) \) and \( t = t_f - \tau \). Let us now focus on the mean of $q$. When we consider
\begin{equation*}
  \int_{0}^{\tau} \left[\frac{2\sqrt{d} e^{-(t_f - \tau')}}{\Gamma(t_f-\tau')}\right] e^{(\tau - \tau')-2\int_{\tau'}^\tau \frac{1}{\Gamma(t_f-\tau'')}d\tau''} d\tau',
\end{equation*}
using that 
\begin{equation*}
    \frac{d}{d\tau'} \exp\left[-2\int_{\tau'}^{\tau} \frac{1}{\Gamma(t_f - \tau'')} d\tau''\right] = \frac{2}{\Gamma(t_f - \tau')}\exp\left[-2\int_{\tau'}^{\tau} \frac{1}{\Gamma(t_f - \tau'')} d\tau''\right],
\end{equation*}
one finds that the mean of $q$ for the evolution with $\omega=0$, starting from any value $q(0)$ at any time $t_f$, is 
\begin{equation}\label{bsol2}
    q(\tau) = q(0)e^{ \tau-2\int_0^\tau \frac{1}{\Gamma(t_f-\tau')}d\tau'} + \sqrt{d} e^{-(t_f - \tau)}\left( 1- \exp\left(-2 \int_{0}^\tau \frac{1}{\Gamma(t_f-\tau')}d\tau' 
    \right)
    \right).
   \end{equation}
Using
\begin{equation*}
    \int_0^\tau \frac{1}{\Gamma(t_f-\tau')}d\tau'=-\frac{1}{2}\log\frac{e^{-2\tau}+(\sigma^2-1)e^{-2 t_f}}{1+(\sigma^2-1)e^{-2 t_f}},
\end{equation*}
we find that
\begin{equation}\label{bsol3}
    q(\tau) = q(0)\; e^\tau \; 
    \frac{e^{-2\tau}+(\sigma^2-1)e^{-2 t_f}}{1+(\sigma^2-1)e^{-2 t_f}} + \sqrt{d} \; e^{-(t_f - \tau)} \; 
    \frac{1-e^{-2\tau}}{1+(\sigma^2-1)e^{-2 t_f}}.
   \end{equation}
One can check that, when $q(0)$ is obtained by the equilibrium process with $\omega=0$, namely $q(0)=\sqrt{d}e^{-t_f}$, then at all times $q(\tau)=\sqrt{d}e^{-(t_f-\tau)}$.

\subsection*{Case: interrupted guidance}
Now let us consider a protocol of interrupted guidance. We start the backward process at $t_f\gg 1$ with a CFG coefficient $\omega>0$. Then at time backward time $\tau_1$ (forward time $t_1= t_f-\tau_1$) we switch to $\omega=0$.
At time $t_1$ the mean of $q$ obtained from the backward process with $\omega>0$ is larger than the value $ \sqrt{d}e^{-t_1} $ which would be obtained with the $\omega=0$ dynamics (the reason is that the extra force in (\ref{basic_backward_eq_q}) is positive). Let us write this mean as 

\begin{equation*}
    q(t_1,\omega)= \sqrt{d}\; e^{-t_1}+\delta q(t_1,\omega).
\end{equation*}

Let us measure the backward time starting from $t=t_1$. We thus write $t=t_1-\tilde \tau$. We can use formula (\ref{bsol3}) with $t_f\to t_1$, $\tau\to \tilde \tau$ and $q(0)\to q(t_1,\omega)$. This gives for the mean value of $q$:
\begin{equation*}
\tilde q(\tilde \tau,\omega)= \sqrt{d} e^{-(t_1-\tilde \tau)}+\delta q(t_1,\omega) \; \frac{e^{-\tilde\tau}+(\sigma^2-1)e^{\tilde\tau-2 t_1}}{1+(\sigma^2-1)e^{-2 t_1}},
\end{equation*}
which, translated in terms of the forward time $t=t_1-\tilde \tau$, gives:
\begin{equation}
    q(t)= \sqrt{d} e^{-t}+ \delta q (t_1,\omega) \; e^{t-t_1}\; \frac{1+(\sigma^2-1)e^{-2t}}{1+(\sigma^2-1)e^{-2t_1}}.
    \label{eq:qav_switch_pred}
\end{equation}

In particular at the end of the backward process, for $\tilde\tau=t_1$ we get 

\begin{equation*}
 q(t=0)= \sqrt{d} +\delta q(t_1,\omega) \; e^{-t_1}\; \frac{\sigma^2}{1+(\sigma^2-1)e^{-2 t_1}}
\end{equation*}
If we choose $t_1=t_s=(1/2)\log d$, and assuming that the dynamics at $t>t_1$ has produced an average $q(t_1)=\sqrt{d} e^{-t_1}+\delta q$, we find that 
\begin{equation*}
 q(t=0)= \sqrt{d} \left(1+\delta q  \; \frac{\sigma^2/d}{1+(\sigma^2-1)/d}\right).
\end{equation*}

\begin{figure}[t]
    \centering
    \includegraphics[width=0.38\textwidth]{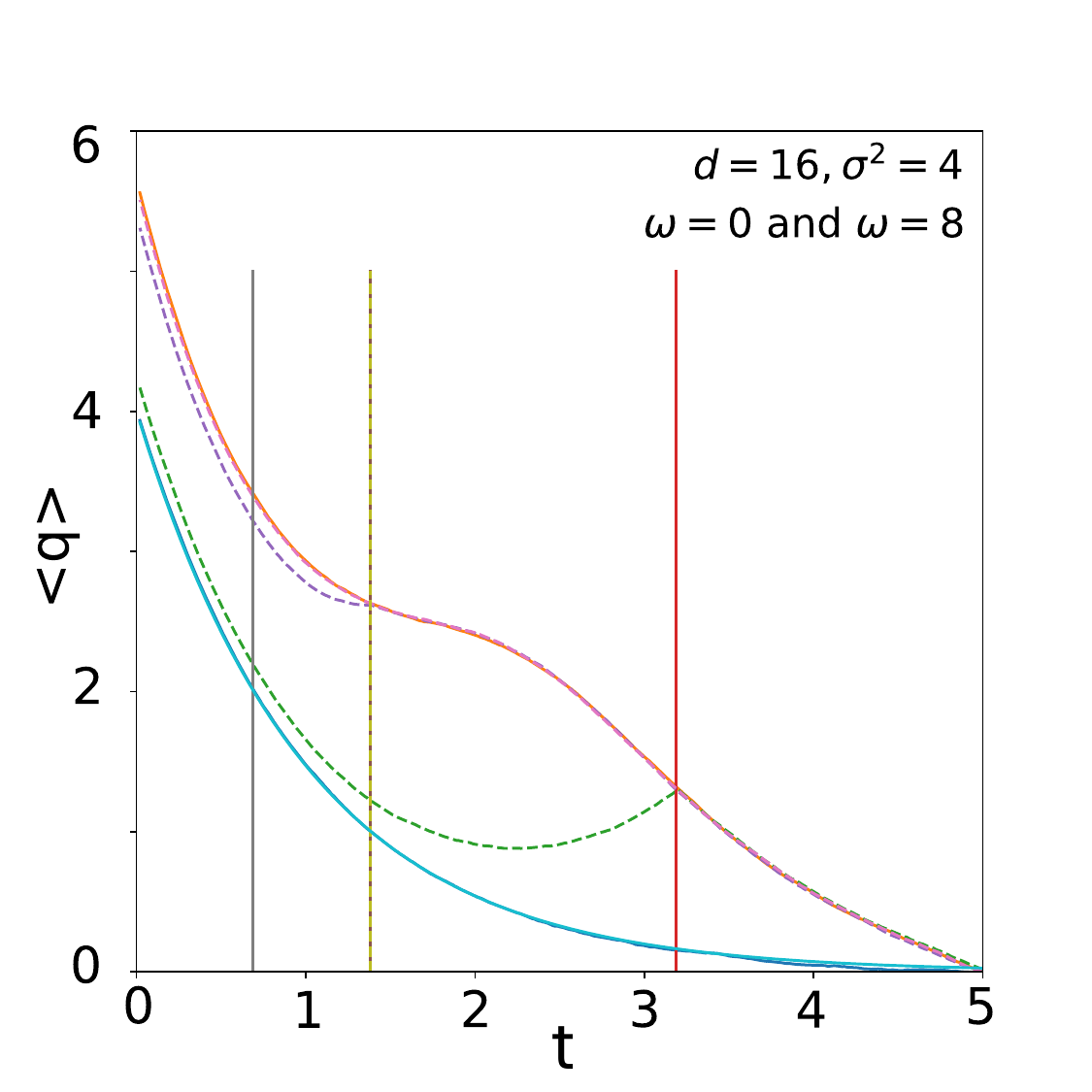}
    \caption{Mean value of $q$ obtained from the backward diffusion in a Gaussian mixture model with $d=16,\sigma^2=4$ (speciation time $t_s=1.38$). The CFG is run with $\omega=8$ from $t=5$ to $t=t_1$, then one switches to the class guidance $\omega=0$. The top curve is when CFG is kept all the time ($t_1=0)$. The bottom curve is the case without CFG ($\omega=0$). Three values of $t_1$ are studied $t_1=0.69,1.38,3.19$ (vertical lines). The dashed curves give the mean value of $q$ for each of these three cases. They are in perfect agreement with the theoretical prediction (\ref{eq:qav_switch_pred}).
       }
    \label{fig:switchomegamean}
\end{figure}

\begin{figure}[t]
    \centering
     \includegraphics[width=1.\linewidth]
     {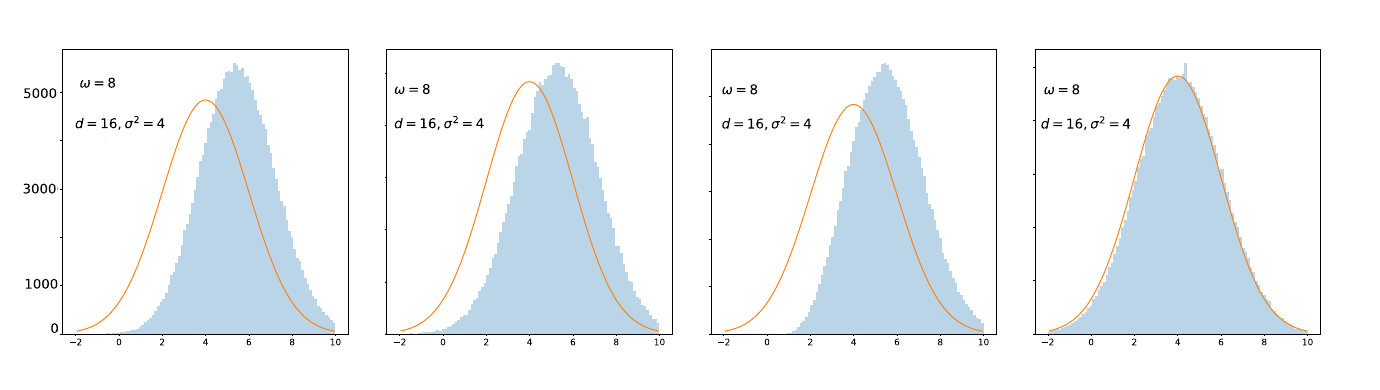}
    \caption{Histograms of $q(t=0)$ obtained from the backward diffusion in a Gaussian mixture model with $d=16,\sigma^2=4$ (the speciation time is $1.38$), run with $200,000$ trajectories. Left: CFG with $\omega=8$ is applied at all times. The final distribution has a larger mean and a smaller variance than the desired class distribution (full line). Next three figures:
    The CFG is run with $\omega=8$ from $t=5$ to $t=t_1$, then one switches to standard CFG $\omega=0$. From left to right, $t_1=0.69,1.38,3.19$. 
    The mean values of $q$ in the four cases are respectively $5.56, 5.51,5.29,4.12 $ and the standard deviations $1.68, 1.74 ,1.87 , 1.98$, with targets $\mu=4, \sigma=2$. 
    In order to minimize the bias due to CFG one must interrupt it before the speciation takes place in the background diffusion, hence at $t_1>t_s$.  
    }
    \label{fig:switchomega}
\end{figure}

This shows that the guidance interrupted at $t_s$ gives a good result only in the limit $\sigma^2/d\ll 1$. 
Figs. \ref{fig:switchomegamean} and \ref{fig:switchomega} illustrate the effect of the choice of $t_1$. 

\subsection*{CFG contribution to the magnetization in Regime \rom{1}}

Using Equation (14), one can derive the equation for the average \( \langle q(\tau) \rangle_\omega \):

\begin{align}
    \frac{d\langle q(\tau) \rangle_\omega}{d\tau} 
    &= \langle q(\tau) \rangle_\omega \left( 1 - \frac{2}{\Gamma(t_f-\tau)} \right) 
    + \frac{2\sqrt{d}}{\Gamma(t_f-\tau)} e^{-(t_f-\tau)} \nonumber \\
    &\quad + 2\omega\sqrt{d} \frac{e^{-(t_f-\tau)}}{\Gamma(t_f-\tau)} 
    \left\langle 1 - \tanh\left( \frac{q \sqrt{d} e^{-(t_f-\tau)}}{\Gamma(t_f-\tau)} \right) \right\rangle_\omega.
\end{align}

The extra \( \omega \) term is strictly positive. Therefore, we have:

\begin{equation*}
    \langle q(\tau) \rangle_\omega \geq \langle q(\tau) \rangle_{\omega=0}, \quad \forall \tau.
\end{equation*}

Moreover, using that the right-hand side is less than or equal to:

\begin{equation*}
    \langle q(\tau) \rangle_\omega \left( 1 - \frac{2}{\Gamma(t_f-\tau)} \right) 
    + \frac{2(1+\omega)\sqrt{d}}{\Gamma(t_f-\tau)} e^{-(t_f-\tau)},
\end{equation*}

which corresponds to the backward equation one would obtain if \( \|\vec{m}\|^2 =  (1+\omega)d \). We then find:

\begin{equation*}
   \langle q(\tau) \rangle_{\omega=0} <  \langle q(\tau) \rangle_\omega <\sqrt{d} e^{-t} (1+\omega).
\end{equation*}

We conclude that \( \langle q(\tau) \rangle_\omega \) gets an extra contribution due to CFG of the order \( \sqrt{d} e^{-t} \).

CFG indeed shifts the mean value. The amount of shift is of order $\sqrt{d}e^{-t}$ in Regime \rom{1}. However, as we shall see next the CFG has almost no effect in Regime \rom{2}, so we can use the result of the previous section to argue that the total shift due to CFG is the one of CFG in Regime \rom{1} followed by a switch at $\omega=0$ in Regime \rom{2}, i.e., it is of order one.  

\subsection*{CFG contribution to the score in Regime \rom{1} vs in Regime \rom{2}}

Another interesting inequality can be derived for the difference between the CFG and the standard, non-guided score, \( S_{\text{CFG}} - S_C \), evaluated on trajectories corresponding to CFG:

\begin{align}
    S_{\text{CFG}} - S_C 
    &= \omega \frac{\sqrt{d} e^{-(t_f-\tau)}}{\Gamma(t_f-\tau)} 
    \left( 1 - \tanh\left( \frac{q \sqrt{d} e^{-(t_f-\tau)}}{\Gamma(t_f-\tau)} \right) \right).
\end{align}

We use the fact that for the same thermal noise, we have \( q_\omega(\tau) \geq q_{\omega=0}(\tau) \) because the CFG force is always equal or larger than the $\omega=0$ one. Therefore for a given (the same) thermal history we have: 

\begin{equation}
    -\tanh\left(\frac{q_\omega(\tau)\sqrt{d} e^{-(t_f-\tau)}}{\Gamma(t_f-\tau)}\right) 
    \leq -\tanh\left(\frac{q_{\omega=0}(\tau)\sqrt{d} e^{-(t_f-\tau)}}{\Gamma(t_f-\tau)}\right),
\end{equation}

and we can obtain:

\begin{align}
    S_{\text{CFG}} - S_C 
    &\leq \frac{\sqrt{d} e^{-(t_f-\tau)}}{\Gamma(t_f-\tau)} 
    \left( 1 - \tanh\left(\frac{q_{\omega=0}(\tau)\sqrt{d} e^{-(t_f-\tau)}}{\Gamma(t_f-\tau)}\right) \right).
\end{align}

This inequality tells us, as expected, that the extra CFG contribution to the score is very small at the beginning of the backward process. Its mean increases, and is of the order of one during the backward process in Regime \rom{1}. However, after the speciation time $q_{\omega=0}(\tau)$ is a Gaussian variable with a mean $\sqrt{d}e^{-(t_f-\tau)}$ much larger than the square root of the variance. Therefore, replacing the fluctuating variable by its mean we obtain  
\begin{align}
    S_{\text{CFG}} - S_C 
    &\leq \frac{\sqrt{d} e^{-(t_f-\tau)}}{\Gamma(t_f-\tau)} 
    \left( 1 - \tanh\left(\frac{{d} e^{-2(t_f-\tau)}}{\Gamma(t_f-\tau)}\right) \right).
\end{align}
In Regime \rom{2}, $t_f-\tau$ is of order one, and using the asymptotic form of the hyperbolic tangent one finds that
\begin{align}
    S_{\text{CFG}} - S_C 
    &\leq \frac{\sqrt{d} e^{-(t_f-\tau)}}{\Gamma(t_f-\tau)} 
   \exp\left(-2\frac{{d} e^{-2(t_f-\tau)}}{\Gamma(t_f-\tau)}\right) .
\end{align}
Therefore in Regime \rom{2} the extra contribution to the score is exponentially small in $d$ and its effect is completely negligible with respect to the one in Regime \rom{1}. 
\subsection*{Analysis of the CFG effect on the variance}

Let us derive the equation for \( \langle q^2(\tau) \rangle_\omega - \langle q(\tau) \rangle_\omega^2 \).

Using Itô calculus, we have (multiplying by \( q(\tau) \) in the equation for \( \frac{dq(\tau)}{d\tau} \)):

\begin{align}
    \frac{d q^2(\tau) }{d\tau} 
    &= 2 + 2  q^2(\tau)  \left( 1 - \frac{2}{\Gamma(t_f-\tau)} \right) 
    + 2  q(\tau)  \frac{2\sqrt{d}}{\Gamma(t_f-\tau)} e^{-(t_f-\tau)} \nonumber \\
    &\quad + 2 \frac{2\omega\sqrt{d}}{\Gamma(t_f-\tau)} e^{-(t_f-\tau)}
    \left( q(\tau) - q(\tau) \tanh\left( \frac{q(\tau)\sqrt{d} e^{-(t_f-\tau)}}{\Gamma(t_f-\tau)} \right) \right) \nonumber \\
    &\quad + 2q(\tau) \eta(\tau).
\end{align}

Taking the average and subtracting \( 2\langle q(\tau) \rangle_\omega \frac{d\langle q(\tau) \rangle_\omega}{d\tau} \), we find the equation for 
\( \langle q^2(\tau) \rangle_\omega - \langle q(\tau) \rangle_\omega^2 \):

\begin{align}
    \frac{d \langle q^2(\tau) \rangle_\omega - \langle q(\tau) \rangle_\omega^2}{d\tau} 
    &= 2 + 2 \left( \langle q^2(\tau) \rangle_\omega - \langle q(\tau) \rangle_\omega^2\right) \left( 1 - \frac{2}{\Gamma(t_f-\tau)} \right) \nonumber \\
    &\quad + \omega \frac{4\sqrt{d} e^{-(t_f-\tau)}}{\Gamma(t_f-\tau)} 
    \left( \langle q(\tau) \rangle_\omega 
    \left\langle \tanh\left(\frac{q\sqrt{d} e^{-(t_f-\tau)}}{\Gamma(t_f-\tau)} \right) \right\rangle_\omega \right. \nonumber \\
    &\quad \left. - \langle q(\tau) \tanh\left(\frac{q\sqrt{d} e^{-(t_f-\tau)}}{\Gamma(t_f-\tau)} \right) \rangle_\omega \right).
\end{align}

%\rd{**Question:** Can one determine the sign of the CFG extra term?}

At the beginning of the backward process, one can expand \( \tanh(x) \) and observe that the term in the parentheses is proportional to:

\begin{equation}
    -\left(\langle q(\tau)^2 \rangle_\omega - \langle q(\tau) \rangle_\omega^2\right),
\end{equation}

which is negative. Therefore, we can conclude that the classifier-free-guidance-added term will result in shrinkage of the variance.

As for the mean, the main CFG effect on the variance is produced in Regime \rom{1}, since the CFG score term is exponentially small in Regime \rom{2}.

\section{Experimental details: Gaussian mixtures}
\label{sec:appx_E_gm}
In this section, we present experimental details for the numerical simulations involving Gaussian mixtures, describing the procedures and the hyperparameter configurations.

\paragraph{Numerical simulations.} In the case of a mixture of two Gaussian clusters centered on $\pm \vec{m} \in \mathbb{R}^d$ with variance $\sigma^2$, the score function reads as

\begin{align*}
    S_{t_{CFG}}(\vec{x}(t),c) = -\frac{x(t)}{\Gamma_t} + \omega\frac{\vec{m} e^{-t}}{\Gamma_t}\left\{c-\tanh{\left(\frac{\vec{x}(t)\cdot\vec{m}e^{-t}}{\Gamma_t}\right)}\right\} + \frac{c\vec{m} e^{t}}{\Gamma_t},
\end{align*}

where $\Gamma_t=\Delta_t+\sigma^2 e^{-2 t}$, with $\Delta_t=1-e^{-2 t}$. We can then discretize the stochastic differential equation associated to the backward process as
\begin{align*}
    \vec{x}(t+1)=\vec{x}(t)+\eta\left[\vec{x}(t)+2 S_{t_{CFG}}(\vec{x}(t),c)\right]+\vec{\eta} \sqrt{2 \tau/L},
\end{align*}

where $\vec{\eta} \sim \mathcal{N}(0, I)$, with $t_f=8$ the time horizon and $t_f/L=0.01$. We use $\vec{m}=[1, \ldots, 1], \sigma^2=1$, and each point is obtained by averaging over 100 initial conditions. The speciation time $t_s$ is calculated as $t_s = -\frac1{2}\log d$. Throughout the  experiments, we plot the evolution of $q(t)=\frac{\vec{x}\cdot\vec{m}}{|\vec{m}|}$, conditioning the guidance on the positive class with $c=1$.

\section{Experimental details: Real-world analyses}
\label{sec:appx_F_real_world}

\subsection{Assets}
In \Cref{tab:assets} we list the datasets and models
used in our work along with their licensing.

\begin{table}[ht]
    \centering
    \caption{Assets used for  our work. }
    {\scriptsize
    \begin{tabular}{lc}
        \toprule
        Name & License/Link \\
        \midrule
        COCO'14 & \url{https://www.cocodataset.org}\\
        ImageNet & \url{https://www.image-net.org}\\
        CC12M & \url{https://github.com/google-research-datasets/conceptual-12m}\\
        YFCC100M & \url{https://www.multimediacommons.org}\\
        Florence-2 & \url{https://huggingface.co/microsoft/Florence-2-large/blob/main/LICENSE} \\
        \midrule
        DiT & \url{https://github.com/facebookresearch/DiT/blob/main/LICENSE.txt} \\
        EDM2 &\url{https://github.com/NVlabs/edm2/blob/main/LICENSE.txt} \\
        MMDiT & \url{https://github.com/lucidrains/mmdit/blob/main/LICENSE} \\
        MDTv2 & \url{https://github.com/sail-sg/MDT/blob/main/LICENSE} \\
        \bottomrule
    \end{tabular}}
    \label{tab:assets}
\end{table}

\subsection{Performing the time reparameterization}
\label{sec:time_reparameterization}
In the second part of the paper, we evaluate the score of DiT models, in discrete time, as introduced by \citet{peebles2023scalable}. In this context, the forward process has a linear variance schedule $\left\{\beta_t^{\prime}\right\}_{t^{\prime}=1}^L$, where $L$ is the time horizon given as a number of steps. Here, the variance evolves linearly from $\beta_1=10^{-4}$ to $\beta_{1000}=2 \times 10^{-2}$. An unguided sample, at timestep $t^{\prime}$, denoted $\vec{x}\left(t^{\prime}\right)$ can be expressed readily from its initial state, $\vec{x}(0)=\vec{a}$, as
\begin{align*}
    \vec{x}\left(t^{\prime}\right)=\sqrt{\bar{\alpha}}\left(t^{\prime}\right) \vec{a}+\sqrt{1-\bar{\alpha}\left(t^{\prime}\right)} \vec{\xi}\left(t^{\prime}\right)
\end{align*}

where $\bar{\alpha}\left(t^{\prime}\right)=\prod_{s=1}^{t^{\prime}}\left(1-\beta_s\right)$ and $\vec{\xi}$ is standard Gaussian noise. This equation corresponds to the discretization of the Ornstein-Uhlenbeck Eq. (\ref{eqn:ou}) under the following timestep $t^{\prime}$ reparameterization,

\begin{align*}
    t=-\frac{1}{2} \log \left(\bar{\alpha}\left(t^{\prime}\right)\right),
\end{align*}

where time $t$ is as defined in the main manuscript. This gives the map between our theoretical timescale used in Gaussian mixtures, and the one used in real-world settings. We note that, as the neural network predicts the noise, in order to calculate the score, one needs to normalize the output by the standard deviation (depending on the variance schedule). In this case, this corresponds to dividing the neural network output by $\sigma(t')=\sqrt{1-\bar{\alpha}(t')}$. \textbf{In numerical experiments, we divide the CFG-added-term} by $\sigma(t')+1$ to avoid numerical errors. This is theoretically justified due to the fact that, as discussed in main paper, the score difference $|S_{t'}(\vec{x},c)-S_{t'}(\vec{x})|$ for large forward times decays exponentially (as $e^{-t'}$) to zero.

For completeness, we present the full comparison of numerical simulations to real-world using the time-reparameterization to plot the timesteps on the same time-scale. Our findings are portrayed in Figure \ref{fig:correct_time}. As each framework uses a separate time reparameterization, the x-axis needs to be recalculated accordingly. For the EDM2 framework \citep{karras2022elucidating}, this can be done as follows: given a noise schedule $\sigma(t)$, the reparameterization can be calculated as $t' (t)=(1/2) \log(1+\sigma^2(t))$, assuming that $s(t)=1$. For the case $s(t)$, one needs to resort to equation \Cref{eqn:backw_dyn}.

\subsection{Hyperparameter configurations}
\label{sec:appx_hyperparam_configs}

Here, we give exact hyperparameters used for reproducing all our experiments. The real-world experiments are performed using NVIDIA H100  Tensor Core - 80GB HBM3. The EDM2-S model has a model size of 280 Mparams and 102 Gflops, whereas the DiT-XL/2 model has model size of 675 Mparams and 525 Gflops. Parameter $\alpha$ is tuned in $(0.3, 0.95)$ with an increment of $0.05$ and parameter $\omega$ is tuned in $(1.,12.)$ with an increment of $0.05$. To tune $\omega$, we first perform a small grid search of the increment of $1.$ and then do a further extensive search of the best performing $\omega_{prelim}$ in the range $(\omega_{prelim}-2., \omega_{prelim}+2.)$ with the $0.05$ increment. We begin with the hyerparameters used in our figures.

In \textbf{Figure \ref{fig:intro_img}}, we plot the generation of images starting from 7  initial seeds for the DiT/XL-2 model trained on ImageNet-1K ($256\times256$) for (1) conditional model without using guidance, (2) standard CFG with $\omega=4.$, and Power-Law CFG with $\alpha=0.9, \omega=8$.

In \textbf{Figure \ref{fig:intro_motiv}}, the first two plots correspond to the histograms of the samples generated using the backward process with dimensions $d\in\{2,200\}$ and guidance parameter $\omega\in\{0, 0.2, 15\}$, with $\sigma^2=1$, averaged over $10,000$ trajectories. The last two plots correspond to the actual trajectories projected onto the target mean $+\vec{m}$ for values of $\omega\in\{0.,5.,10.,15.,20.\}$.

In \textbf{Figure \ref{fig:2_diffusion_regimes}}, we plot the evolution of the 1D backward dynamics with means at $\pm4$ and unit variance. The potential plotted corresponds to equation $V(q,t)=\frac1{2}q^2-2\log\cosh(q e^{-(t-t_s)})$. For the derivation of this potential, see Appendix B.2 in \citet{biroli2024dynamical}. 

In \textbf{left part} of \textbf{Figure \ref{fig:6_cfg_score_diff}}, we evaluate the difference of the conditional and unconditional score for Gaussian mixtures, $|S_t(x,c)-S_t(x)|$ for dimensions $d\in\{1,5,20,50,200\}$ with 2 classes, using $\omega=5, \sigma^2=1$, averaged over $10,000$ trajectories. For the \textbf{middle part,} we use $\omega=5., d=200$ and change the $\alpha\in\{-0.5,-0.3,-0.1,0.,0.2,0.5,0.9\}$. The forward time in the equation goes from $0.$ to $8.$, which we denote as \textit{sampling time} going from $0$ (corresponding to data) to $1$ (corresponding to noise) for simplicity of the exposition. In \textbf{right part}, we plot the score difference for four real-world models: DiT/XL-2, EDM2-S, MMDiT and MDTv2, all trained using the diffusion objective. As each of the models' default hyperparameters have different number of sampling steps, we normalize the x-axis from 0 to 1: e.g.,  DiT framework uses 250 sampling steps, EDM2 uses 32 and for the text-to-image models we use 50 sampling steps. The y-axis is normalized to be between 0 and 1 for easier readability. All hyperparameters are set to the default ones.

In \textbf{Figure \ref{fig:sens_analysis}}, we perform sensitivity analysis for EDM2-S trained on ImageNet-1K ($512\times512$), taking $\alpha$ from $0.$ to $0.99$ with 20 evenly spaced values, and $\omega$ from $1.$ to $10.$ with 20 evenly spaced values as well. The right plot involves $\alpha$ values of $0.2, 0.4, 0.6, 0.8, 0.9$ with $\omega$ in the range of $1.$ to $12.5$ with evenly spaced 20 points.

In \textbf{Figure \ref{fig:increasing_omega}} we show generations of DiT/XL-2 trained on ImageNet-1K ($256\times256$). The red panel contains generations from weak and strong standard CFG (corresponding to $\omega=2.$ and $\omega=5.$ respectively). The green panel corresponds to power-law CFG ($\alpha=0.9$) with weak and strong guidance (corresponding to $\omega=2.$ and $\omega=10.$). The blue panel corresponds to combinations of $\alpha$ and $\omega$ $(0,2.5)$, $(0.25, 4.)$, $(0.5,6.)$ and $(0.9, 8.)$.

In \textbf{Figure \ref{fig:4_potentials}}, we examine the following functions:
\begin{align*}
V_{\text{class}}(q, t; c) &= \frac{1}{2}q^2 - ce^{-(t-t_s)}q + 2 \\
V_{\text{extra}}(q, t; c) &= -ce^{-(t-t_s)}q + \log\left(\cosh\left(qe^{-(t-t_s)}\right)\right) + \log(2),
\end{align*}
where the plots correspond to $V_{\text{class}}$, $V_{\text{extr}}$ and $(V_{\text{class}}+\omega V_{\text{extr}})$ with $\omega=3$ respectively. We select $c=1$, and fix the speciation time to $t_s=.5$. The additive constants are added for clarity only.

In \textbf{Figure \ref{fig:switchomegamean}}, we plot the backward diffusion in a Gaussian mixture model with $d=16,\sigma^2=4, \omega=8$. The CFG is either run at all times (top curve), stopped at times $t_1$ or not used at all (bottom curve).

In \textbf{Figure \ref{fig:switchomega}}, we perform linear CFG with $\omega=8$ from $t=5$ to $t=t_1$, after which we turn CFG off ($\omega=0$) at times $t_1=0.69,1.38,3.19$.

\begin{figure*}[t]
    \centering
    \begin{NiceTabular}{ccc}
        \includegraphics[width=0.3\textwidth]{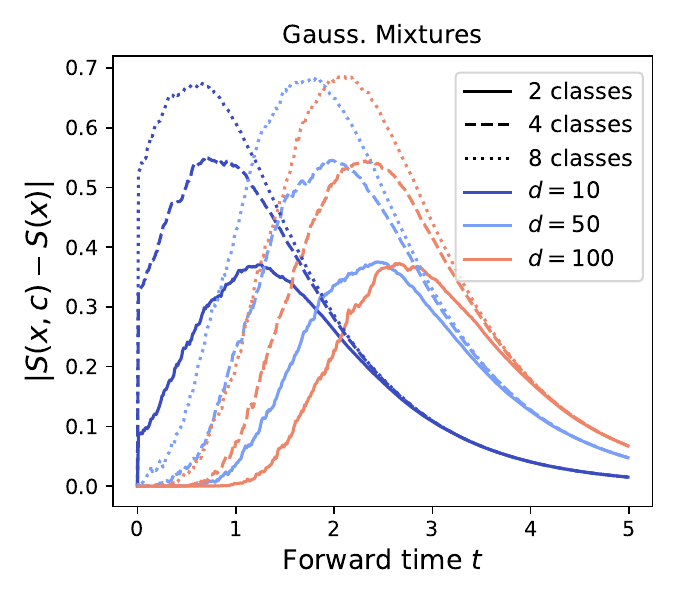} &
        \includegraphics[width=0.3\textwidth]{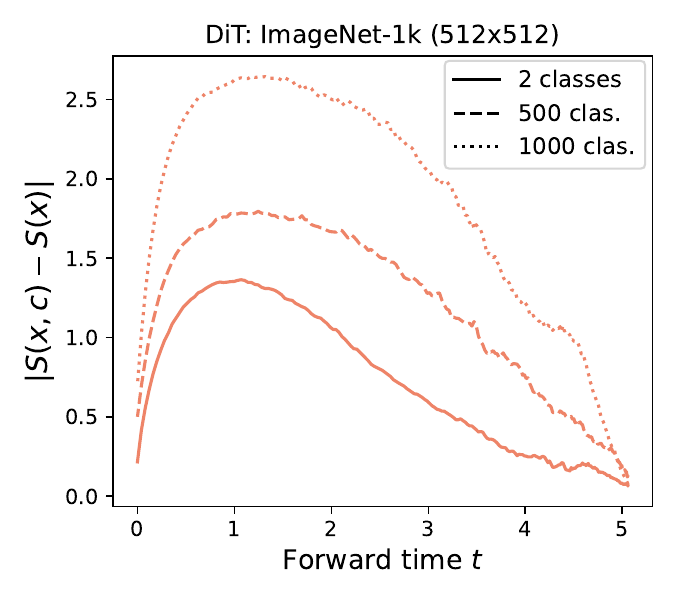} &
        \includegraphics[width=0.3\textwidth]{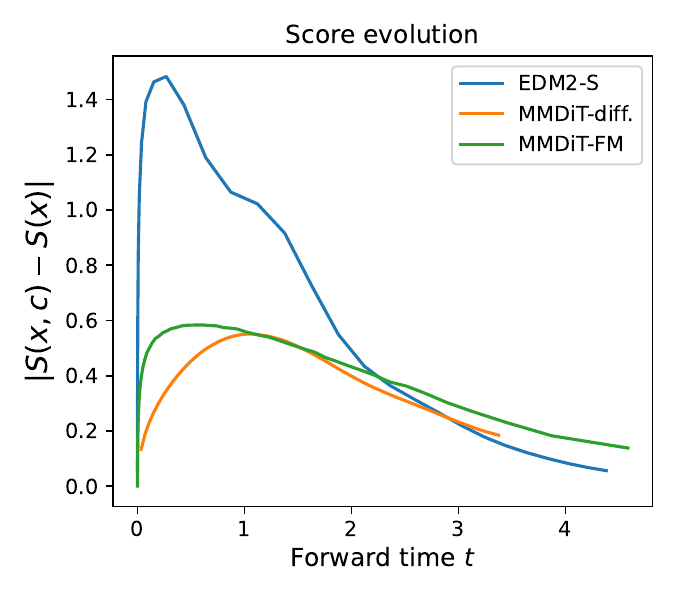}
    \end{NiceTabular}
      \caption[Caption for LOF]{\textbf{Evolution of the score differences for numerical simulations and real-world experiments} projected onto the same time-scale for direct comparison. \textbf{First:} Numerically simulating mixture of two, four, and eight Gaussians with equidistant means on a sphere ($r=\sqrt{d}$), with varying dimension $d$, with $\omega=4, \sigma^2=1$, averaged over 10,000 trajectories. As $d$ increases, the score difference starts to increase at an earlier backward time $\tau$. Additionally, as the number of classes increases, the magnitude of the score difference grows, as well as the duration of large difference between the scores. \textbf{Second:} Three DiT/XL-2 models trained on ImageNet-1K using 2, 500, and 1000 classes (image size $512\times 512$). We observe a similar pattern: as $d$ increases, the score difference becomes larger at an earlier time. Furthermore, as the number of classes increases, the magnitude of the score difference increases, together with the duration for which the difference remains large. \textbf{Third:} evolution of the remaining models used in our experiments (EDM2-S, MMDiT and MDTv2). We observe a similar behavior to theory and the DiT/XL-2 models.}
      \label{fig:correct_time}
\end{figure*}

In \textbf{Figure \ref{fig:correct_time}}, we use DiT/XL-2 model trained on 2, 500 and 1000 classes. For 2 classes, we have selected the same classes as in \citet{biroli2024dynamical}, and for the 500 classes we selected the first 500 classes in ImageNet-1K. The x-axis represents the Forward time $t$, where the parameterization is obtained as explained in \Cref{sec:time_reparameterization}.

In \textbf{Figures \ref{fig:newnonlin1}-\ref{fig:newnonlin2}}, we perform the same experimenet as in Figure \ref{fig:correct_time} and use $d=16$ and $\sigma^2=4$.

Finally, in the \textbf{first column} of \textbf{Figure \ref{fig:large}}, we plot the estimated Jensen-Shannon Divergence between the target samples corresponding to a randomly selected class and the diffusion particles throughout the backward trajectory. Note that this is performed in latent space. For obtaining the \textbf{middle column}, we first take all data samples from one class, embed them into the latent space and calculate the centroid corresponding to this class. Then, we normalize the centroid (making it unit norm) and plot the dot product of the particles throughout the backward diffusion process with the calculated centroid. The \textbf{right column} corresponds to the score difference. Across all experiments, we selected $\omega=4$, sampled using DDPM \citep{ho2020denoisingdiffusionprobabilisticmodels} using 250 sampling steps, averaged over 25 samples. All other hyperparameter configurations are set to the default.

\textbf{Table \ref{tab:hyperparameters}} displays the hyperparameters used to obtain the results  given in Table \ref{tab:1}. The evaluation code relied on EvalGIM library by \citet{hall2024evalgim}. 

\begin{table}[t]
\centering
\caption{Hyperparameter configurations used throughout the experiments.}
\resizebox{\columnwidth}{!}{
\begin{tabular}{c|cccccc}
\toprule
$(\alpha,\omega)$ & DiT/XL-2      & EDM2-S          & Diff. MMDiT CC12m & Diff. MDTv2 IMN-1K & FM MMDiT COCO   & FM MMDiT CC12M \\
\midrule
Standard          & $(0., 1.5)$   & $(0.,1.4)$      & $(0., 1.55)$      & $(0.,1,2)$         &$ (0.,2.)$       & $(0.,2.1)$ \\
Non-linear        & $(0.75,4.85)$ & $(0.85, 11.4)$  & $(0.6,7.0)$       & $(0.8, 8.5)$       & $(0.7,10.15)$   & $(0.6,8.05)$ \\
Non-lin. + Limited& $(0.8,4.95)$  & $(0.9,12.05)$   & $(0.55,8.25)$     & $(0.85,8.25)$      & $(0.75,10.05)$  & $(0.65,7.85)$ \\
Non-lin. + CADS   & $(0.7,4.75)$  & $(0.80, 11.75)$ & $(0.75, 8.15)$    & $(0.80,8.40)$      & $(0.75, 10.75)$ & $(0.55,7.90)$\\
\bottomrule
\end{tabular}
}
\label{tab:hyperparameters}
\end{table}

\subsection{Further results}

Here, we detail the remaining experiments conducted. We provide the following:
\begin{itemize}
    \item Diversity and coverage metrics corresponding to experiments in \Cref{tab:1} (see Table \ref{tab_4_appx})
    \item Ablation studies showing that Power-Law CFG outperforms standard linear CFG when changing the number of steps (see Tables \ref{tab_5_appx}-\ref{tab_10_appx})
    \item Sensitivity analysis showing the FID benefit for increasing value of $\alpha$ (see Section \ref{sec:sens_anal}, Figures \ref{fig:11_appx}-\ref{fig:13_appx})
    \item Further qualitative analyses of power-law CFG for either fixed $\omega$ and varying $\alpha$ or varying $\omega$ and varying $\alpha$ (see Sections \ref{sec:qualit_anal_fix_alpha} and \ref{sec:qualit_anal_fix_omega})
    \item Further generation examples of DiT/XL-2 and MMDiT diffusion model (see Sections \ref{appx:ex_by_dit} and \ref{appx:ex_by_tti})
\end{itemize}

\paragraph{Diversity and coverage metrics.} In Table \ref{tab_4_appx} present additional quantitative evaluations of our method, focusing on diversity and coverage metrics (as described in \citet{hall2024evalgim}), which complement the results shown in \Cref{tab:1}. Our analysis compares power-law CFG to standard CFG and state-of-the-art guidance methods, including combinations with CADS~\citep{sadat2023cads} and limited-guidance~\citep{kynkaanniemi2024applying}, which proved to be the most competitive approaches. As demonstrated in the main manuscript, power-law CFG generally outperforms standard CFG (indicated by arrows in the table). Moreover, when combined with CADS and limited-interval guidance, it yields improved results over existing methods in many cases.

\paragraph{Ablation studies.} In Tables \ref{tab_5_appx} through \ref{tab_10_appx}, we conduct ablation studies on two class-conditional and four text-to-image models, demonstrating that non-linear power-law CFG consistently surpasses standard CFG across varying sampling steps. The results show improved FID performance and enhanced outcomes across multiple metrics when using the non-linear approach compared to standard CFG.

\paragraph{Sensitivity analysis.} In Section \ref{sec:sens_anal}, we present additional sensitivity analyses that build on Section \ref{sec:nonlin} and Figure \ref{fig:sens_analysis}, demonstrating that high values of $\alpha$ consistently enhance performance, improving robustness and stability during $\omega$ tuning. As noted in the main manuscript, while power-law CFG introduces an additional hyperparameter, $\alpha$, extensive hyperparameter tuning was unnecessary, with large values like $\alpha=0.9$ consistently performing well. This is evidenced in Section \ref{sec:sens_anal}, Figures \ref{fig:11_appx} to \ref{fig:13_appx}, which show that higher $\alpha$ values reliably improve FID scores. Class-conditional models (Figure \ref{fig:11_appx}) exhibit greater benefits than text-to-image models (Figures \ref{fig:12_appx} and \ref{fig:13_appx}), though both show improved performance with Power-Law CFG compared to standard CFG.

\paragraph{Further qualitative analyses.} In Sections \ref{sec:qualit_anal_fix_alpha} and \ref{sec:qualit_anal_fix_omega}, we provide additional qualitative examples for DiT-XL/2. Specifically, we conduct two studies: one varying the guidance parameter $\omega$ with a fixed $\alpha$, and another varying $\alpha$ with a fixed $\omega$. When $\alpha$ is fixed, increasing $\omega$ can lead to issues such as complete mode collapse (e.g., for the \textit{jellyfish} class), oversaturation (e.g., for the \textit{bee} class), or a significant loss of diversity (e.g., for the \textit{dung beetle} class), which are common artifacts of standard classifier-free guidance. These effects are mitigated when using a non-linear power-law guidance approach. The second study explores the impact of increasing $\alpha$ while keeping $\omega$ constant, demonstrating enhanced diversity as $\alpha$ strength increases.

\paragraph{Further generation examples.} In Sections \ref{appx:ex_by_dit} and \ref{appx:ex_by_tti}, we present additional generation examples for class-conditional (DiT/XL-2) and text-to-image (MMDiT) models, demonstrating how power-law CFG enhances image details, thereby improving image quality and fidelity for individual images, and increases diversity when examining a set of images for a specific class.

% \ref{sec:sens_anal}
% \ref{sec:qualit_anal_fix_omega}
% \ref{sec:qualit_anal_fix_alpha}
% \ref{appx:ex_by_dit}
% \ref{appx:ex_by_tti}

\begin{table}[ht]
  \centering
  \caption{Comparison of EDM2-S on ImageNet-1K 512x512 data, Diffusion trained text-to-image MMDiT on CC12m data, and Flow-matching trained text-to-image MMDiT on COCO data. \textbf{Bolded} are the best results and \underline{underlined} are the second best.}
  \resizebox{\textwidth}{!}{\begin{tabular}{l|HHHcc|HHHcc|HHHcc|HHHcc}
    \toprule
    Model & \multicolumn{5}{c}{EDM2-S (CC, IM-1K 512)} & \multicolumn{5}{c}{DiT/XL-2 (CC, IM-1K 256)} & \multicolumn{5}{c}{Diff. MMDiT (T2IM, CC12m)} & \multicolumn{5}{c}{FM MMDiT (T2IM, COCO)} \\
    & FID & Precision & Recall & Density & Coverage & FID & Precision & Recall & Density & Coverage & FID & Precision & Recall & Density & Coverage & FID & Precision & Recall & Density & Coverage \\
    \midrule
    Standard \citep{ho2022classifier} & 2.29 & 0.751 & 0.582 & 0.850 & 0.764 & & & & 0.951 & 0.801 & 8.58 & 0.661 & 0.569 & 1.091 & 0.840 & 5.20 & 0.629 & 0.594 & 0.902 & 0.772 \\
    Scheduler \citep{wang2024analysis} & 2.03 & 0.762 & 0.591 & \textbf{0.867} & \underline{0.780} & & & & \underline{1.117} & 0.790 & 8.30 & 0.681 & 0.559 & 1.266 & 0.860 & 5.00 & 0.606 & 0.623 & 0.908 & 0.795 \\
    Limited \citep{kynkaanniemi2024applying} & 1.87 & 0.760 & 0.598 & 0.845 & 0.777 & & & & \textbf{1.130} & 0.840 & 8.58 & 0.680 & 0.553 & 1.258 & 0.857 & 5.00 & 0.609 & 0.602 & 0.915 & \textbf{0.808} \\
    Cosine \citep{gao2023masked} & 2.15 & 0.770 & 0.619 & 0.850 & 0.769 & & & & 1.102 & 0.822 & 8.29 & 0.659 & 0.564 & 1.106 & 0.840 & 5.14 & 0.630 & 0.616 & 0.920 & 0.802 \\
    CADS \citep{sadat2023cads} & \underline{1.60} & \textbf{0.792} & 0.619 & 0.854 & 0.765 & & & & 0.999 & 0.853 & 8.32 & \textbf{0.692} & 0.559 & 1.222 & 0.860 & 4.91 & \underline{0.633} & 0.613 & \underline{0.923} & 0.779 \\
    APG \citep{sadat2024eliminating} & 2.13 & 0.756 & \textbf{0.640} & 0.845 & 0.760 & & & & 1.033 & \underline{0.867} & 8.49 & 0.661 & \underline{0.571} & 1.095 & 0.858 & 5.23 & 0.614 & \textbf{0.631} & 0.915 & 0.797 \\
    REG \citep{xia2024rectified} & 1.99 & 0.761 & 0.608 & 0.850 & 0.771 & & & & 1.112 & 0.833 & \underline{8.10} & 0.673 & 0.540 & 1.091 & 0.855 & 5.06 & 0.619 & 0.619 & 0.903 & 0.783 \\
    CFG++ \citep{chung2024cfg++} & N/A & N/A & N/A & N/A & N/A & & & & N/A & N/A & 8.35 & 0.668 & 0.552 & 1.265 & 0.859 & 4.85 & 0.632 & 0.629 & 0.919 & 0.784 \\
    \midrule
    Power-law CFG (Ours) & 1.62 ($\textcolor{green}{\downarrow}$) & \underline{0.780} ($\textcolor{green}{\uparrow}$) & \underline{0.631} ($\textcolor{green}{\uparrow}$) & 0.845 ($\textcolor{red}{\downarrow}$) & 0.760 ($\textcolor{green}{\uparrow}$) & 1.62 ($\textcolor{green}{\downarrow}$) & \underline{0.780} ($\textcolor{green}{\uparrow}$) & \underline{0.631} ($\textcolor{green}{\uparrow}$) & 0.986 ($\textcolor{green}{\uparrow}$) & 0.844 ($\textcolor{green}{\uparrow}$) & 8.11 ($\textcolor{green}{\downarrow}$) & 0.670 ($\textcolor{green}{\uparrow}$) & 0.553 ($\textcolor{red}{\downarrow}$) & 1.128 ($\textcolor{green}{\uparrow}$) & 0.850 ($\textcolor{green}{\uparrow}$) & \underline{4.81} ($\textcolor{green}{\downarrow}$) & 0.621 ($\textcolor{red}{\downarrow}$) & 0.619 ($\textcolor{green}{\uparrow}$) & 0.918 ($\textcolor{green}{\uparrow}$) & 0.778 ($\textcolor{green}{\uparrow}$) \\
    Power-law CFG + Limited (Ours) & 1.73 ($\textcolor{green}{\downarrow}$) & 0.752 ($\textcolor{red}{\downarrow}$) & 0.600 ($\textcolor{green}{\uparrow}$) & 0.850 ($\textcolor{green}{\uparrow}$) & 0.778 ($\textcolor{green}{\uparrow}$) & 1.73 ($\textcolor{green}{\downarrow}$) & 0.752 ($\textcolor{red}{\downarrow}$) & 0.600 ($\textcolor{green}{\uparrow}$) & 1.115 ($\textcolor{red}{\downarrow}$) & 0.835 ($\textcolor{red}{\downarrow}$) & 8.27 ($\textcolor{green}{\downarrow}$) & \textbf{0.692} ($\textcolor{green}{\uparrow}$) & 0.555 ($\textcolor{green}{\uparrow}$) & \textbf{1.286} ($\textcolor{green}{\uparrow}$) & \underline{0.860} ($\textcolor{green}{\uparrow}$) & 4.84 ($\textcolor{green}{\downarrow}$) & 0.615 ($\textcolor{green}{\uparrow}$) & 0.622 ($\textcolor{green}{\uparrow}$) & 0.920 ($\textcolor{green}{\uparrow}$) & 0.795 ($\textcolor{red}{\downarrow}$) \\
    Power-law CFG + CADS (Ours) & \textbf{1.52} ($\textcolor{green}{\downarrow}$) & 0.770 ($\textcolor{red}{\downarrow}$) & 0.622 ($\textcolor{green}{\uparrow}$) & \underline{0.862} ($\textcolor{green}{\uparrow}$) & \textbf{0.782} ($\textcolor{green}{\uparrow}$) & \textbf{1.52} ($\textcolor{green}{\downarrow}$) & 0.770 ($\textcolor{red}{\downarrow}$) & 0.622 ($\textcolor{green}{\uparrow}$) & 1.071 ($\textcolor{green}{\uparrow}$) & \textbf{0.876} ($\textcolor{green}{\uparrow}$) & \textbf{7.98} ($\textcolor{green}{\downarrow}$) & \underline{0.690} ($\textcolor{red}{\downarrow}$) & \textbf{0.573} ($\textcolor{green}{\uparrow}$) & \underline{1.279} ($\textcolor{green}{\uparrow}$) & \textbf{0.862} ($\textcolor{green}{\uparrow}$) & \textbf{4.71} ($\textcolor{green}{\downarrow}$) & \textbf{0.640} ($\textcolor{green}{\uparrow}$) & \underline{0.624} ($\textcolor{red}{\downarrow}$) & \textbf{0.924} ($\textcolor{green}{\uparrow}$) & \underline{0.804} ($\textcolor{green}{\uparrow}$) \\
    \bottomrule
    \label{tab_4_appx}
  \end{tabular}}
% \end{table}
\vspace{0.25cm}

% \begin{table}[ht]
\centering
\caption{\textbf{Ablation study:} Changing the number of sampling steps for Class-conditional: DiT ImageNet-1K 256x256}
\resizebox{\columnwidth}{!}{
\begin{tabular}{cc|cc|ccccc}
\toprule 
Version & Num. steps & $\mathbf{\alpha}$ & $\mathbf{\omega}$ & \textbf{FID} $(\downarrow)$& \textbf{IS} $(\uparrow)$ & \textbf{Precision} $(\uparrow)$ & \textbf{Recall} $(\uparrow)$ & \textbf{sFID} $(\downarrow)$ \\ 
\midrule
\multirow{5}{*}{Stand. CFG} & 50  & 0 & 1.5  & 3.33 & 259.88 & 0.8163 & 0.5474 & 7.406 \\ 
                            & 100 & 0 & 1.4  & 2.64 & 233.72 & 0.8027 & 0.5831 & 5.720 \\ 
                            & 150 & 0 & 1.3  & 2.38 & 233.52 & 0.8032 & 0.5936 & 5.462 \\ 
                            & 200 & 0 & 1.35 & 2.29 & 234.92 & 0.8031 & 0.5950 & 5.331 \\ 
                            & 250 & 0 & 1.5  & 2.27 & 278.30 & 0.8291 & 0.5840 & 4.601 \\ 
\hline
\multirow{5}{*}{Non-lin. CFG} & 50 & 0.6 & 4.35  & 3.03 & 284.55 & 0.8215 & 0.5757 & 7.110 \\ 
                            & 100 & 0.6  & 3.4  & 2.32 & 274.36 & 0.8199 & 0.6012 & 5.432 \\ 
                            & 150 & 0.6  & 3.4  & 2.19 & 274.39 & 0.8202 & 0.6071 & 5.512 \\ 
                            & 200 & 0.75 & 4.8  & 2.17 & 276.98 & 0.8204 & 0.5956 & 5.567 \\ 
                            & 250 & 0.75 & 4.85 & 2.05 & 279.90 & 0.8310 & 0.5950 & 4.670 \\ 
\bottomrule
\label{tab_5_appx}
\end{tabular}
}
% \end{table}
\vspace{0.25cm}

% \begin{table}[ht]
\centering
\caption{\textbf{Ablation study:} Changing the number of sampling steps for Class-conditional: EDM2-S ImageNet-1K 512x512}
\resizebox{\columnwidth}{!}{
\begin{tabular}{cc|ccc|ccc}
\toprule
Version & Num. Steps & $\mathbf{\alpha}$ & $\mathbf\omega$ & \textbf{FID} $(\downarrow)$ & $\mathbf{\alpha}$ & $\mathbf\omega$ & $\textbf{FID}_{\textbf{DINO}}$ $(\downarrow)$ \\
\midrule
\multirow{4}{*}{Stand. CFG} & 8  & 0 & 1.95 & 4.78 & 0 & 2.3  & 103.33\\
                            & 16 & 0 & 1.50 & 2.52 & 0 & 2.3  & 57.47\\
                            & 32 & 0 & 1.40 & 2.29 & 0 & 2.3  & 54.78\\
                            & 64 & 0 & 1.50 & 2.25 & 0 & 2.15 & 54.39 \\
\hline
\multirow{4}{*}{Non-lin. CFG} & 8  & 0.05 & 2.30 & 4.74  & -0.25 & 1.5 & 100.81  \\
                             & 16 & 0.25 & 2.30 & 2.32  & -0.05 & 2.15 & 56.92  \\
                             & 32 & 0.85 & 11.40 & 1.93 & 0.35  & 2.5 & 52.77  \\
                             & 64 & 0.85 & 11.30 & 1.89 & 0.35  & 2.1 & 52.56  \\
\bottomrule
\end{tabular}
}
\end{table}

\clearpage

\begin{table}[ht]
\centering
\caption{\textbf{Ablation study:} Changing the number of sampling steps for Diffusion text-to-image: MMDiT CC12m}
\resizebox{\textwidth}{!}{
\begin{tabular}{cc|cc|cccccccc}
\toprule
Version & Num. Steps & $\mathbf{\alpha}$ & $\mathbf{\omega}$ & \textbf{FID} $(\downarrow)$ & \textbf{Clip score} $(\uparrow)$ & \textbf{Coverage} $(\uparrow)$ & \textbf{Density} $(\uparrow)$ & \textbf{Precision} $(\uparrow)$ & \textbf{Recall} $(\uparrow)$ \\
\midrule
\multirow{4}{*}{Stand. CFG} & 20  & 0 & 1.75 & 8.98 & 22.581 & 0.8392 & 1.104 & 0.6623 & 0.5545 \\
                            & 35  & 0 & 1.75 & 8.79 & 22.532 & 0.8450 & 1.124 & 0.6717 & 0.5590 \\
                            & 50  & 0 & 1.55 & 8.58 & 22.111 & 0.8401 & 1.109 & 0.6612 & 0.5692 \\
                            & 100 & 0 & 1.75 & 8.38 & 22.298 & 0.8462 & 1.117 & 0.6765 & 0.5698 \\
\hline
\multirow{4}{*}{Non-lin. CFG} & 20  & 0.25 & 3.05 & 8.94 & 22.773 & 0.8424 & 1.114 & 0.6619 & 0.5495 \\
                             & 35  & 0.65 & 7.5 & 8.40 & 22.590 & 0.8491 & 1.126 & 0.6638 & 0.5582 \\
                             & 50  & 0.60 & 7.0 & 8.11 & 22.415 & 0.8503 & 1.128 & 0.6703 & 0.5532 \\
                             & 100 & 0.75 & 9.5 & 8.02 & 22.563 & 0.8472 & 1.115 & 0.6747 & 0.5723 \\
\bottomrule
\end{tabular}
}
% \end{table}
\vspace{0.25cm}
% \begin{table}[ht]
\centering
\caption{\textbf{Ablation study:} Changing the number of sampling steps for Diffusion text-to-image: MDTv2 ImageNet-1K 512x512}
\resizebox{\textwidth}{!}{\begin{tabular}{cc|cc|cccccccc}
\toprule
Version & Num. Steps & $\mathbf{\alpha}$ & $\mathbf{\omega}$ & \textbf{FID} $(\downarrow)$ & \textbf{Clip score} $(\uparrow)$ & \textbf{Coverage} $(\uparrow)$ & \textbf{Density} $(\uparrow)$ & \textbf{Precision} $(\uparrow)$ & \textbf{Recall} $(\uparrow)$ \\
\midrule 
\multirow{4}{*}{Stand. CFG} & 20 & 0 & 1.55 & 5.30  & 23.949 & 0.8218 & 1.167 & 0.7475 & 0.5133 \\
                            & 30 & 0 & 1.55 & 4.09  & 23.998 & 0.8292 & 1.233 & 0.7492 & 0.5264 \\
                            & 40 & 0 & 1.6  & 3.85  & 24.011 & 0.8311 & 1.178 & 0.7602 & 0.5294 \\
                            & 50 & 0 & 1.2  & 3.68  & 24.306 & 0.8318 & 1.150 & 0.7510 & 0.5989 \\
\hline 
\multirow{4}{*}{Non-lin. CFG} & 20 & 0.6 & 6.0 & 4.88 & 24.154 & 0.8251 & 1.236 & 0.7503 & 0.4916 \\
                             & 30 & 0.6 & 6.0 & 4.03 & 24.033 & 0.8344 & 1.205 & 0.7583 & 0.5332 \\
                             & 40 & 0.7 & 7.0 & 3.73 & 23.367 & 0.8353 & 1.181 & 0.7557 & 0.5546 \\
                             & 50 & 0.8 & 8.5 & 3.57 & 25.339 & 0.8361 & 1.170 & 0.7513 & 0.5609 \\
\bottomrule
\end{tabular}}
% \end{table}
\vspace{0.25cm}
% \begin{table}[ht]
\centering
\caption{\textbf{Ablation study:} Changing the number of sampling steps for Flow-Matching text-to-image: MMDiT on COCO}
\resizebox{\textwidth}{!}{\begin{tabular}{cc|cc|cccccccc}
\toprule  
Version & Num. Steps & $\mathbf{\alpha}$ & $\mathbf{\omega}$ & \textbf{FID} $(\downarrow)$ & \textbf{Clip score} $(\uparrow)$ & \textbf{Coverage} $(\uparrow)$ & \textbf{Density} $(\uparrow)$ & \textbf{Precision} $(\uparrow)$ & \textbf{Recall} $(\uparrow)$ \\
\midrule
\multirow{4}{*}{Stand. CFG} & 20 & 0 & 2.85 & 6.84 & 26.373 & 0.7529 & 0.8820 & 0.6121 & 0.5604\\
                            & 30 & 0 & 1.95 & 5.84 & 25.948 & 0.7581 & 0.8668 & 0.6051 & 0.5879 \\
                            & 40 & 0 & 2.05 & 5.62 & 25.817 & 0.7651 & 0.8798 & 0.6091 & 0.5978 \\
                            & 50 & 0 & 2.00   & 5.20 & 25.714 & 0.7726 & 0.9026 & 0.6299 & 0.5940\\
\hline 
\multirow{4}{*}{Non-lin. CFG} & 20 & 0.5 & 9.75 & 6.47 & 25.981 & 0.7241 & 0.7762 & 0.5719 & 0.5851 \\
                             & 30 & 0.5 & 9.45 & 5.62 & 26.003 & 0.7577 & 0.8457 & 0.6105 & 0.5874 \\
                             & 40 & 0.6 & 9.05 & 5.45 & 25.113 & 0.7633 & 0.8549 & 0.6149 & 0.6030 \\
                             & 50 & 0.7 & 10.15 & 4.81 & 25.848 & 0.7782 & 0.9183 & 0.6208 & 0.6191 \\
\bottomrule
\end{tabular}}
% \end{table}
\vspace{0.25cm}
% \begin{table}[ht]
\centering
\caption{\textbf{Ablation study:} Changing the number of sampling steps for Flow-Matching text-to-image: MMDiT on CC12m}
\resizebox{\textwidth}{!}{\begin{tabular}{cc|cc|cccccccc}
\toprule
Version & Num. Steps & $\mathbf{\alpha}$ & $\mathbf{\omega}$ & \textbf{FID} $(\downarrow)$ & \textbf{Clip score} $(\uparrow)$ & \textbf{Coverage} $(\uparrow)$ & \textbf{Density} $(\uparrow)$ & \textbf{Precision} $(\uparrow)$ & \textbf{Recall} $(\uparrow)$ \\
\midrule
\multirow{4}{*}{Stand. CFG} & 20 & 0 & 2.75 & 10.75 & 25.224 & 0.8289 & 1.069 & 0.6396 & 0.5803 \\
                            & 30 & 0 & 1.95 & 9.85  & 24.935 & 0.8318 & 1.068 & 0.6946 & 0.6000 \\
                            & 40 & 0 & 2.0  & 9.50  & 25.018 & 0.8461 & 1.103 & 0.7064 & 0.5907 \\
                            & 50 & 0 & 2.1  & 9.46  & 25.133 & 0.8520 & 1.145 & 0.7159 & 0.5946 \\
\hline
\multirow{4}{*}{Non-lin. CFG} & 20 & 0.2 & 3.25 & 10.68 & 25.585 & 0.8301 & 1.075 & 0.7101 & 0.5815 \\
                             & 30 & 0.5 & 10.0 & 9.81  & 25.002 & 0.8338 & 1.085 & 0.6968 & 0.5909 \\
                             & 40 & 0.6 & 9.35 & 9.17  & 24.794 & 0.8352 & 1.087 & 0.6909 & 0.6030 \\
                             & 50 & 0.6 & 8.05  & 9.00  & 24.723 & 0.8397 & 1.087 & 0.6911 & 0.6023 \\
\bottomrule
\label{tab_10_appx}
\end{tabular}}
\end{table}

\ifarxiv
    \begin{figure}[ht]
        \subsubsection{Sensitivity analysis}
        \label{sec:sens_anal}
        \centering
        % \vspace*{-2cm} % adjust this value to fit the figures on the page
        \begin{minipage}{.95\textwidth}
            \begin{subfigure}{0.448\textwidth}
                \includegraphics[width=\textwidth]{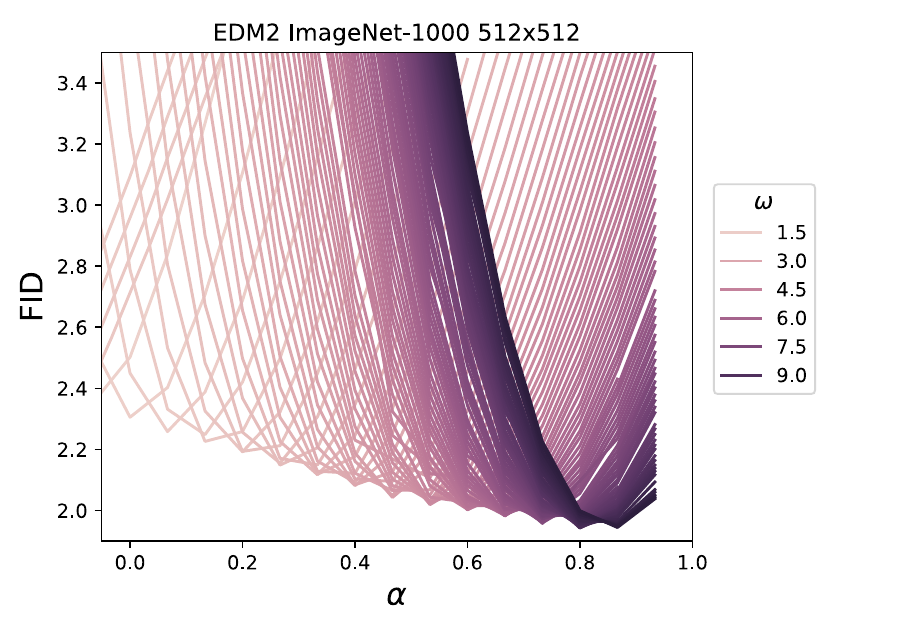}
            \end{subfigure}
            \hfill
            \begin{subfigure}{0.448\textwidth}
                \includegraphics[width=\textwidth]{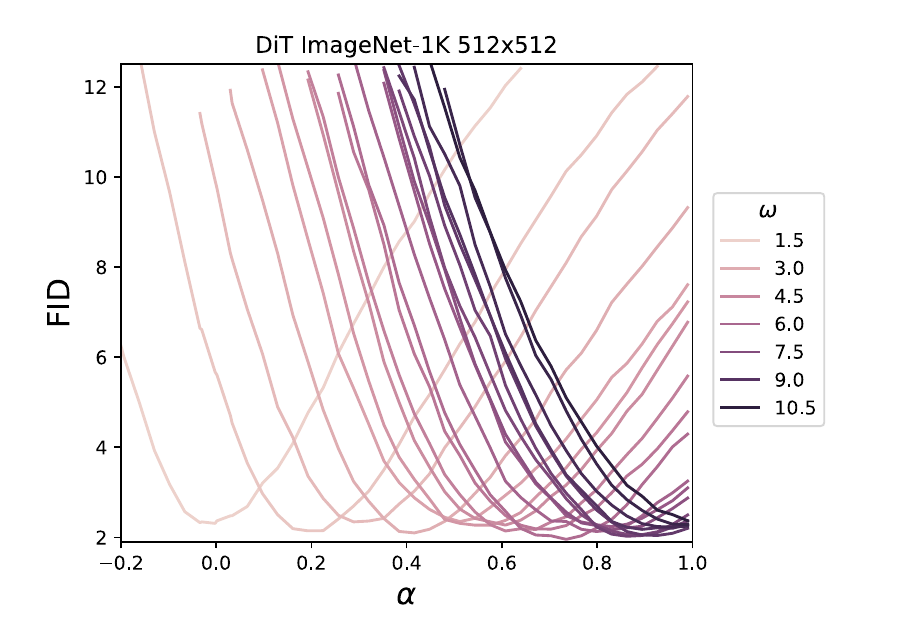}
            \end{subfigure}
            \caption{\textbf{Class-conditional diffusion}: image quality benefits from non-linear scheme, yielding lower FID for larger values of $\alpha$.}
        \label{fig:11_appx}
        \end{minipage}
        
        \vspace*{0.45cm} % adjust this value to separate the figures
        
        \begin{minipage}{.9\textwidth}
            \begin{subfigure}{0.448\textwidth}
                \includegraphics[width=\textwidth]{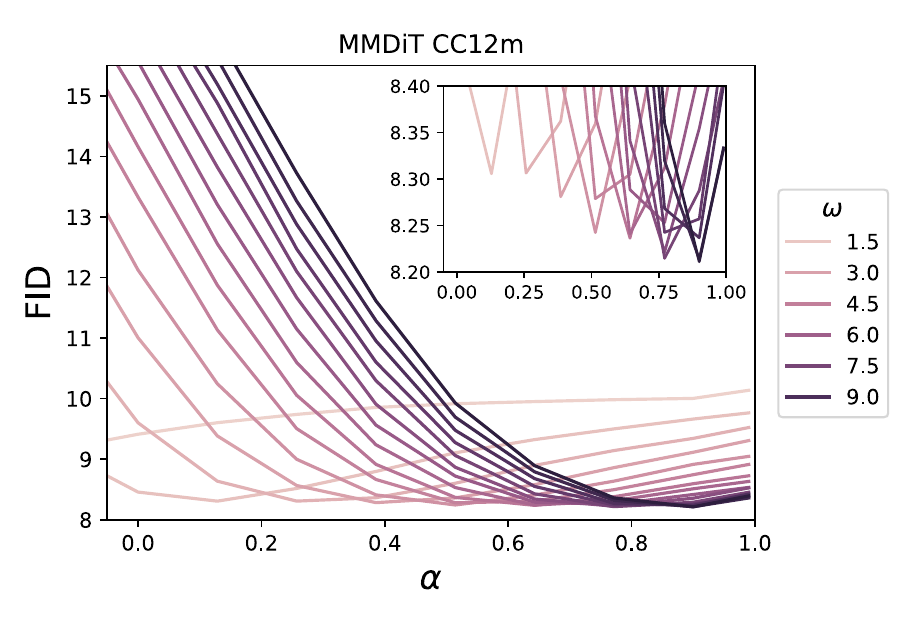}
            \end{subfigure}
            \hfill
            \begin{subfigure}{0.448\textwidth}
                \includegraphics[width=\textwidth]{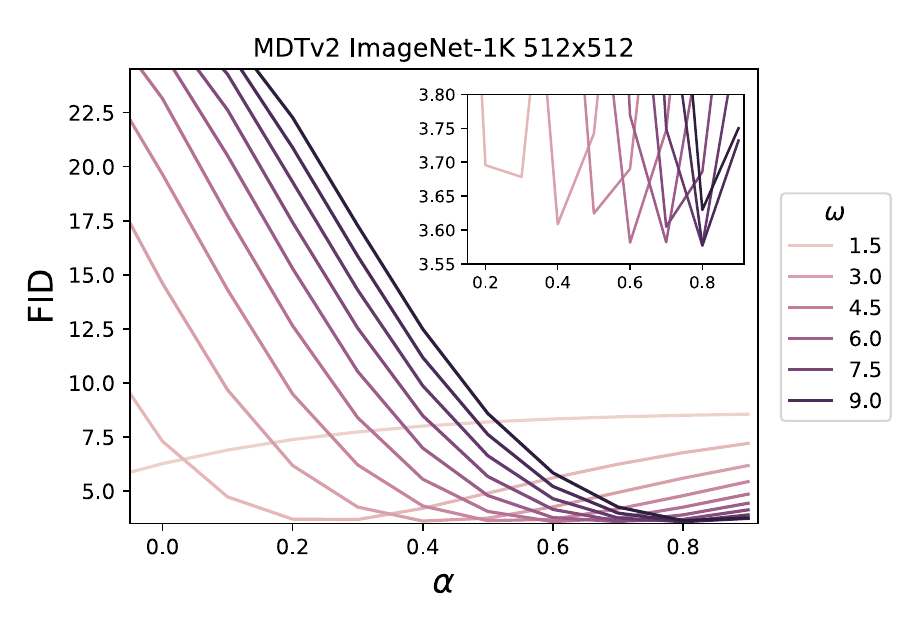}
            \end{subfigure}
            \caption{\textbf{Text-to-image diffusion models}: image quality benefits from non-linear scheme, yielding lower FID for larger values of $\alpha$.}        
            \label{fig:12_appx}
        \end{minipage}
        
        \vspace*{0.45cm} % adjust this value to separate the figures
        
        \begin{minipage}{.9\textwidth}
            \begin{subfigure}{0.448\textwidth}
                \includegraphics[width=\textwidth]{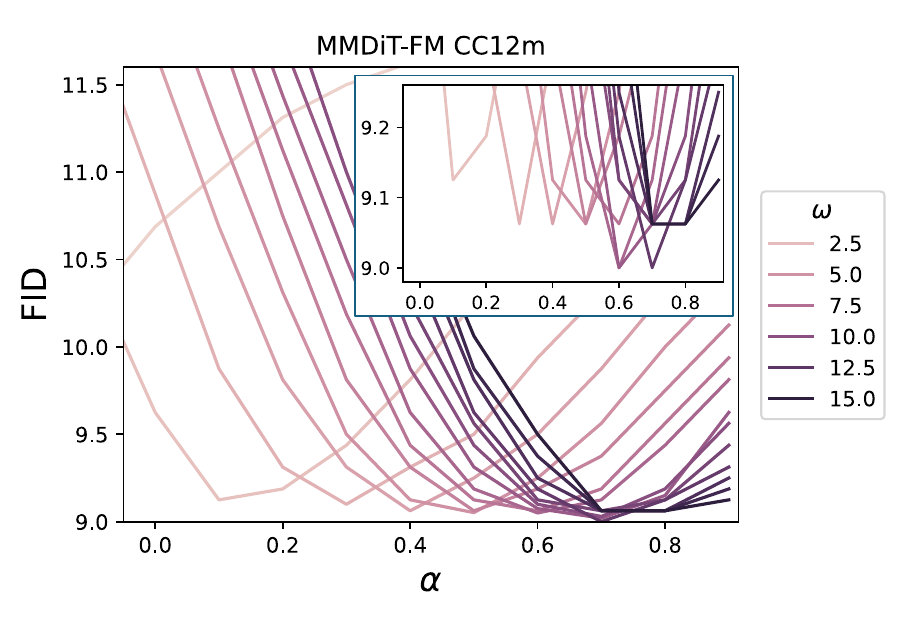}
            \end{subfigure}
            \hfill
            \begin{subfigure}{0.448\textwidth}
                \includegraphics[width=\textwidth]{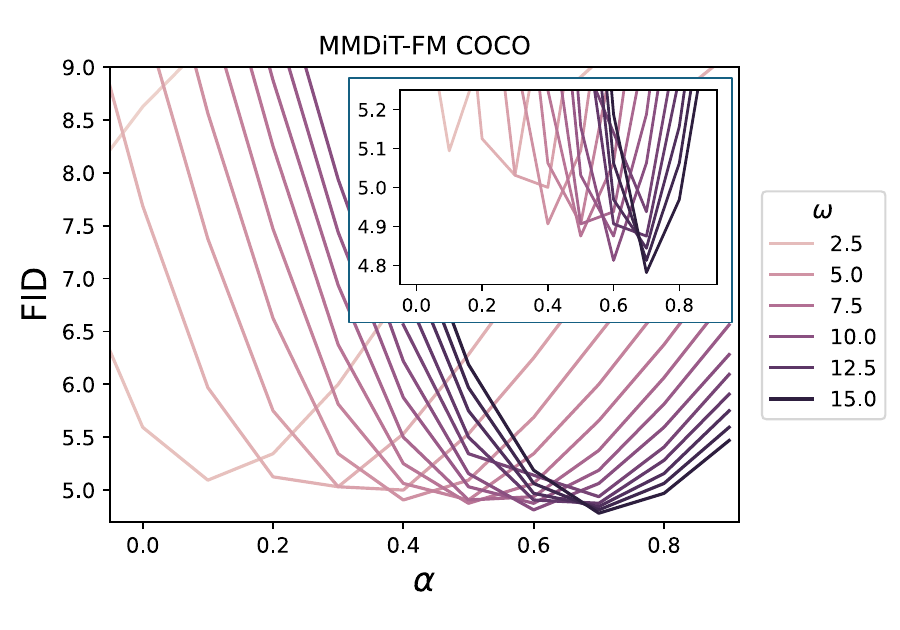}
            \end{subfigure}
            \caption{\textbf{Text-to-image flow matching}: image quality benefits from non-linear scheme, yielding lower FID for larger values of $\alpha$.}
        \label{fig:13_appx}
        \end{minipage}
        % \vspace*{-2cm} % adjust this value to fit the figures on the page
    \end{figure}
\else
    \begin{figure}[ht]
        \subsubsection{Sensitivity analysis}
        \label{sec:sens_anal}
        \centering
        % \vspace*{-2cm} % adjust this value to fit the figures on the page
        \begin{minipage}{1.\textwidth}
            \subfigure{
                \includegraphics[width=0.49\textwidth]{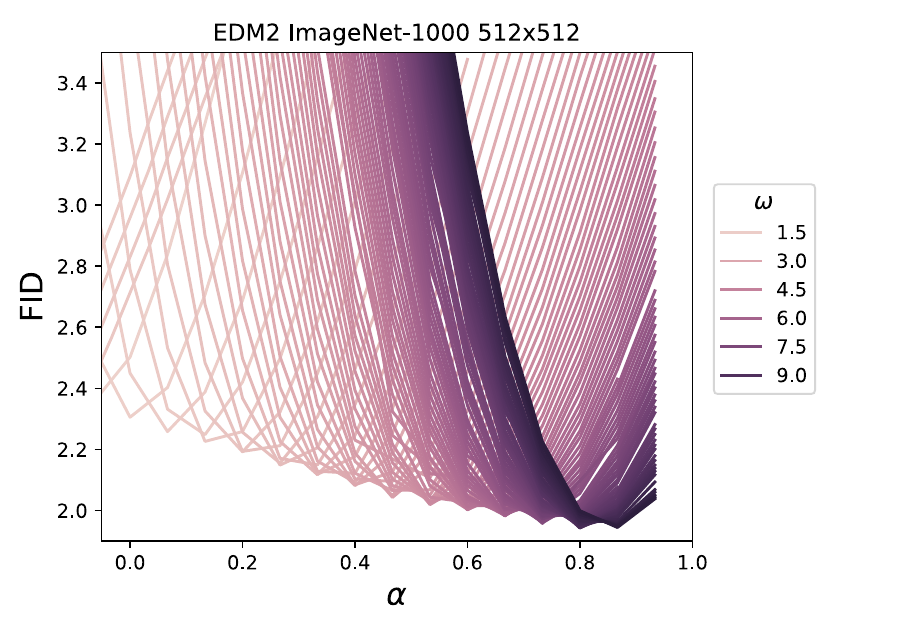}
                
            }
            \subfigure{
                \includegraphics[width=0.49\textwidth]{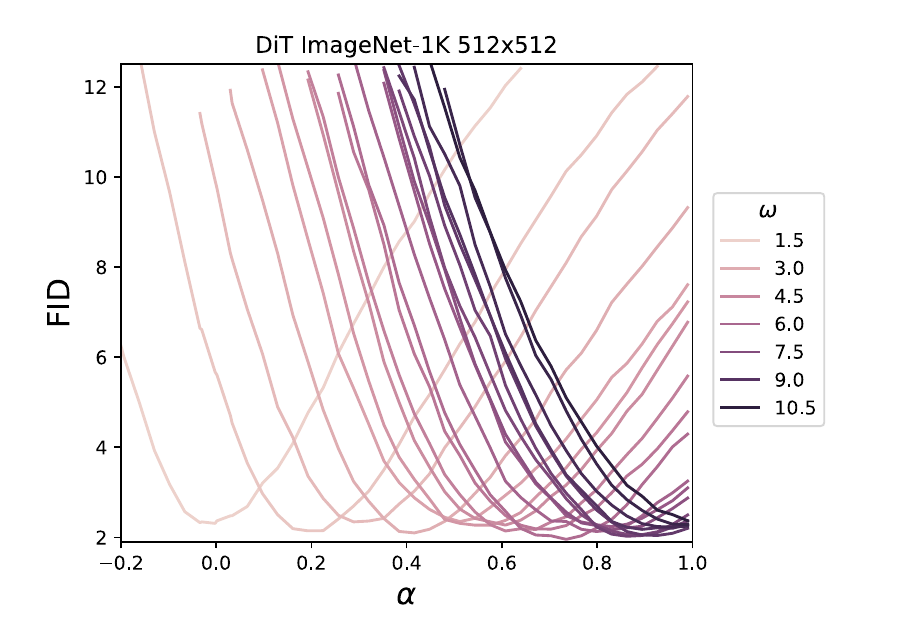}
             
            }
            \caption{\textbf{Class-conditional diffusion}: image quality benefits from non-linear scheme, yielding lower FID for larger values of $\alpha$.}
            \label{fig:11_appx}
        \end{minipage}
    
        \vspace*{0.5cm} % adjust this value to separate the figures
    
        \begin{minipage}{1.\textwidth}
            \subfigure{
                \includegraphics[width=0.49\textwidth]{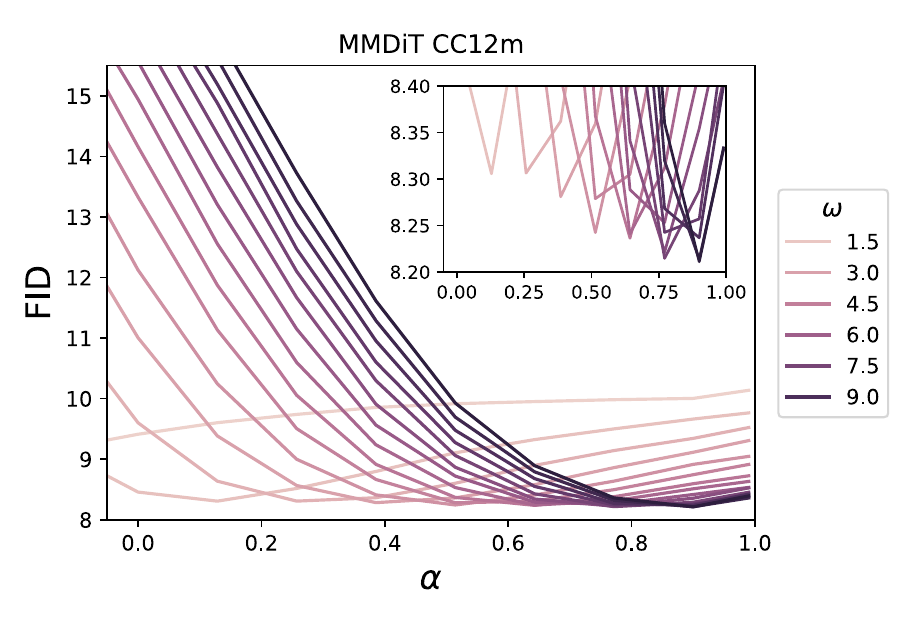}
            }
            \subfigure{
                \includegraphics[width=0.49\textwidth]{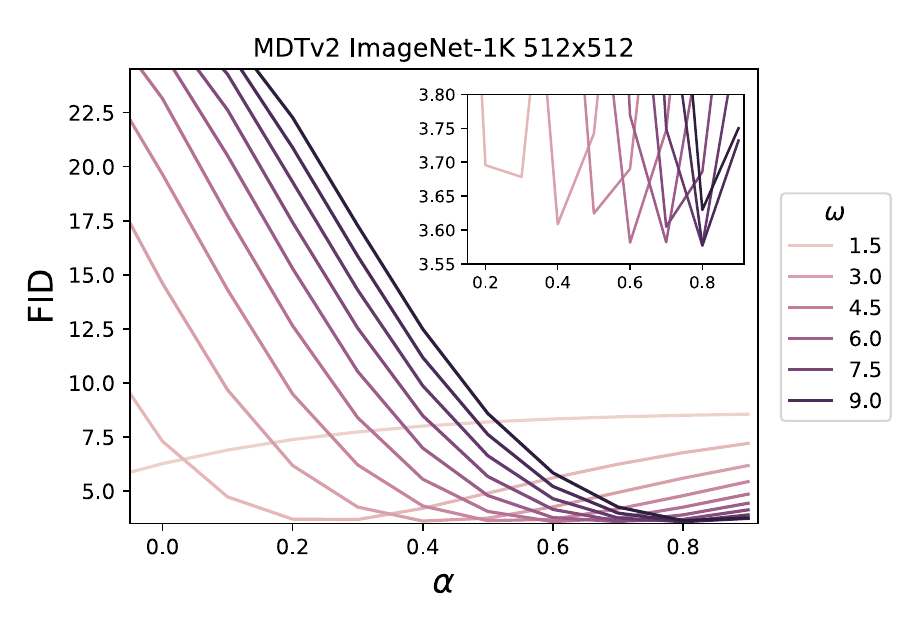}
            }
            \caption{\textbf{Text-to-image diffusion models}: image quality benefits from non-linear scheme, yielding lower FID for larger values of $\alpha$.}        
            \label{fig:12_appx}

        \end{minipage}
    
        \vspace*{0.5cm} % adjust this value to separate the figures

        \begin{minipage}{1.\textwidth}
            \subfigure{
                \includegraphics[width=0.49\textwidth]{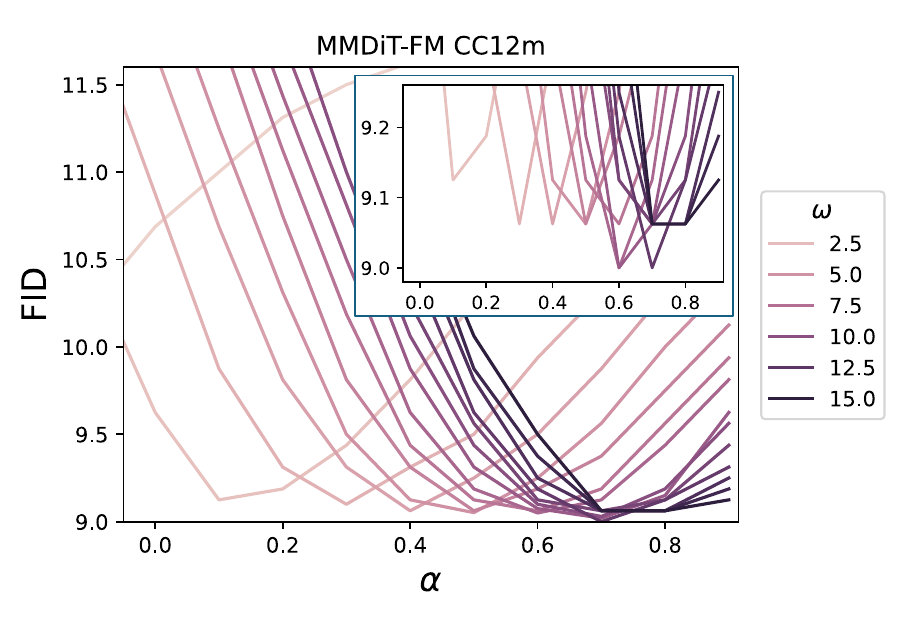}
            }
            \subfigure{
                \includegraphics[width=0.49\textwidth]{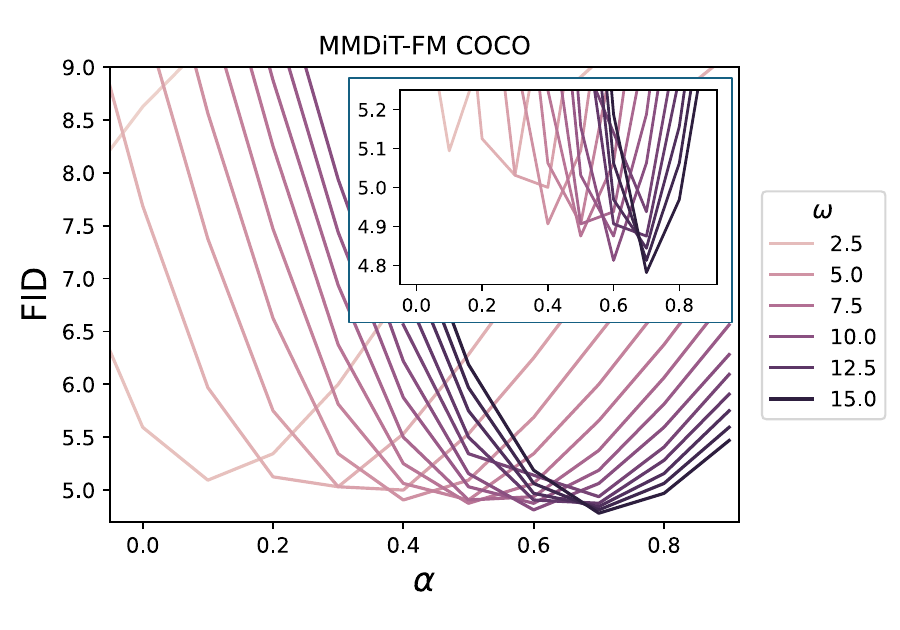}
            }
            \caption{\textbf{Text-to-image flow matching}: image quality benefits from non-linear scheme, yielding lower FID for larger values of $\alpha$.}
        \label{fig:13_appx}
        \end{minipage}
        % \vspace*{-2cm} % adjust this value to fit the figures on the page
    \end{figure}
\fi

\clearpage    

\clearpage

\subsubsection[Qualitative analysis: fixed non-linear parameter]{Qualitative analysis: varying $\omega$, fixed $\alpha$.}
\label{sec:qualit_anal_fix_alpha}

\ifarxiv
    \begin{figure}[h]
        \centering
        \begin{subfigure}[t]{0.48\textwidth} % Adjust width to fit two columns
            \centering
            \includegraphics[width=\linewidth]{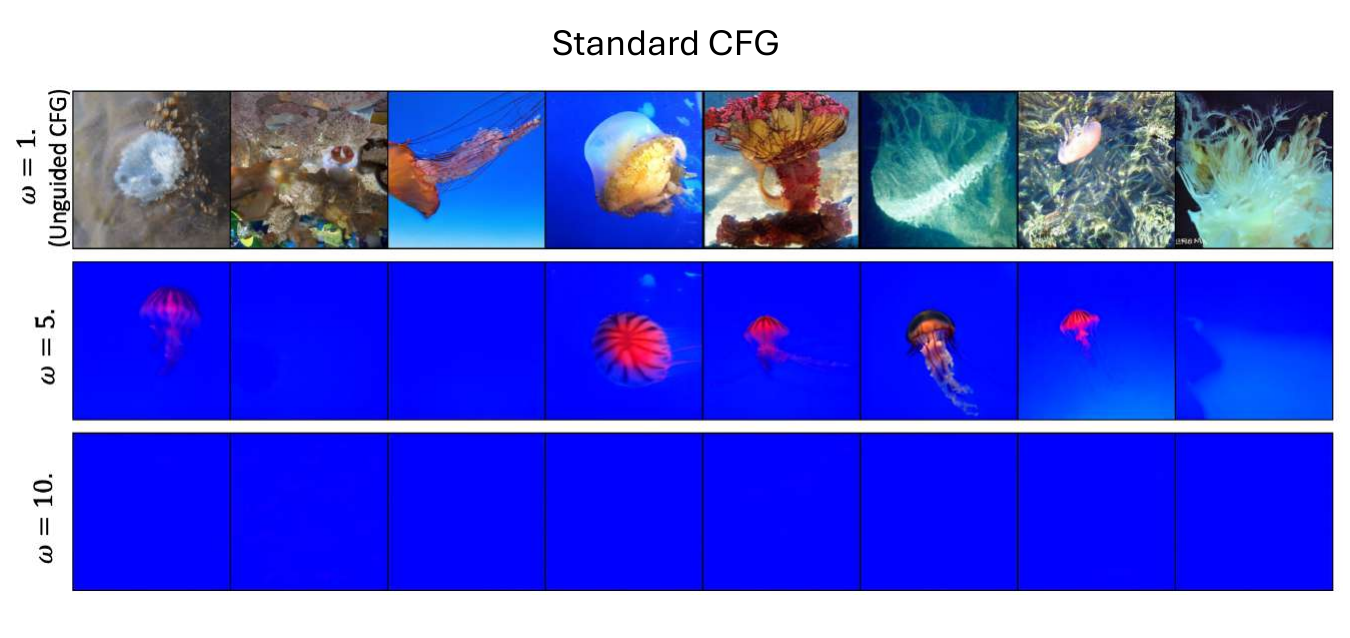}
            \caption{Class \textit{107: jellyfish} with $\alpha=0.$}
        \end{subfigure}%
        \hfill
        \begin{subfigure}[t]{0.48\textwidth} % Adjust width to fit two columns
            \centering
            \includegraphics[width=\linewidth]{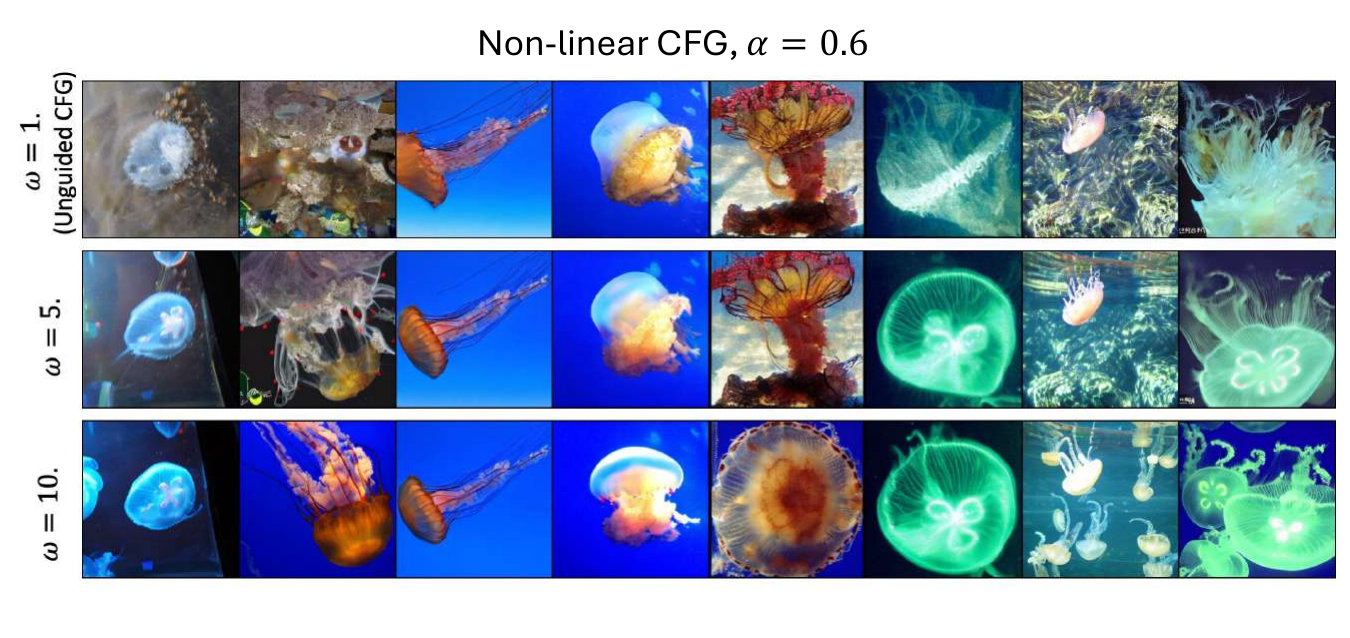}
            \caption{Class \textit{107: jellyfish} with $\alpha=0.9$}
        \end{subfigure}
        \vspace{0.3cm} % Adjust vertical space
        \begin{subfigure}[t]{0.48\textwidth} % Adjust width to fit two columns
            \centering
            \includegraphics[width=\linewidth]{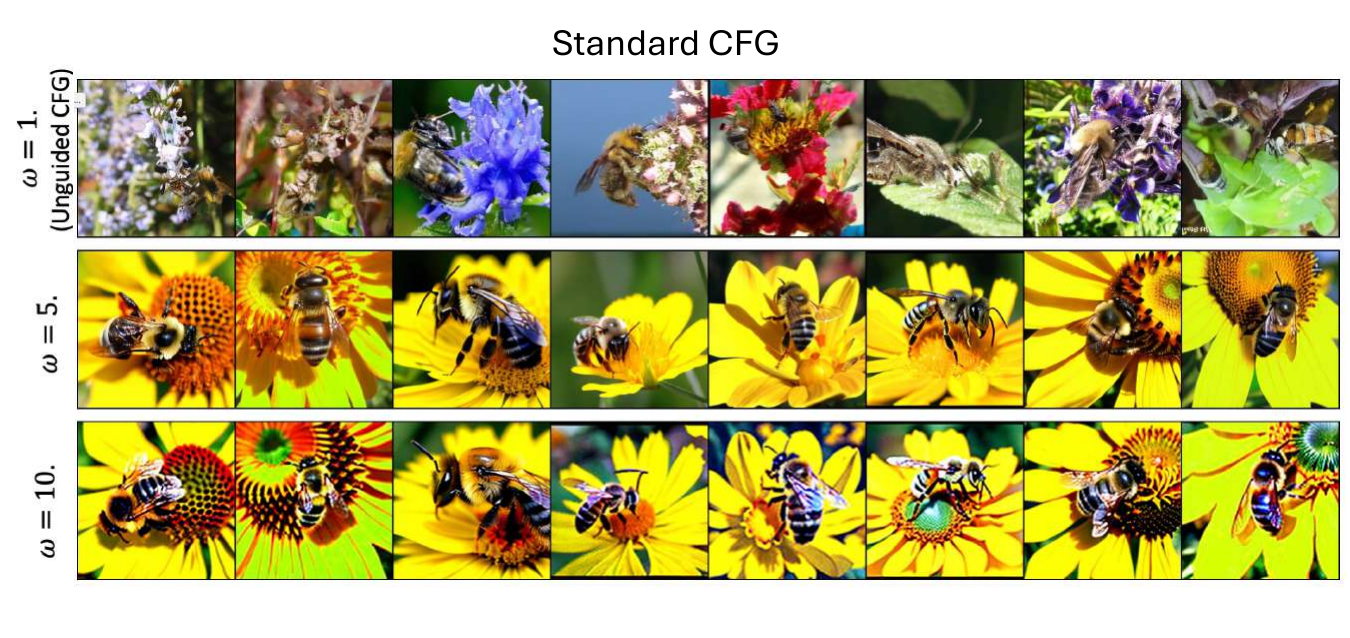}
            \caption{Class \textit{309: bee} with $\alpha=0.$}
        \end{subfigure}%
        \hfill
        \begin{subfigure}[t]{0.48\textwidth} % Adjust width to fit two columns
            \centering
            \includegraphics[width=\linewidth]{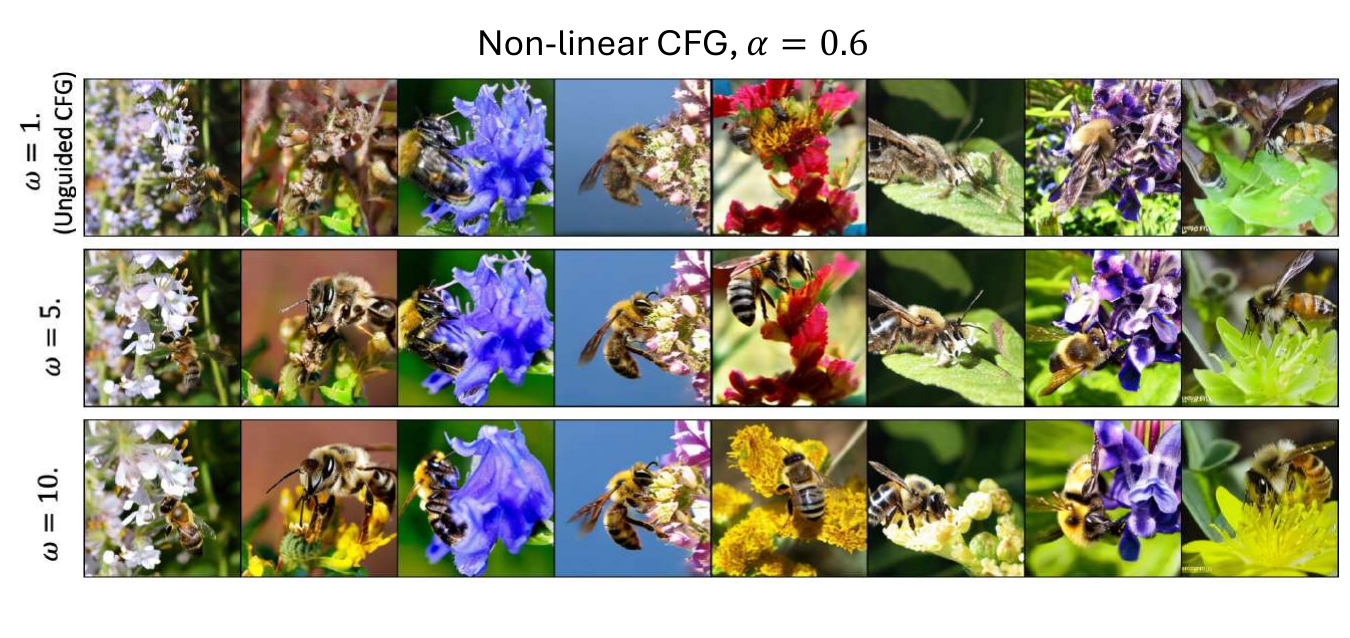}
            \caption{Class \textit{309: bee} with $\alpha=0.9$}
        \end{subfigure}
        \vspace{0.3cm} % Adjust vertical space
        \begin{subfigure}[t]{0.48\textwidth} % Adjust width to fit two columns
            \centering
            \includegraphics[width=\linewidth]{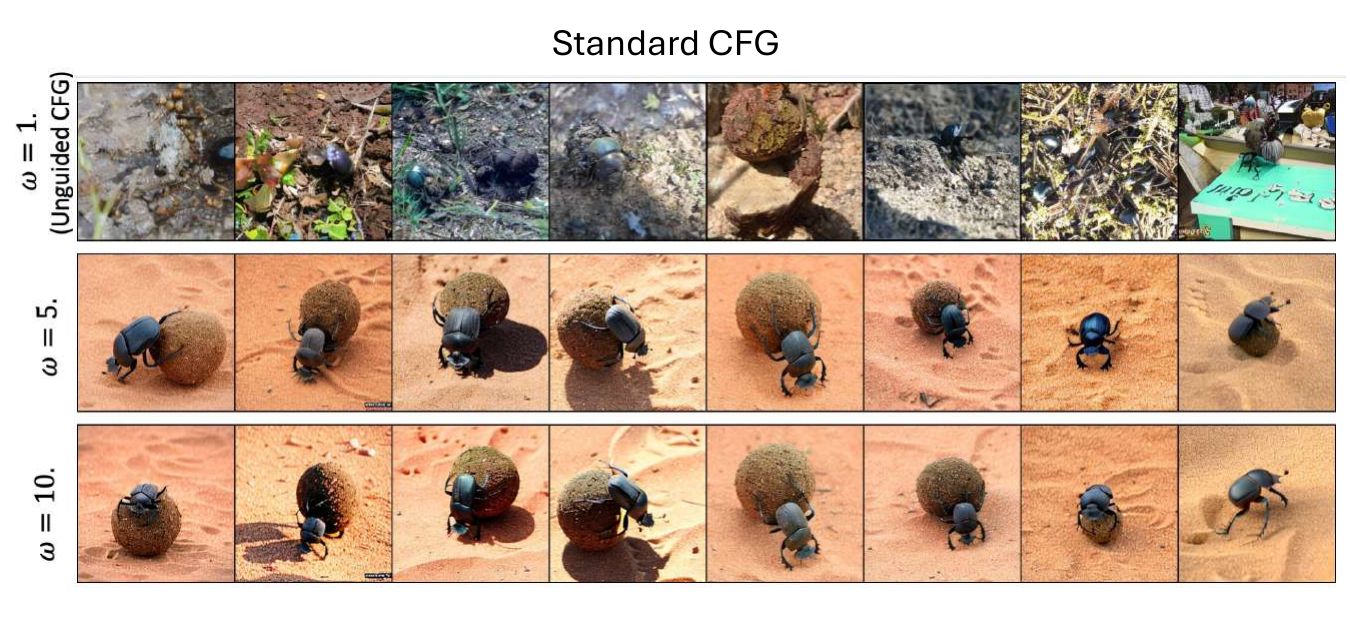}
            \caption{Class \textit{305: dung beetle} with $\alpha=0.$}
        \end{subfigure}%
        \hfill
        \begin{subfigure}[t]{0.48\textwidth} % Adjust width to fit two columns
            \centering
            \includegraphics[width=\linewidth]{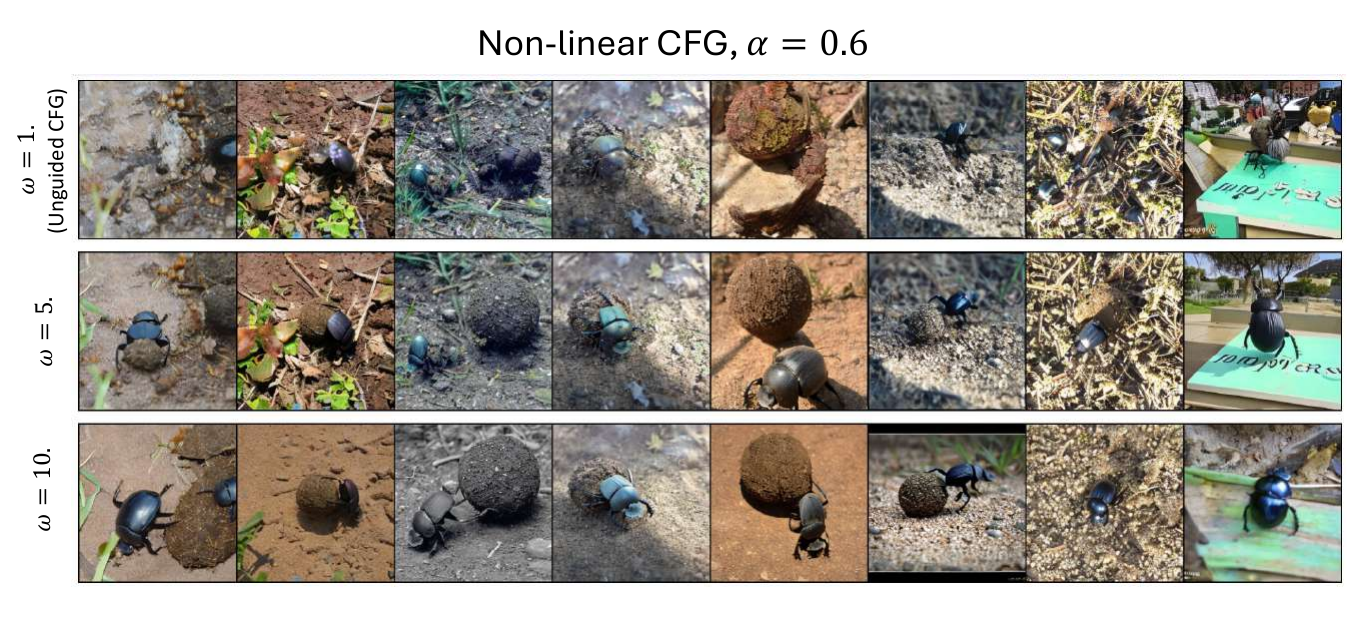}
            \caption{Class \textit{305: dung beetle} with $\alpha=0.9$}
        \end{subfigure}
        \caption{Generated images for different classes with varying values of $\omega$ and $\alpha$. Each panel shows the effect of changing $\alpha$ from 0 to 0.9, demonstrating the impact on diversity and image quality.}
    \end{figure}
\else
    \begin{figure}[h]
        \centering
        \subfigure[Class \textit{107: jellyfish} with $\alpha=0.$]{
            \includegraphics[width=0.48\textwidth]{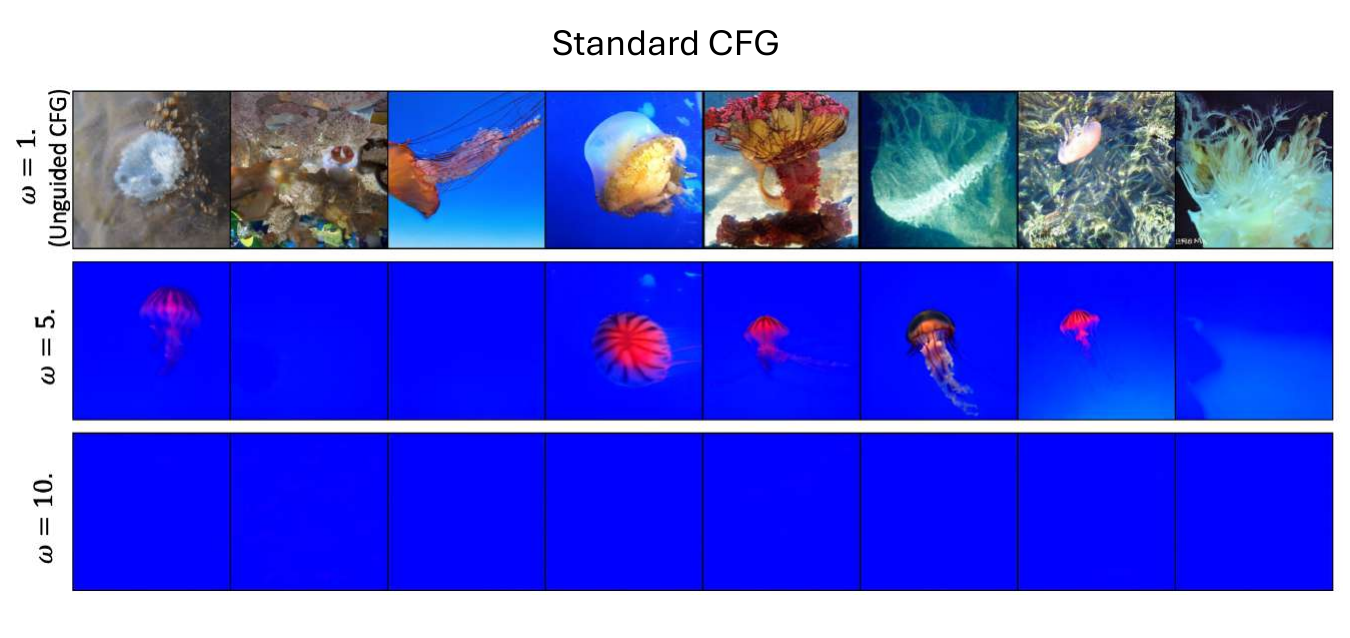}
        }%
        \hfill
        \subfigure[Class \textit{107: jellyfish} with $\alpha=0.9$]{
            \includegraphics[width=0.48\textwidth]{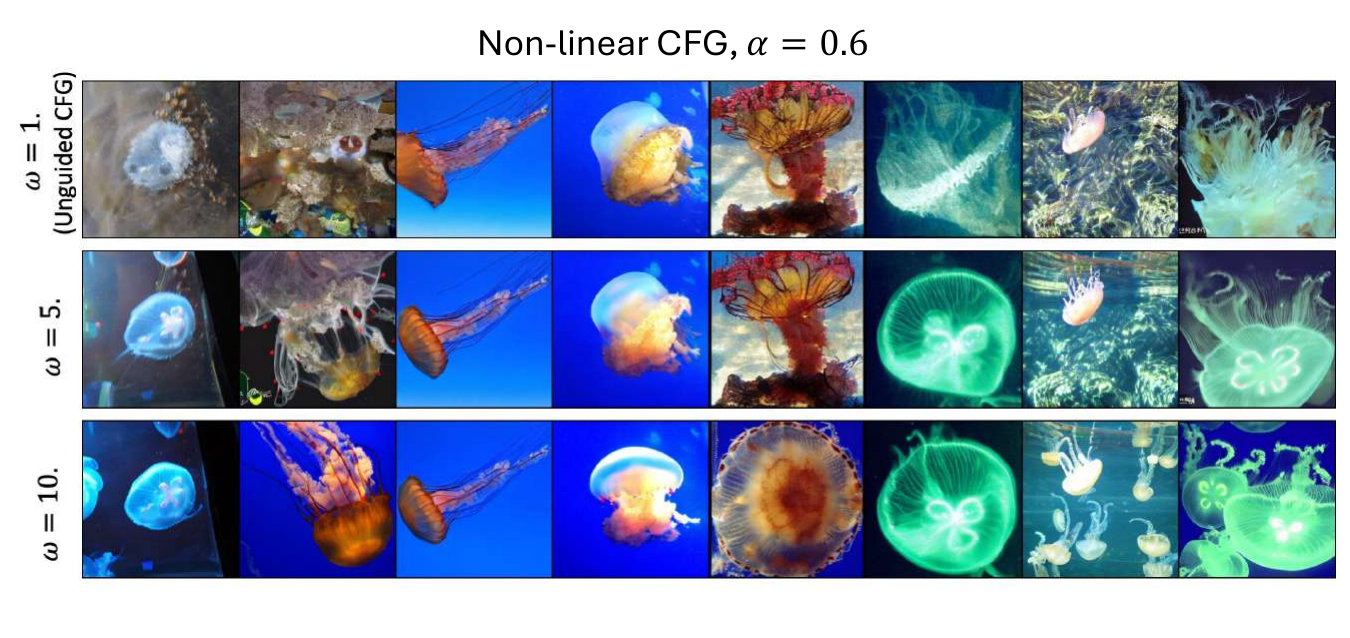}
        }
        \vspace{0.3cm} % Adjust vertical space
        \subfigure[Class \textit{309: bee} with $\alpha=0.$]{
            \includegraphics[width=0.48\textwidth]{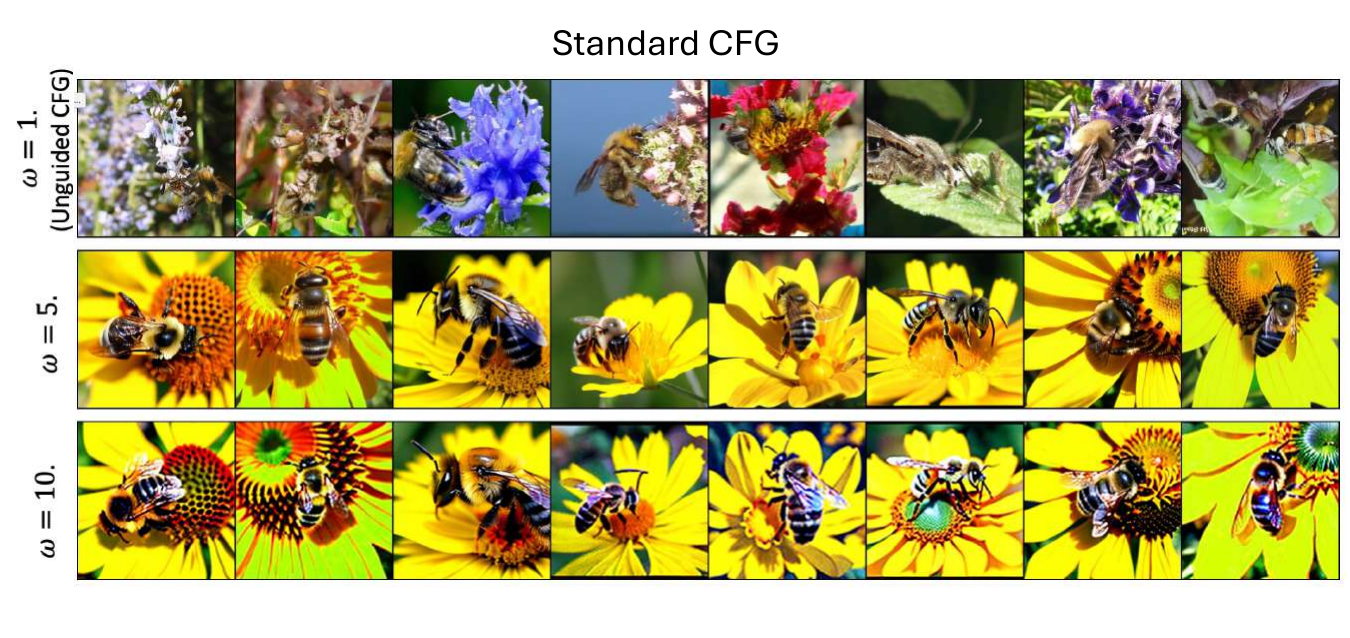}
        }%
        \hfill
        \subfigure[Class \textit{309: bee} with $\alpha=0.9$]{
            \includegraphics[width=0.48\textwidth]{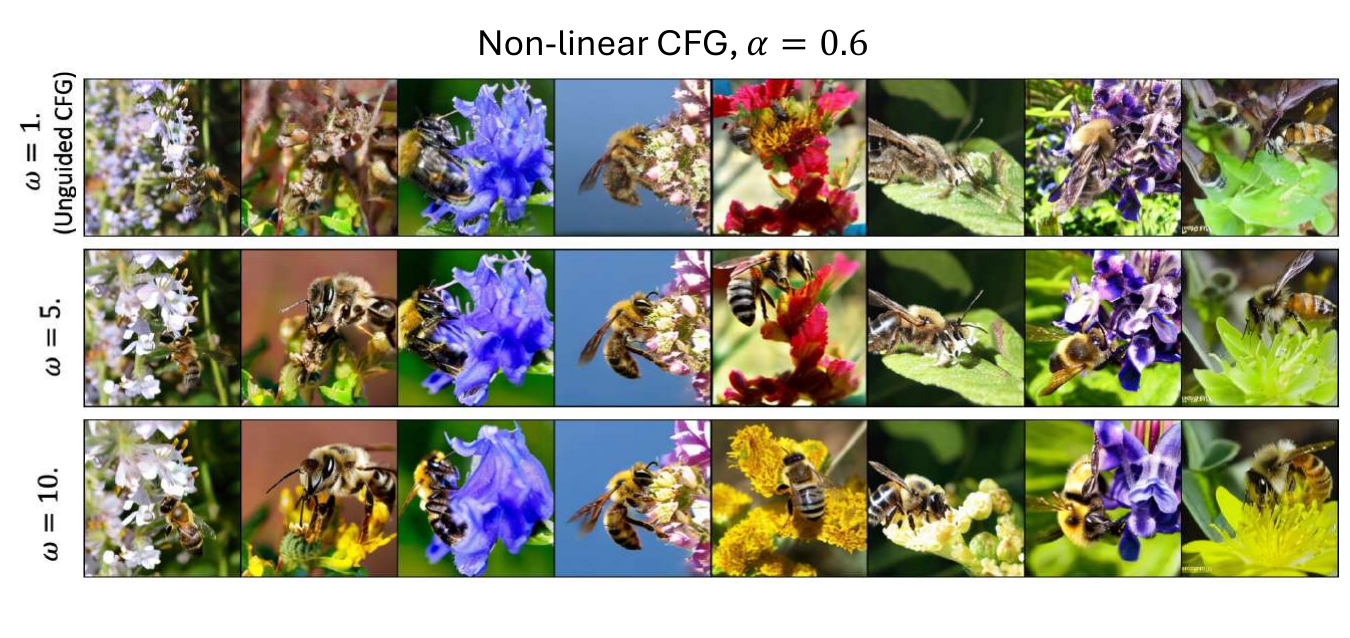}
        }
        \vspace{0.3cm} % Adjust vertical space
        \subfigure[Class \textit{305: dung beetle} with $\alpha=0.$]{
            \includegraphics[width=0.48\textwidth]{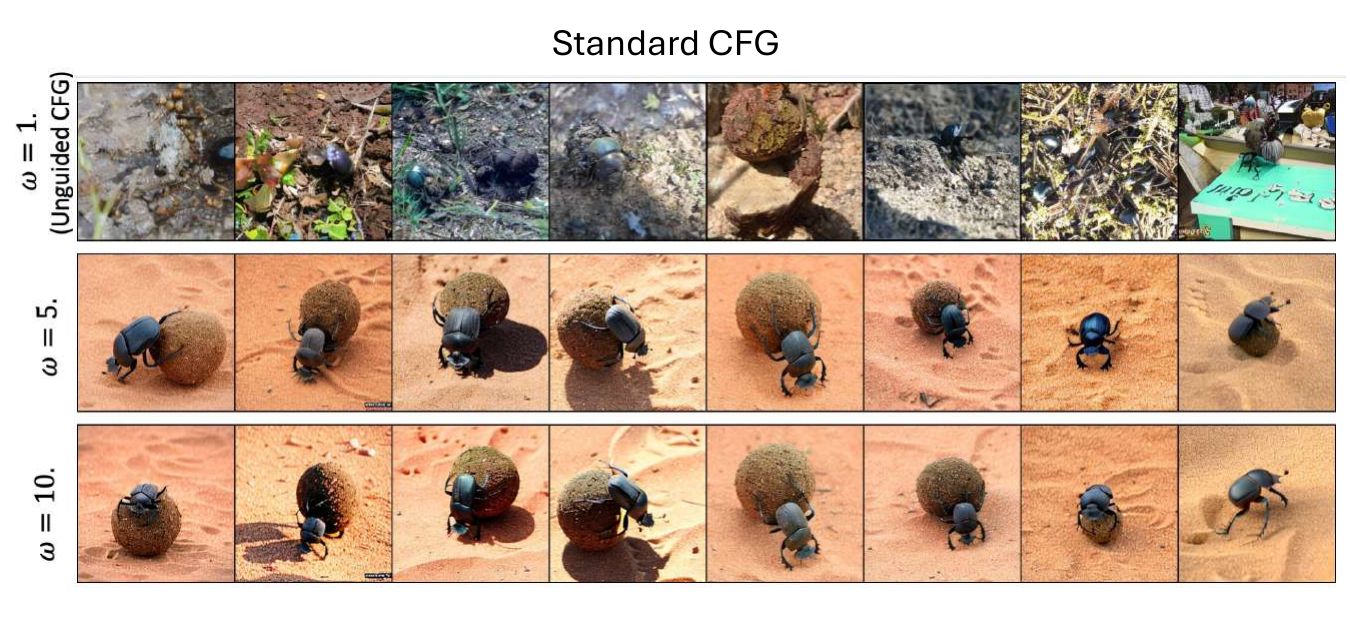}
        }%
        \hfill
        \subfigure[Class \textit{305: dung beetle} with $\alpha=0.9$]{
            \includegraphics[width=0.48\textwidth]{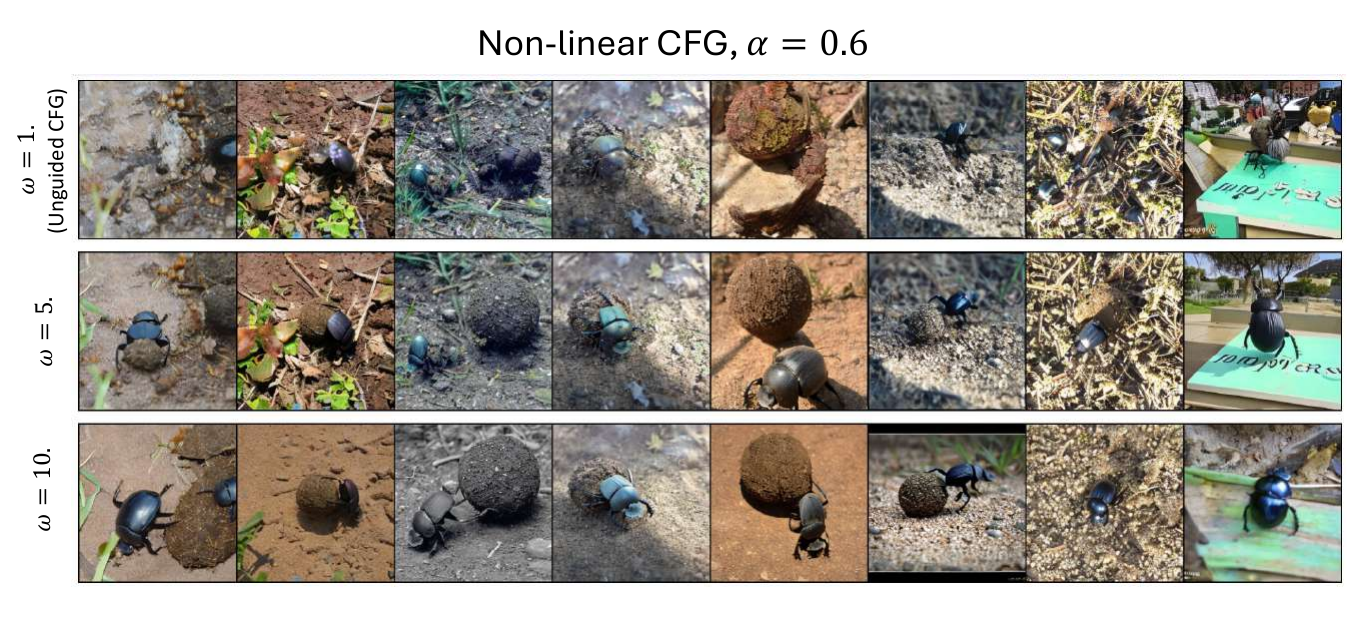}
        }
        \caption{Generated images for different classes for varying values of $\omega$ and $\alpha$. Each panel shows the effect of changing $\alpha$ from 0 to 0.9, demonstrating the impact on diversity and image quality.}
    \end{figure}
\fi

\clearpage

\subsubsection[ualitative analysis: fixed guidance scale]{Qualitative analysis: fixed $\omega$, varying $\alpha$.}
\label{sec:qualit_anal_fix_omega}

\ifarxiv
    \begin{figure}[h]
        \centering
        \begin{subfigure}[t]{0.4\textwidth} % Increase width to 0.48
            \centering
            \includegraphics[width=\linewidth]{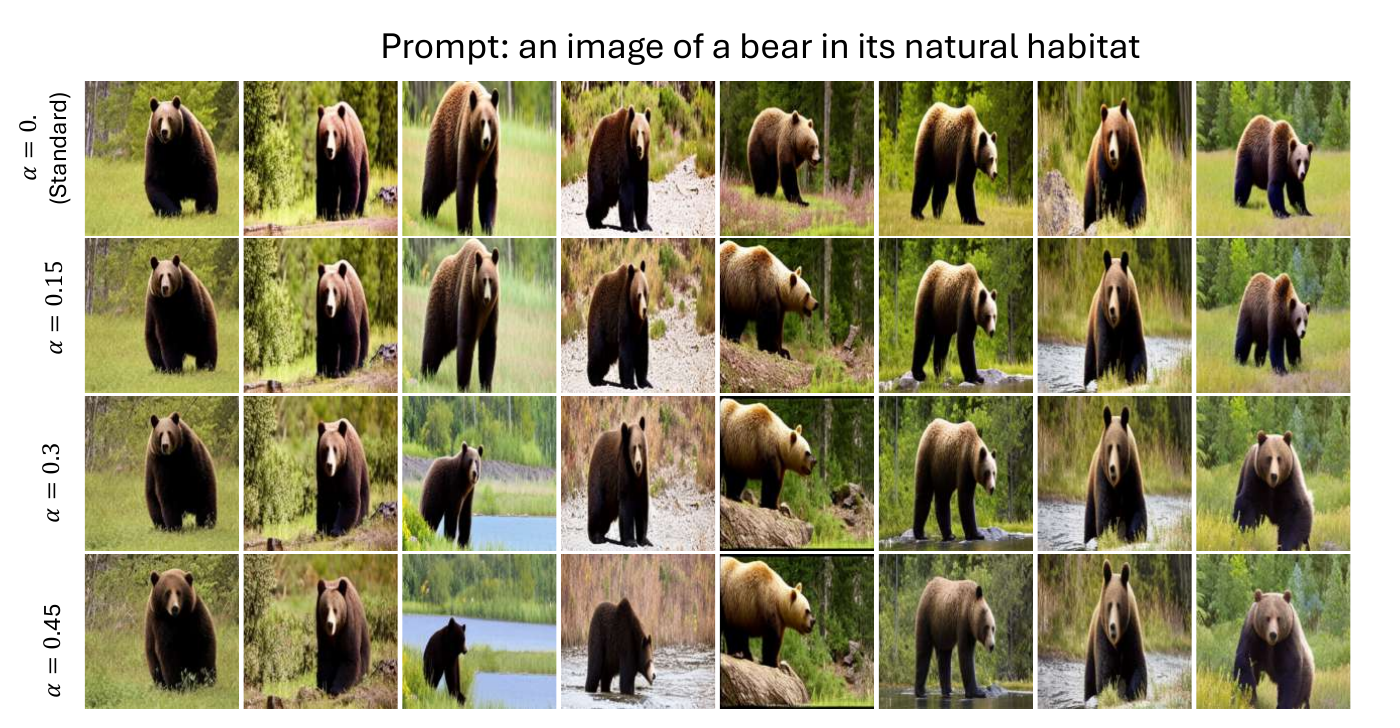}
            \caption{Prompt \textit{bear in its natural habitat}}
        \end{subfigure}%
        \hfill
        \begin{subfigure}[t]{0.4\textwidth} % Increase width to 0.48
            \centering
            \includegraphics[width=\linewidth]{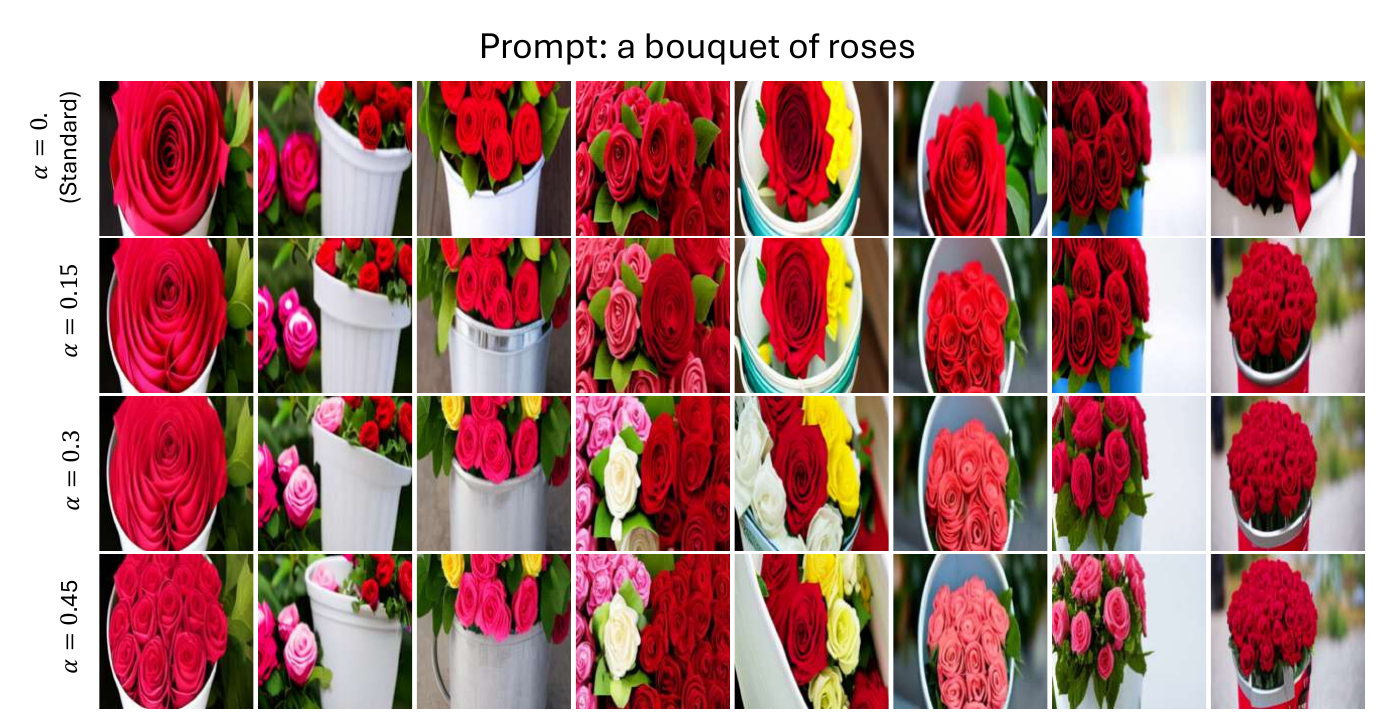}
            \caption{Prompt \textit{a bouquet of roses}}
        \end{subfigure}
        \vspace{0.3cm} % Reduce vertical space
        \begin{subfigure}[t]{0.4\textwidth} % Increase width to 0.48
            \centering
            \includegraphics[width=\linewidth]{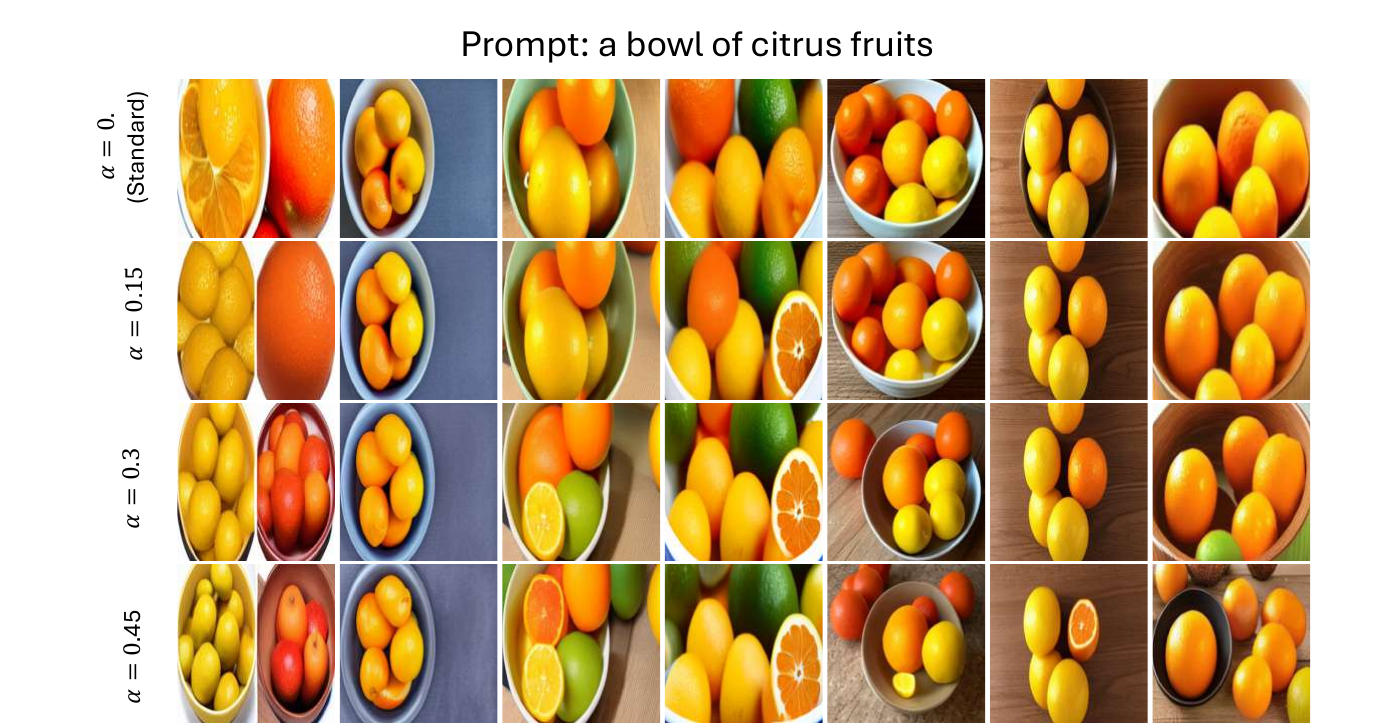}
            \caption{Prompt \textit{a bowl of citrus fruit}}
        \end{subfigure}%
        \hfill
        \begin{subfigure}[t]{0.4\textwidth} % Increase width to 0.48
            \centering
            \includegraphics[width=\linewidth]{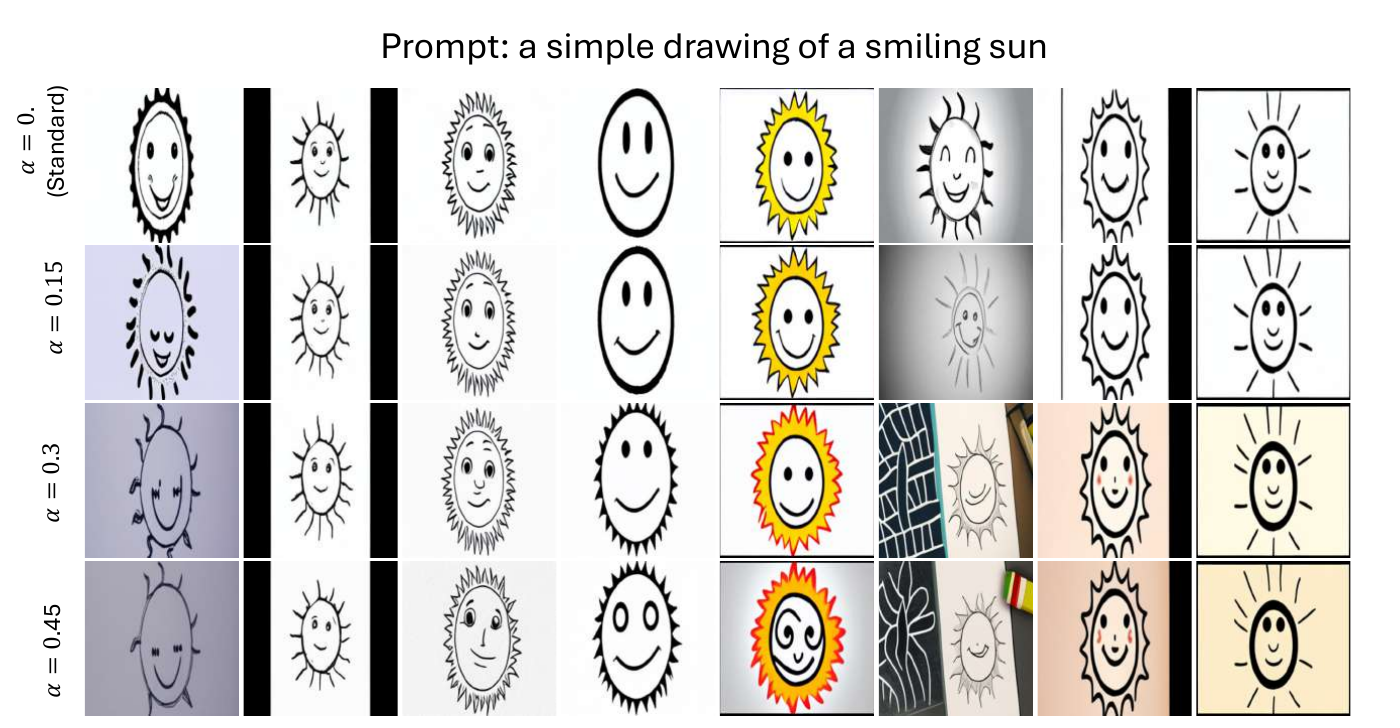}
            \caption{Prompt \textit{a simple drawing of a smiling sun}}
        \end{subfigure}
        \vspace{-.35cm}
        \caption{T2IM generated images for different prompts with $\omega=4.$ and varying value of $\alpha$.}
        \vspace{-.4cm}
    \end{figure}
    
    \begin{figure}[h]
        \centering
        \begin{subfigure}[t]{0.4\textwidth} % Increase width to 0.48
            \centering
            \includegraphics[width=\linewidth]{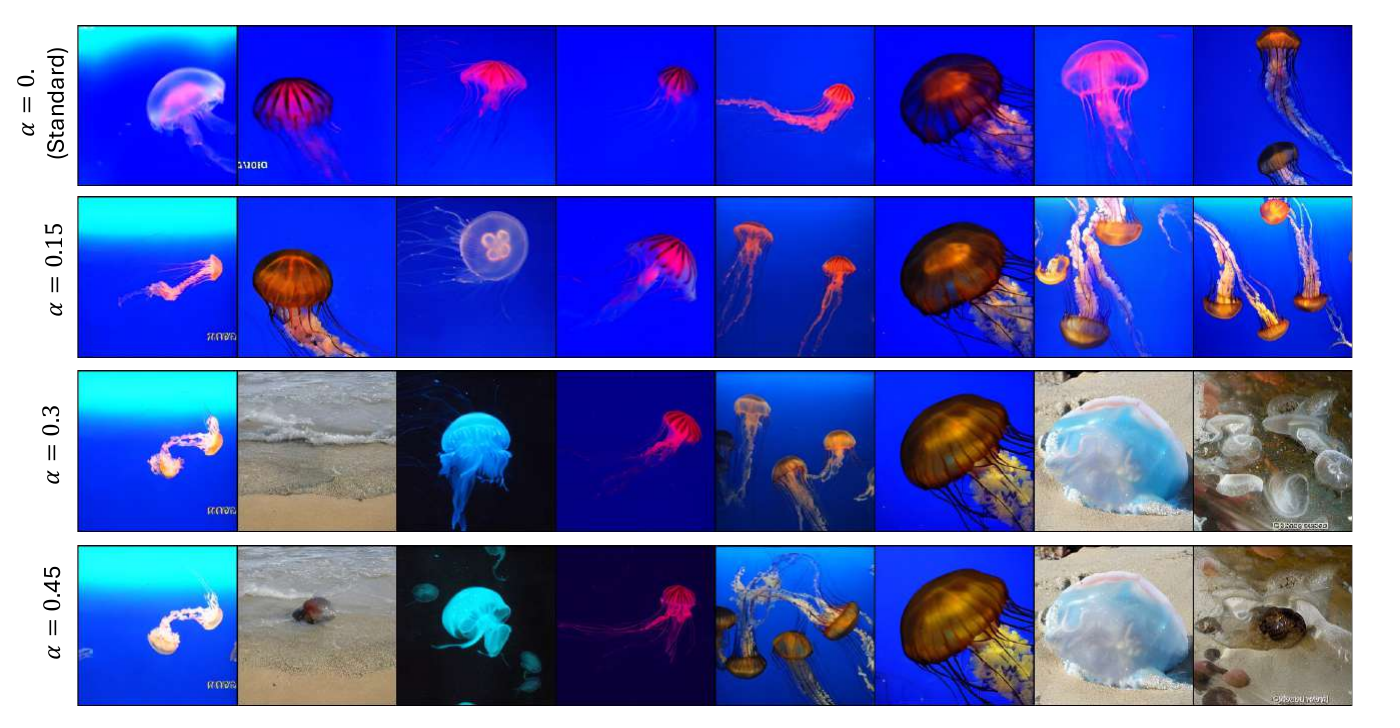}
            \caption{Class \textit{107: jellyfish}}
        \end{subfigure}%
        \hfill
        \begin{subfigure}[t]{0.4\textwidth} % Increase width to 0.48
            \centering
            \includegraphics[width=\linewidth]{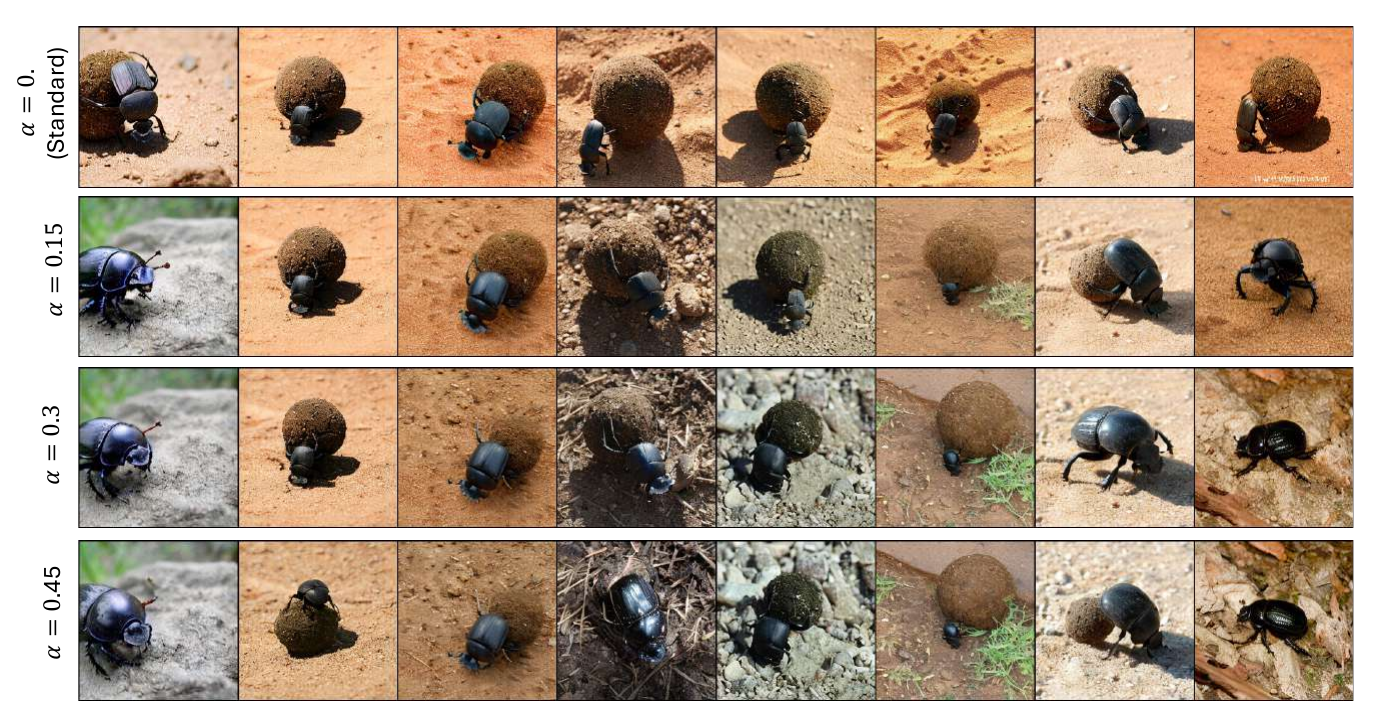}
            \caption{Class \textit{305: dung beetle}}
        \end{subfigure}
        \vspace{0.3cm} % Reduce vertical space
        \begin{subfigure}[t]{0.4\textwidth} % Increase width to 0.48
            \centering
            \includegraphics[width=\linewidth]{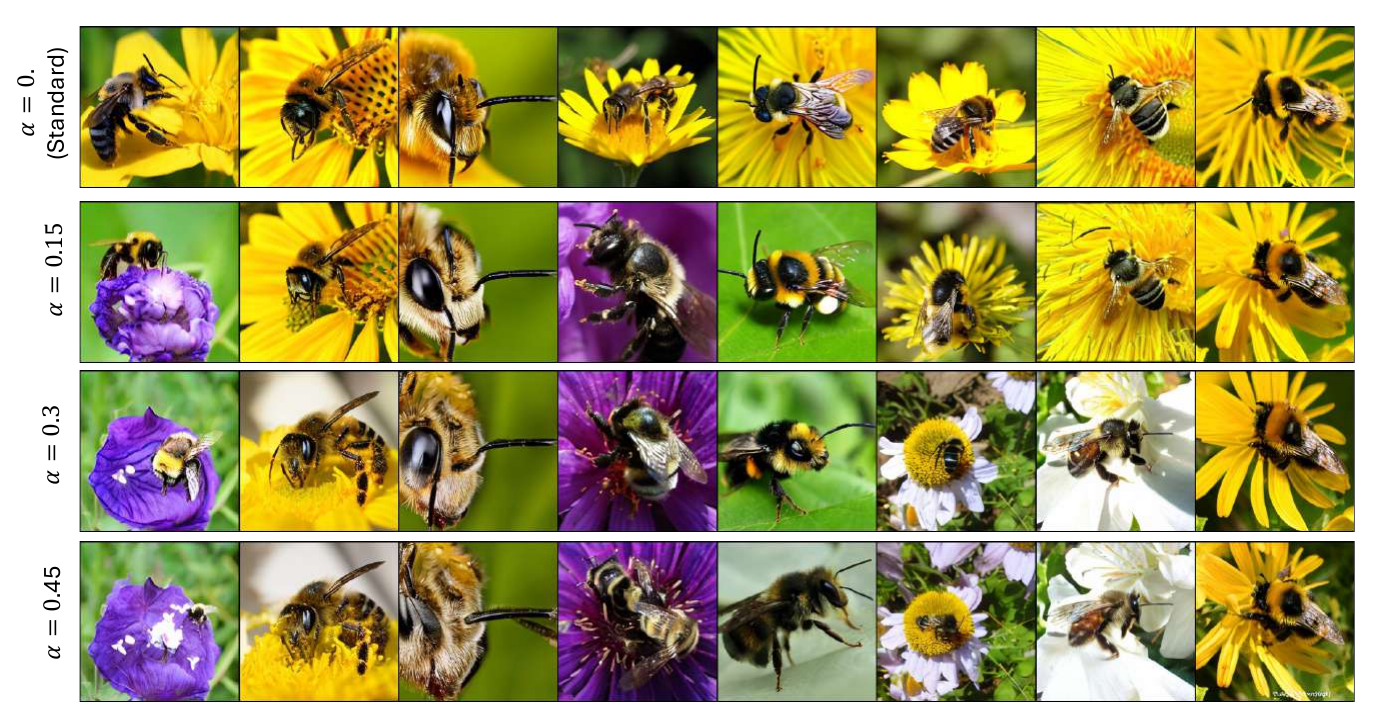}
            \caption{Class \textit{309: bee}}
        \end{subfigure}%
        \hfill
        \begin{subfigure}[t]{0.4\textwidth} % Increase width to 0.48
            \centering
            \includegraphics[width=\linewidth]{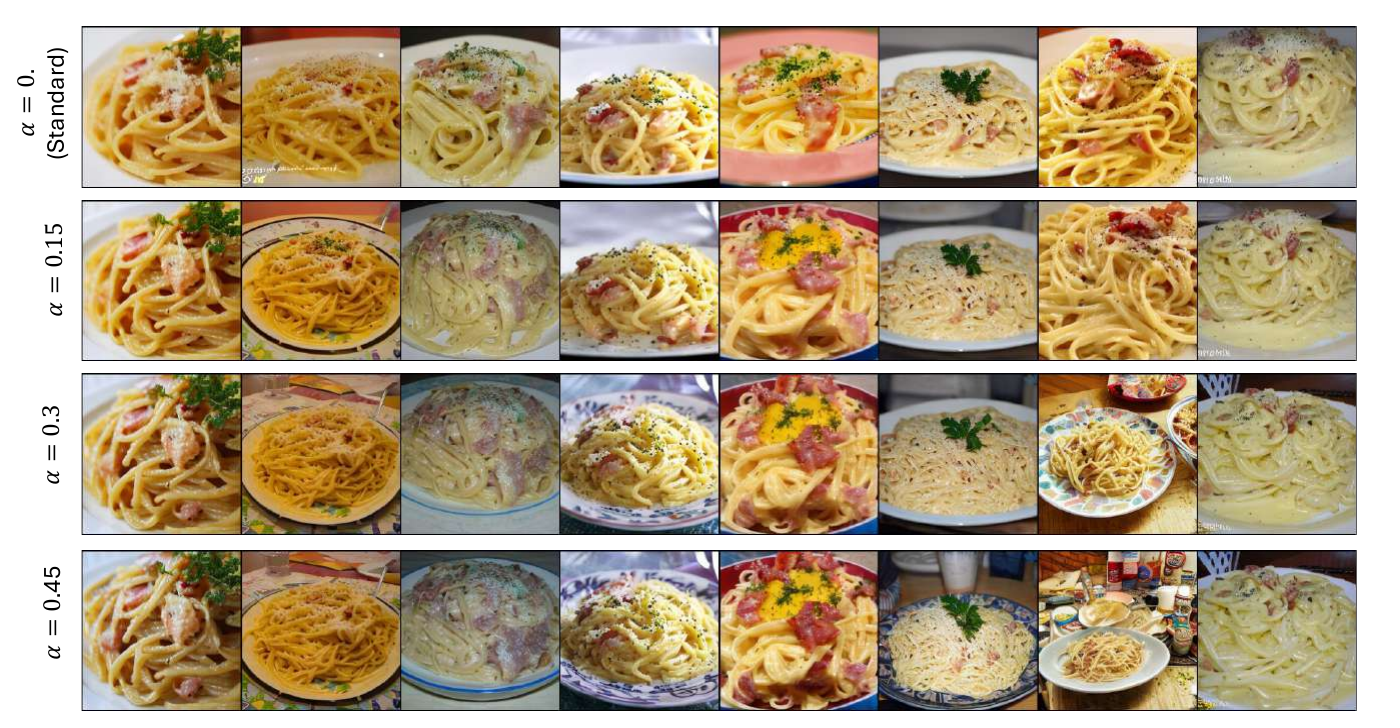}
            \caption{Class \textit{959: carbonara}}
        \end{subfigure}
        \vspace{-.35cm}
        \caption{CC generated images for different classes with $\omega=4.$ and varying value of $\alpha$.}
    \end{figure}
\else
    \begin{figure}[h]
        \centering
        \subfigure[Prompt \textit{bear in its natural habitat}]{
            \includegraphics[width=0.4\textwidth]{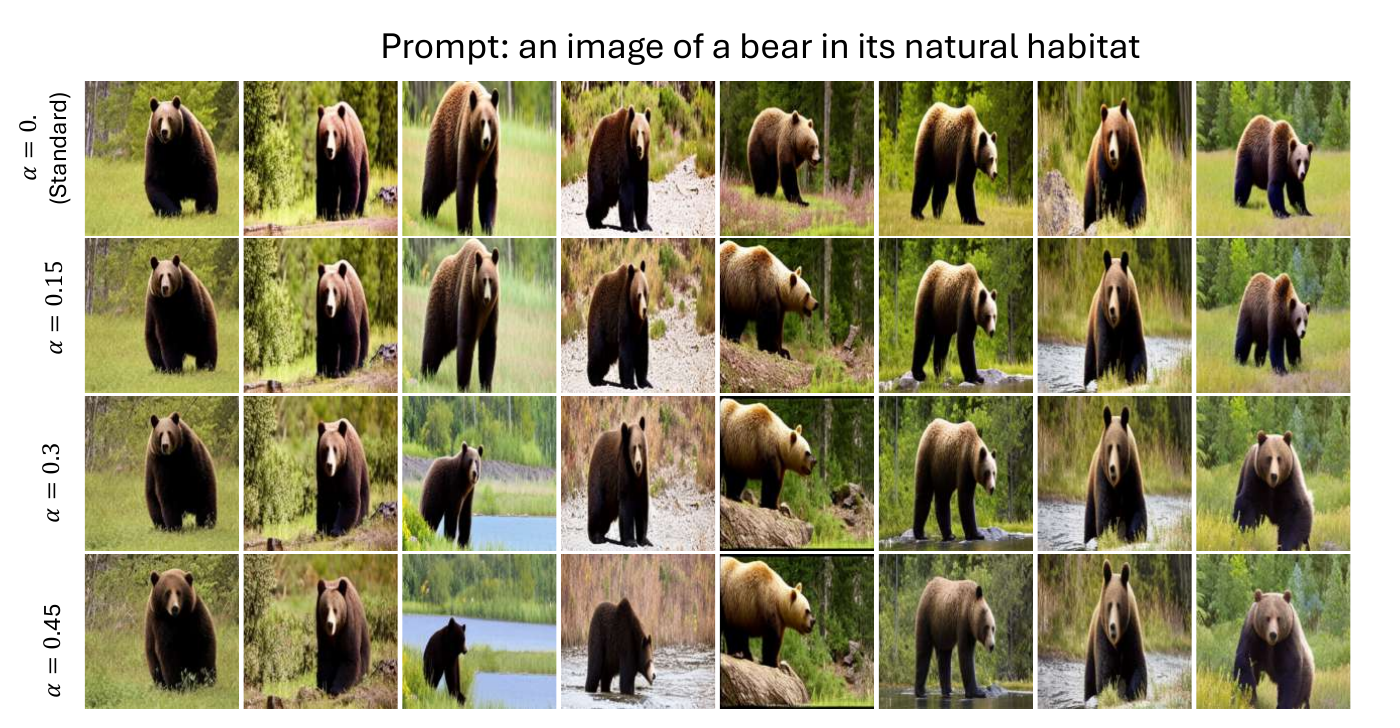}
        }%
        \hfill
        \subfigure[Prompt \textit{a bouquet of roses}]{
            \includegraphics[width=0.4\textwidth]{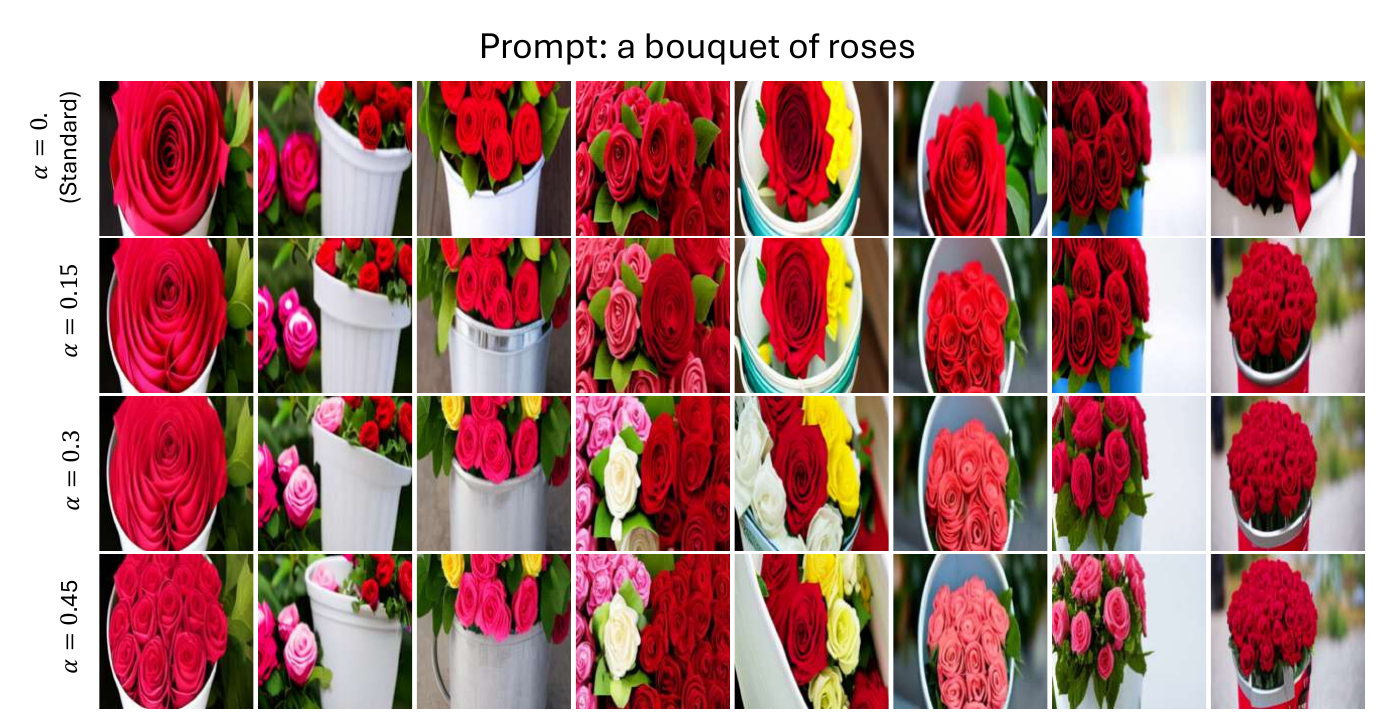}
        }
        \vspace{0.3cm} % Reduce vertical space
        \subfigure[Prompt \textit{a bowl of citrus fruit}]{
            \includegraphics[width=0.4\textwidth]{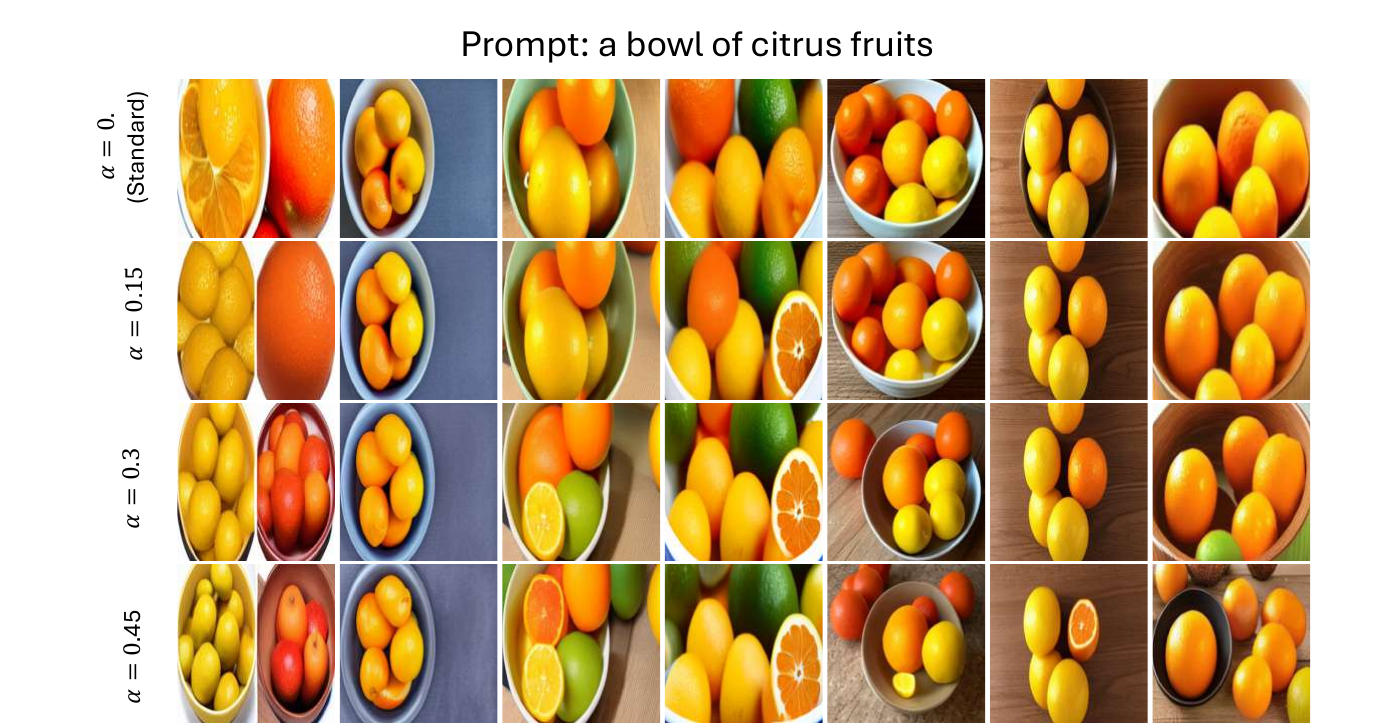}
        }%
        \hfill
        \subfigure[Prompt \textit{a simple drawing of a smiling sun}]{
            \includegraphics[width=0.4\textwidth]{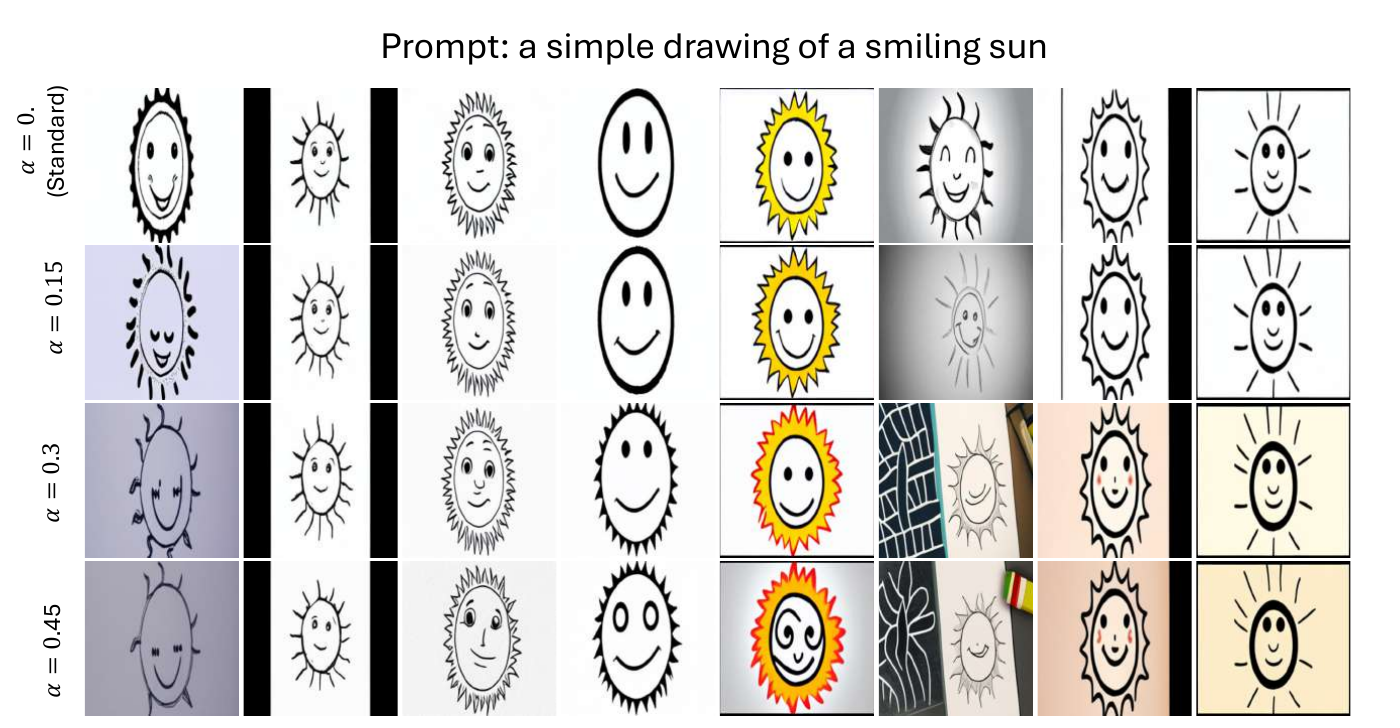}
        }
        \vspace{-0.35cm} % Adjust vertical space
        \caption{T2IM generated images for different prompts with $\omega=4.$ and varying value of $\alpha$.}
        \vspace{-0.4cm} % Adjust vertical space
    \end{figure}
    \begin{figure}[h]
        \centering
        \subfigure[Class \textit{107: jellyfish}]{
            \includegraphics[width=0.4\textwidth]{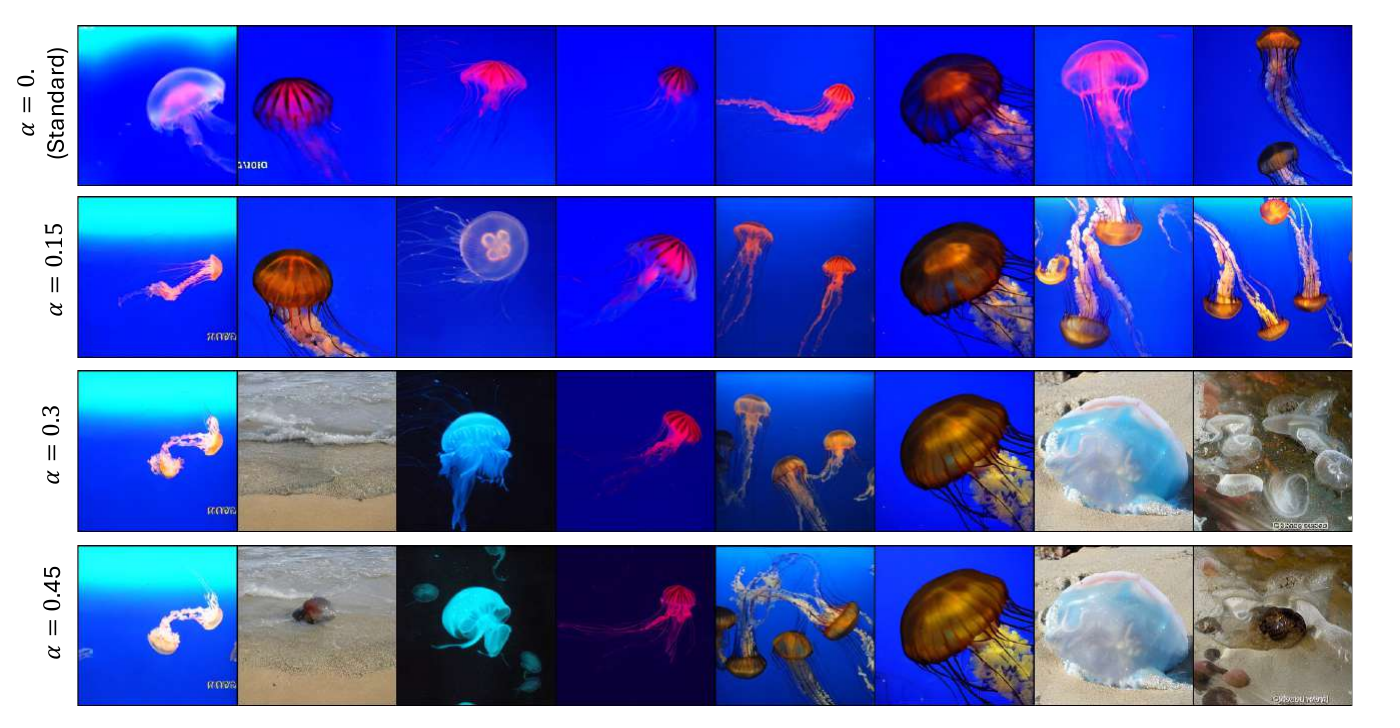}
        }%
        \hfill
        \subfigure[Class \textit{305: dung beetle}]{
            \includegraphics[width=0.4\textwidth]{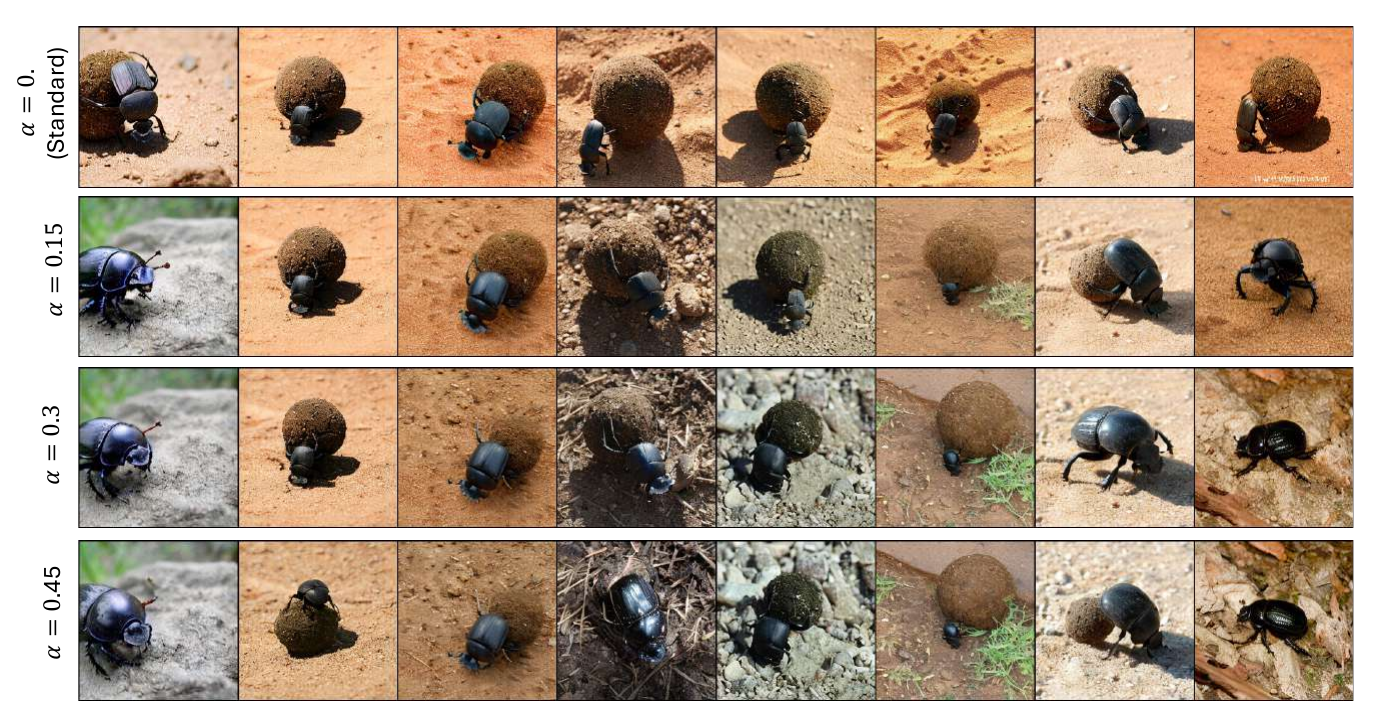}
        }
        \vspace{0.3cm} % Reduce vertical space
        \subfigure[Class \textit{309: bee}]{
            \includegraphics[width=0.4\textwidth]{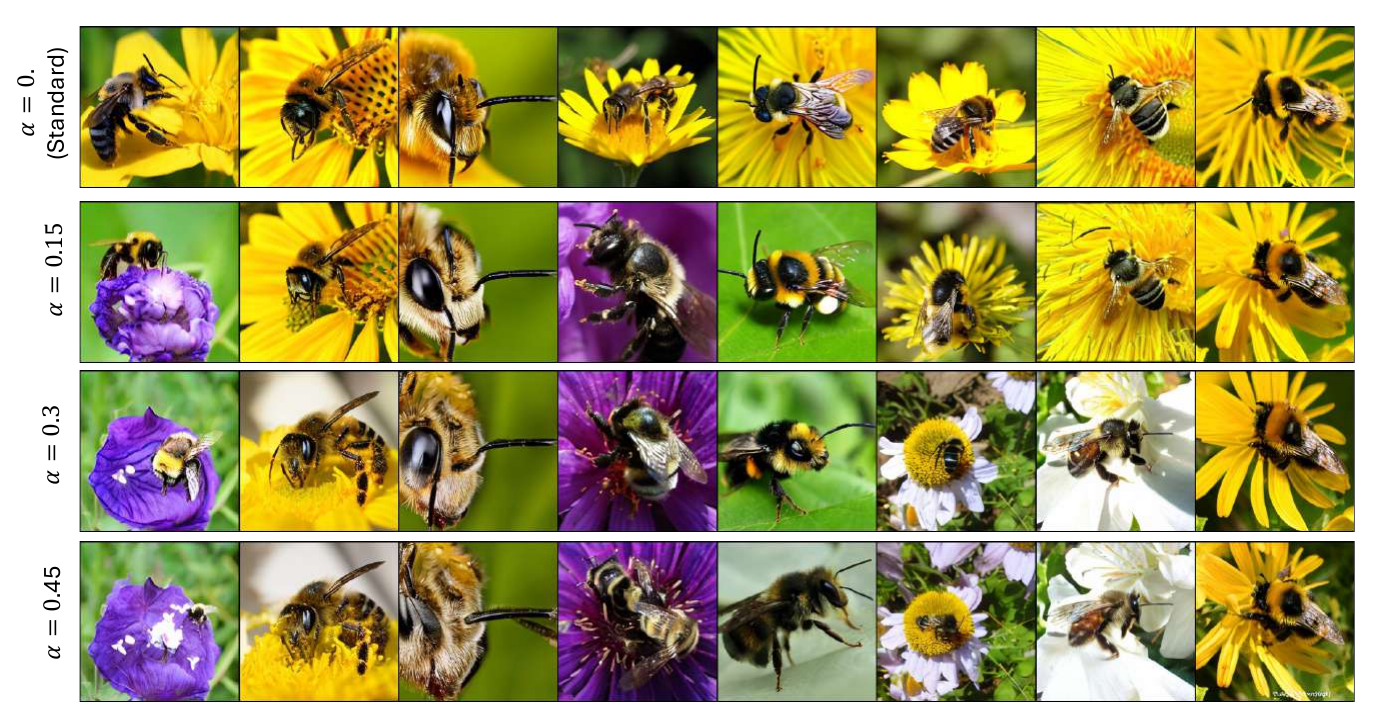}
        }%
        \hfill
        \subfigure[Class \textit{959: carbonara}]{
            \includegraphics[width=0.4\textwidth]{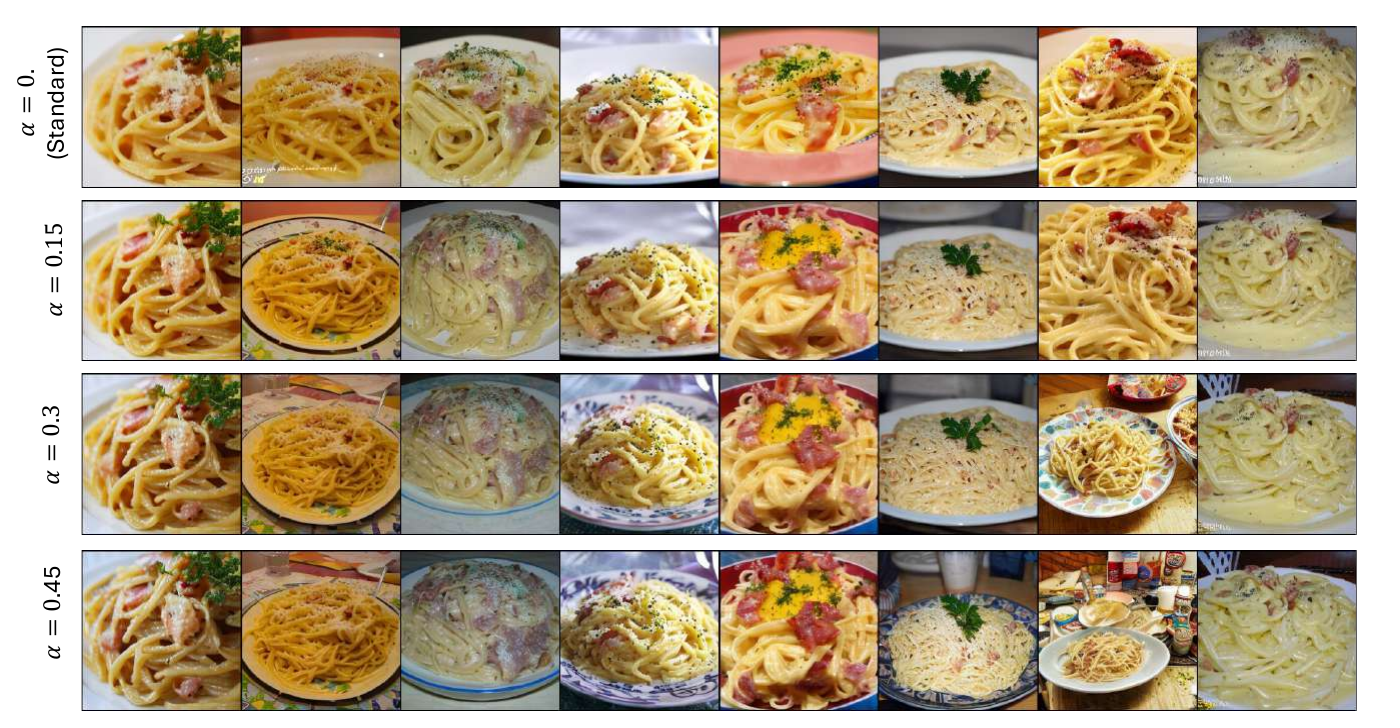}
        }
        \vspace{-0.35cm} % Adjust vertical space
        \caption{CC generated images for different classes with $\omega=4.$ and varying value of $\alpha$.}
    \end{figure}
\fi
\clearpage

\subsubsection{Generated Images by DiT/XL-2 (256x256)}
\label{appx:ex_by_dit}
\begin{figure}[ht]
    \centering
    \includegraphics[width=0.8\linewidth]{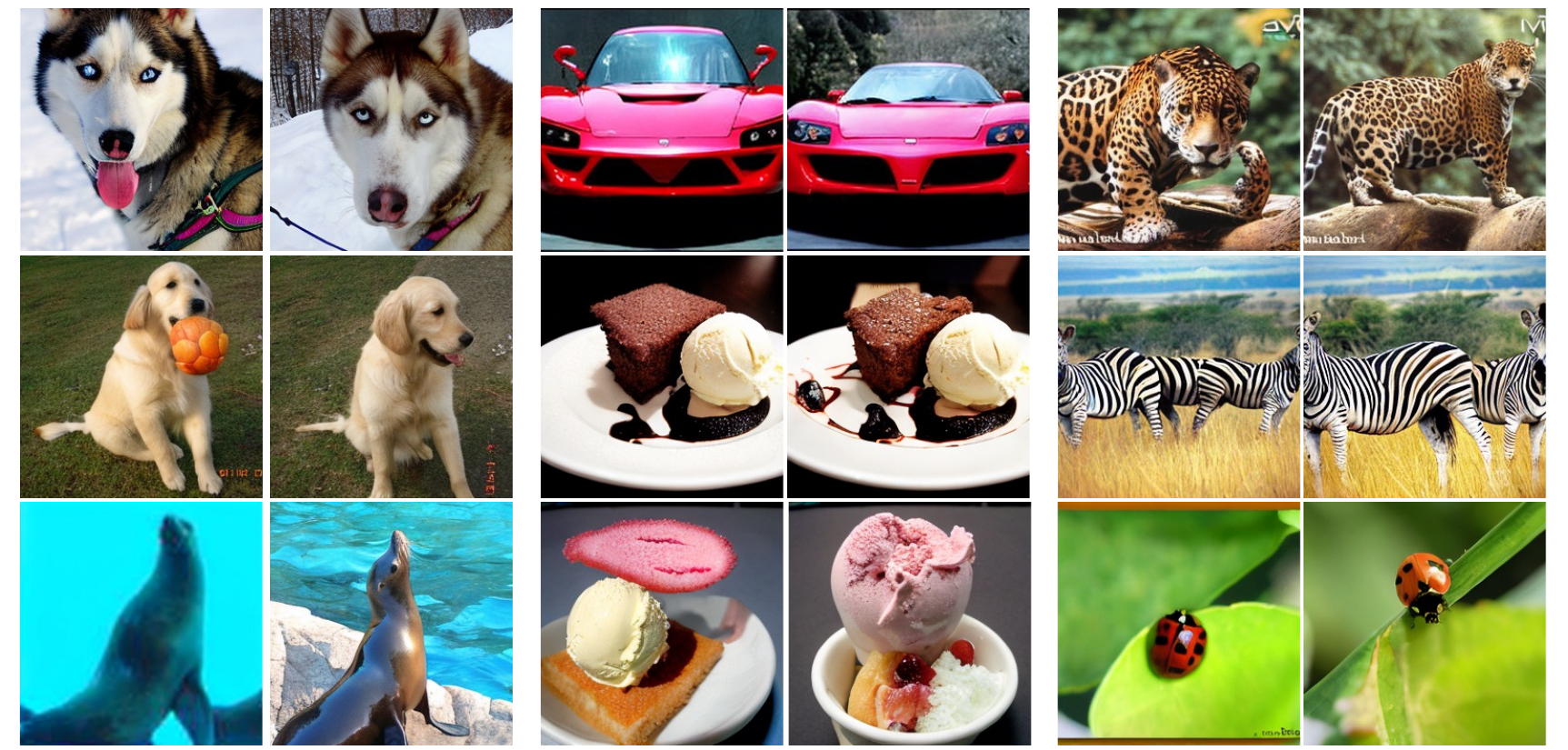}
    \caption{Additional examples generated by DiT/XL-2 using Standard CFG $(\omega=4)$ and Power-Law CFG $(\omega=8., \alpha=0.7)$. Image pairs start from the same noise (same seed). The resulting pairs represent Standard CFG on the left and Power-Law CFG on the right.}
    \label{fig:enter-label}
\end{figure}
\begin{figure}[ht]
    \centering
    \begin{minipage}{0.825\linewidth}
        \centering
        \includegraphics[width=.825\linewidth]{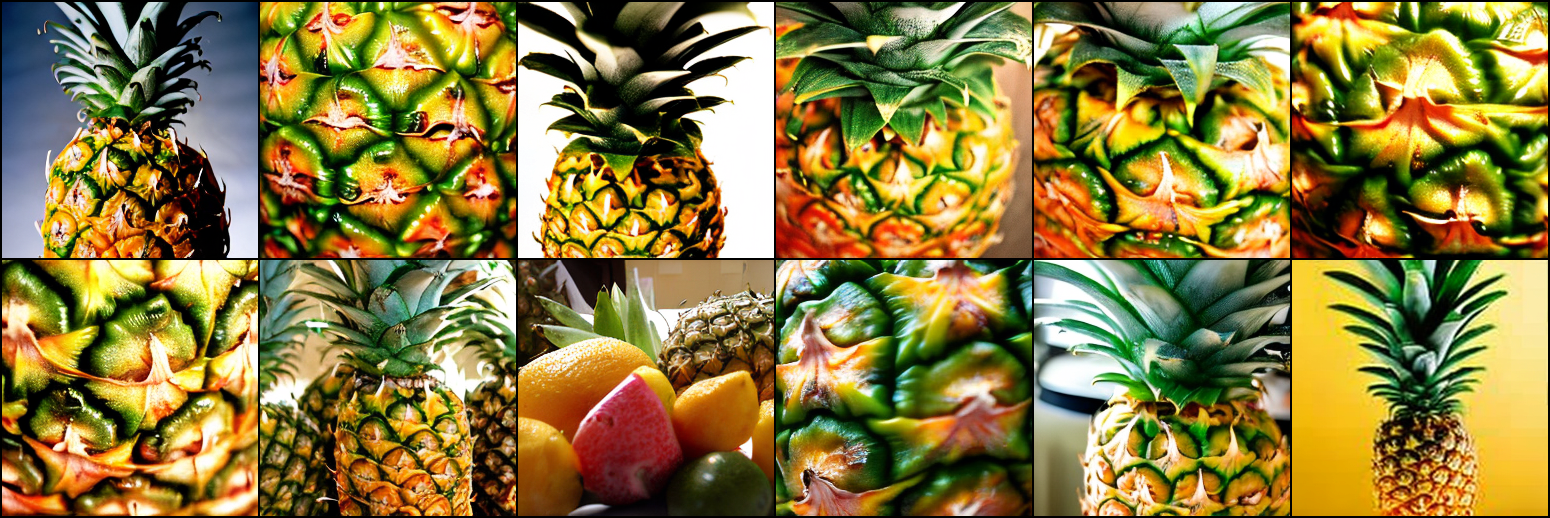}
        \caption{Gen. images conditioned on the class \textit{pineapple} with Standard CFG ($\omega=4$).}
        
        \vspace{1em} % add some space between images
        
        \includegraphics[width=.825\linewidth]{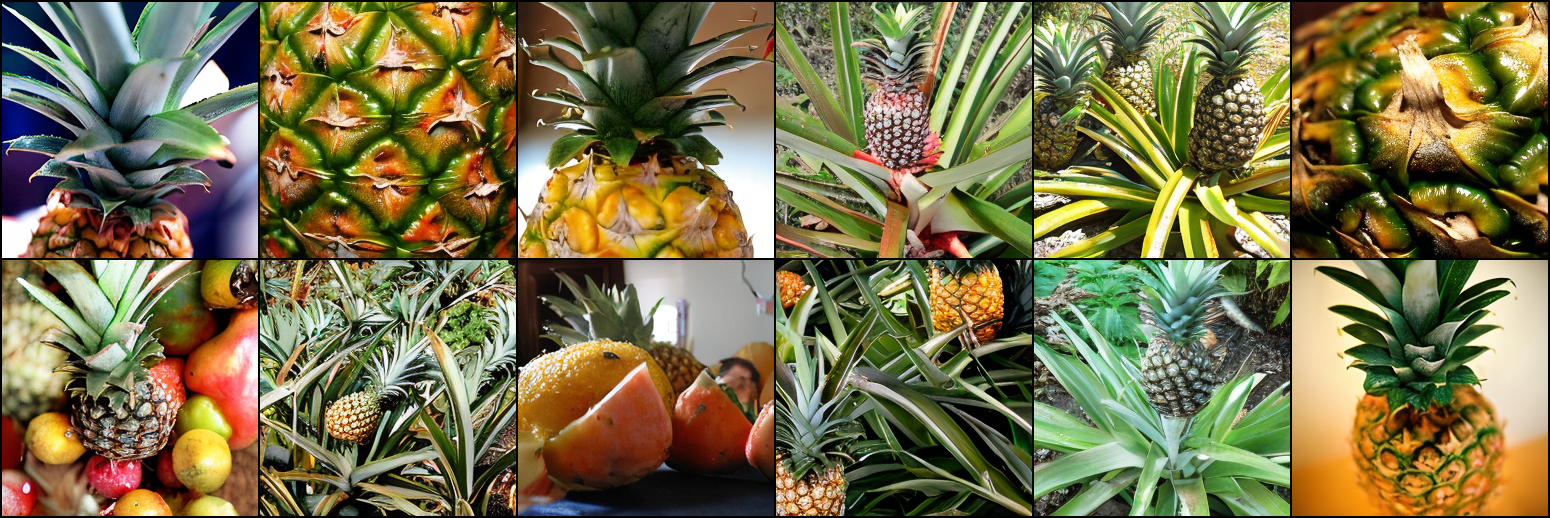} 
        \caption{Gen. images conditioned on class \textit{pineapple} with Power-Law CFG ($\omega=8, \alpha=0.7$).}
    \end{minipage}
\end{figure}
\begin{figure}[ht]
    \centering
    \begin{minipage}{0.95\linewidth}
        \centering
        \includegraphics[width=0.85\linewidth]{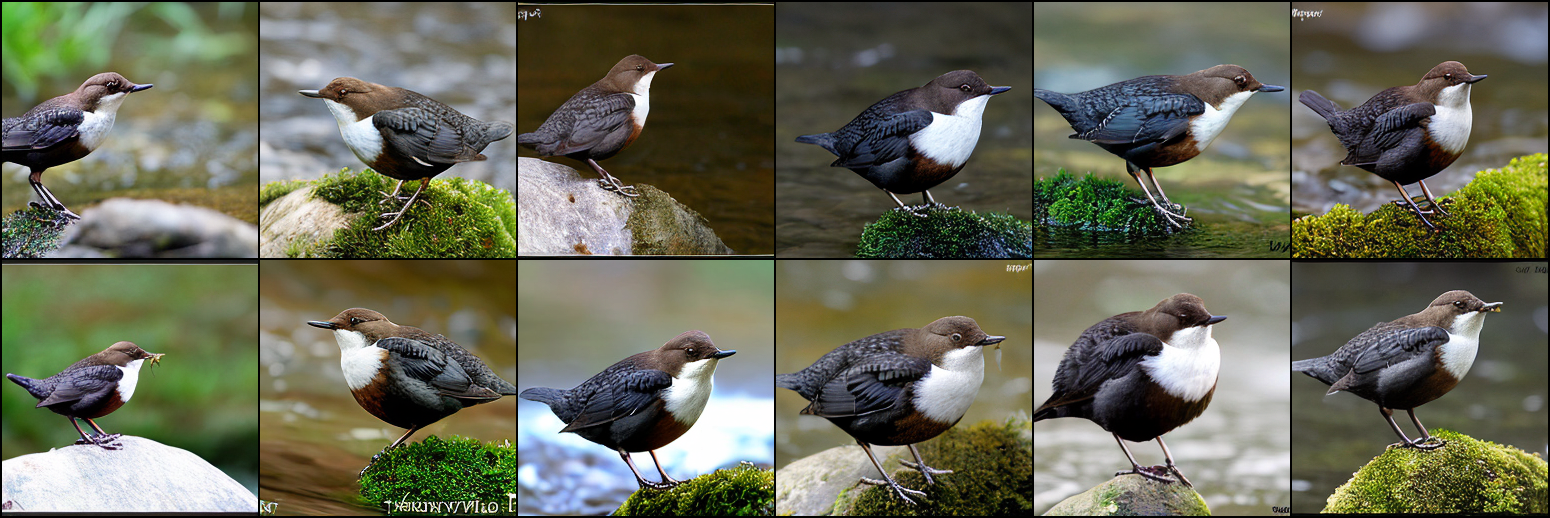}
        \vspace{-.25cm}
        \caption{Gen. images conditioned on the class \textit{water ouzel, dipper} using Standard CFG with $\omega=4$.}
        \vspace{.5cm}
        \includegraphics[width=0.85\linewidth]{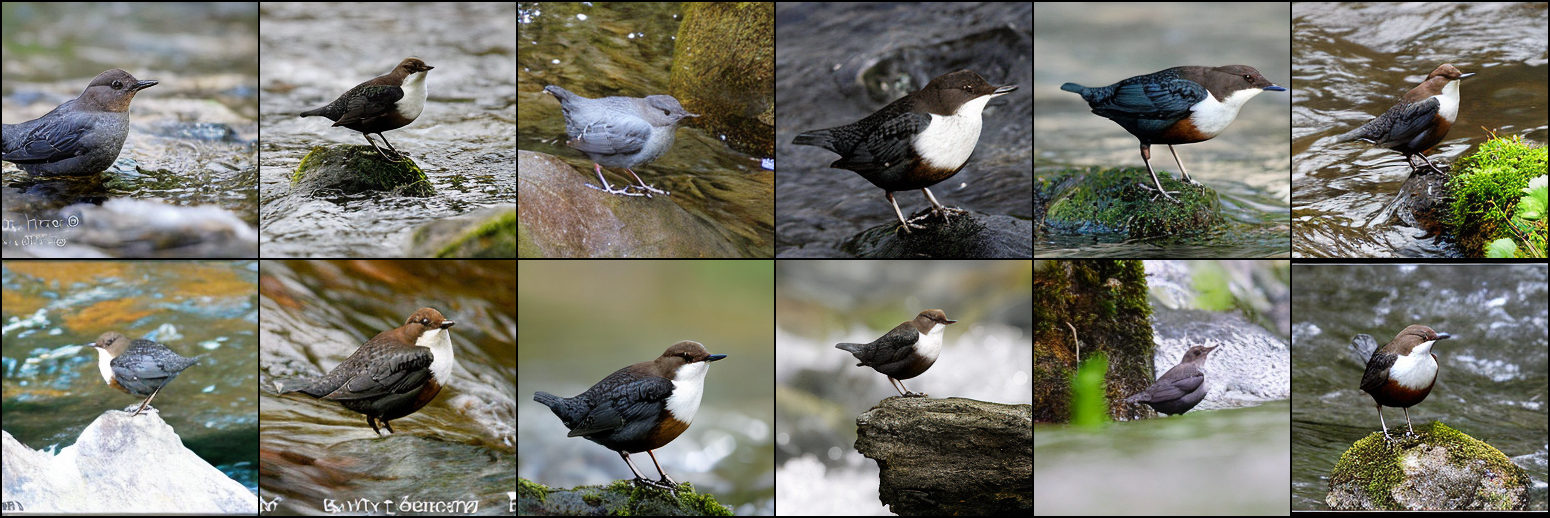} 
        \vspace{-.25cm}
        \caption{Gen. images conditioned on the class \textit{water ouzel, dipper} using Power-Law CFG with $\omega=8., \alpha=0.7$.}
        \vspace{.5cm}
        % \caption{}
        % \label{}
    \end{minipage}

    \begin{minipage}{0.95\linewidth}
        \centering
        \includegraphics[width=0.85\linewidth]{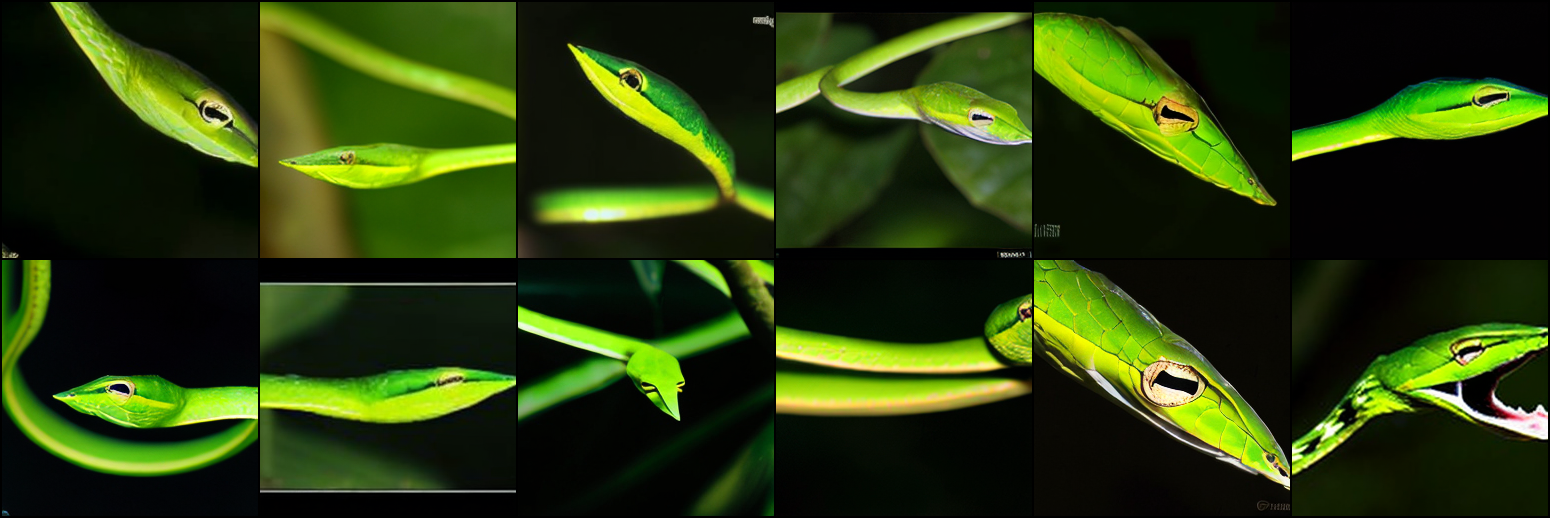}
        \vspace{-.25cm}
        \caption{Gen. images conditioned on the class \textit{vine snake} using Standard CFG with $\omega=4.$.}
        \vspace{.5cm}
        \includegraphics[width=0.85\linewidth]{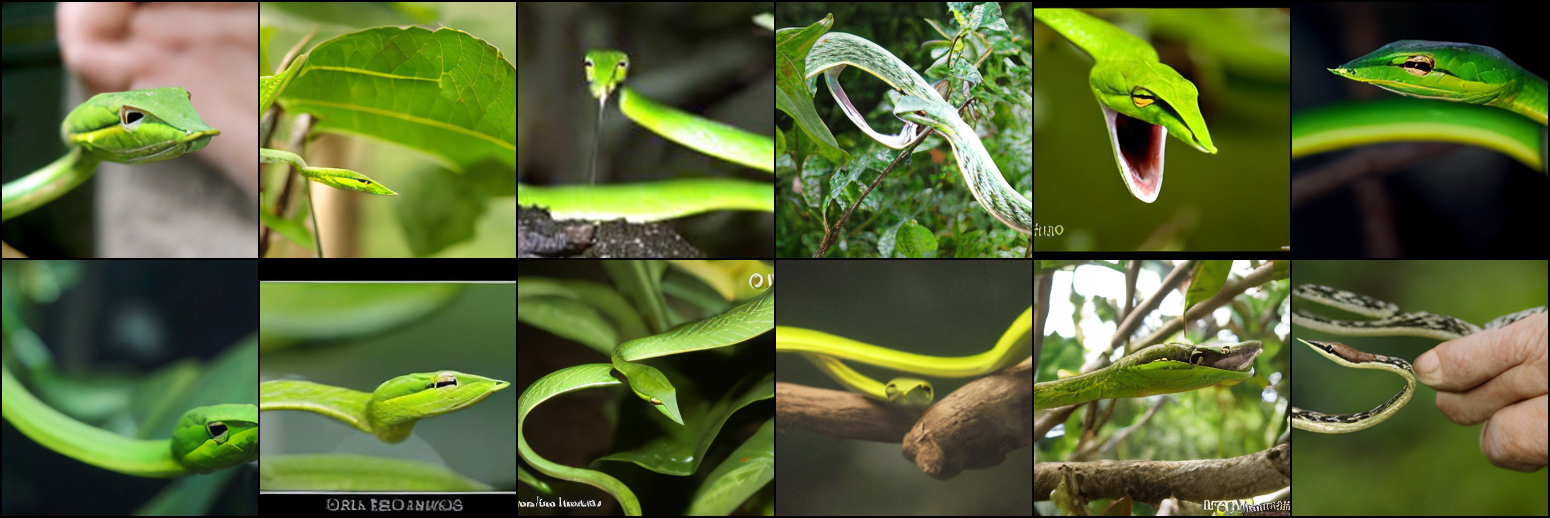} 
        \vspace{-.25cm}
        \caption{Gen. images conditioned on the class \textit{vine snake} using Power-Law CFG with $\omega=8., \alpha=0.7$.}
        % \label{}
    \end{minipage}
    % \label{}
\end{figure}

\clearpage

\subsubsection{Generated Images by MMDiT model (diffusion objective, resolution 512x512)}
\label{appx:ex_by_tti}

\ifarxiv
    \begin{figure}[ht]
        \centering
        \begin{minipage}{0.8\linewidth}
            \centering
            \includegraphics[width=.7\linewidth]{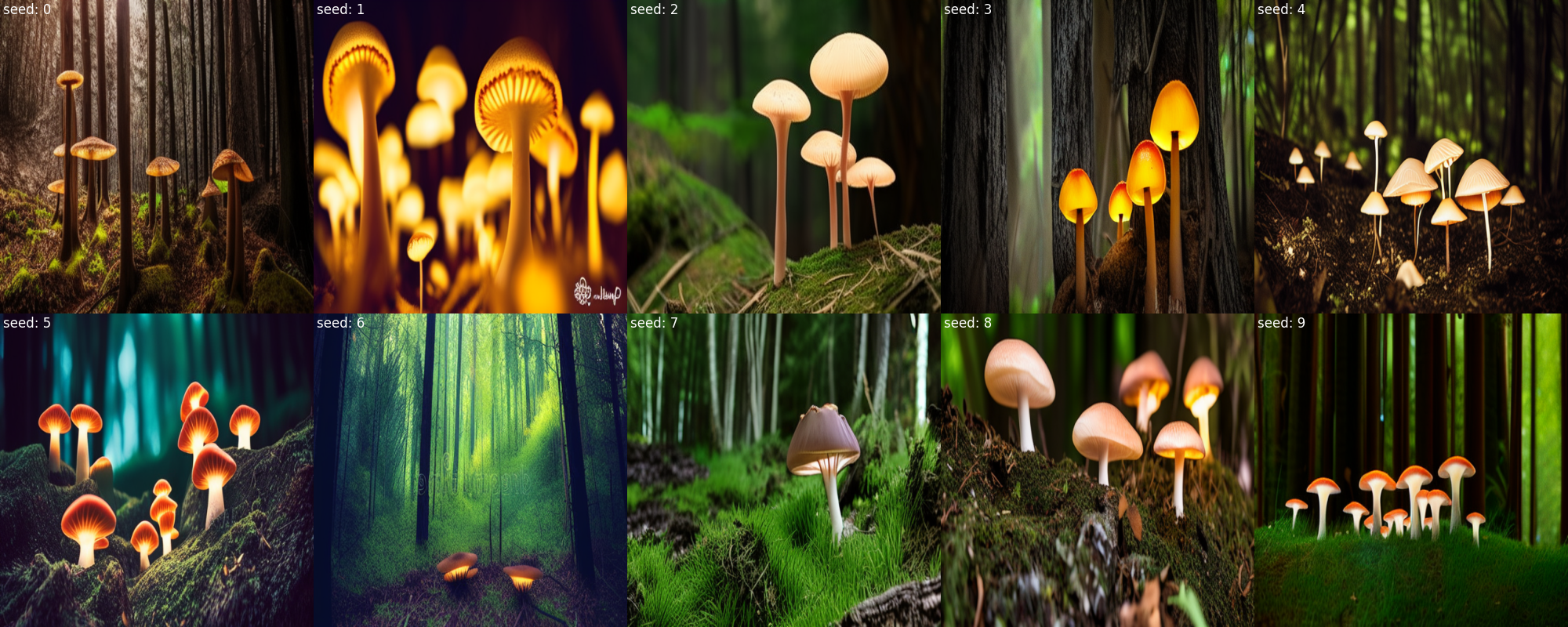}
            \includegraphics[width=.7\linewidth]{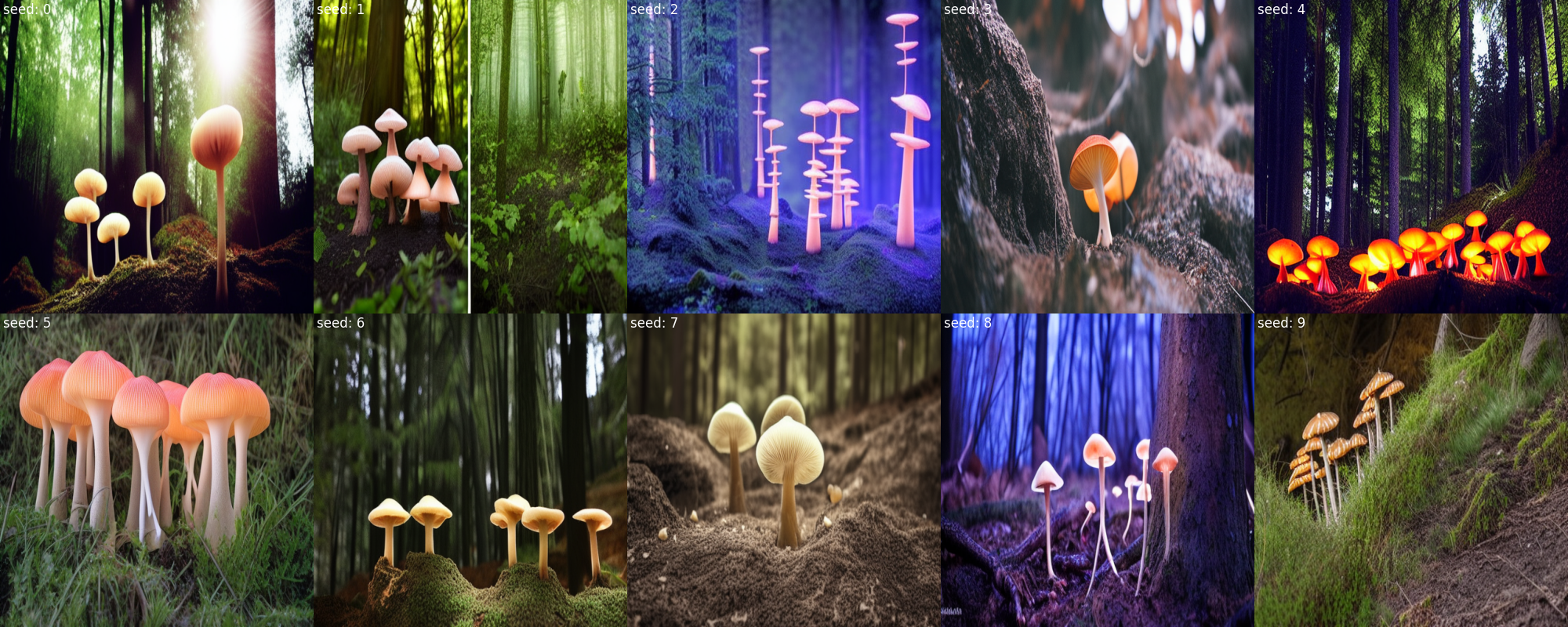} 
        \vspace{-.25cm}
            \caption{Images generated conditioned on the textual prompt \textit{Glowing mushrooms in a dark forest.} using Standard CFG with $\omega=3$ (top two rows) and Power-Law CFG with $\omega=10, \alpha=0.8$ (bottom two rows).}
            \vspace{0.5cm}
        \end{minipage}
    
        \begin{minipage}{.8\linewidth}
            \centering
            \includegraphics[width=.7\linewidth]{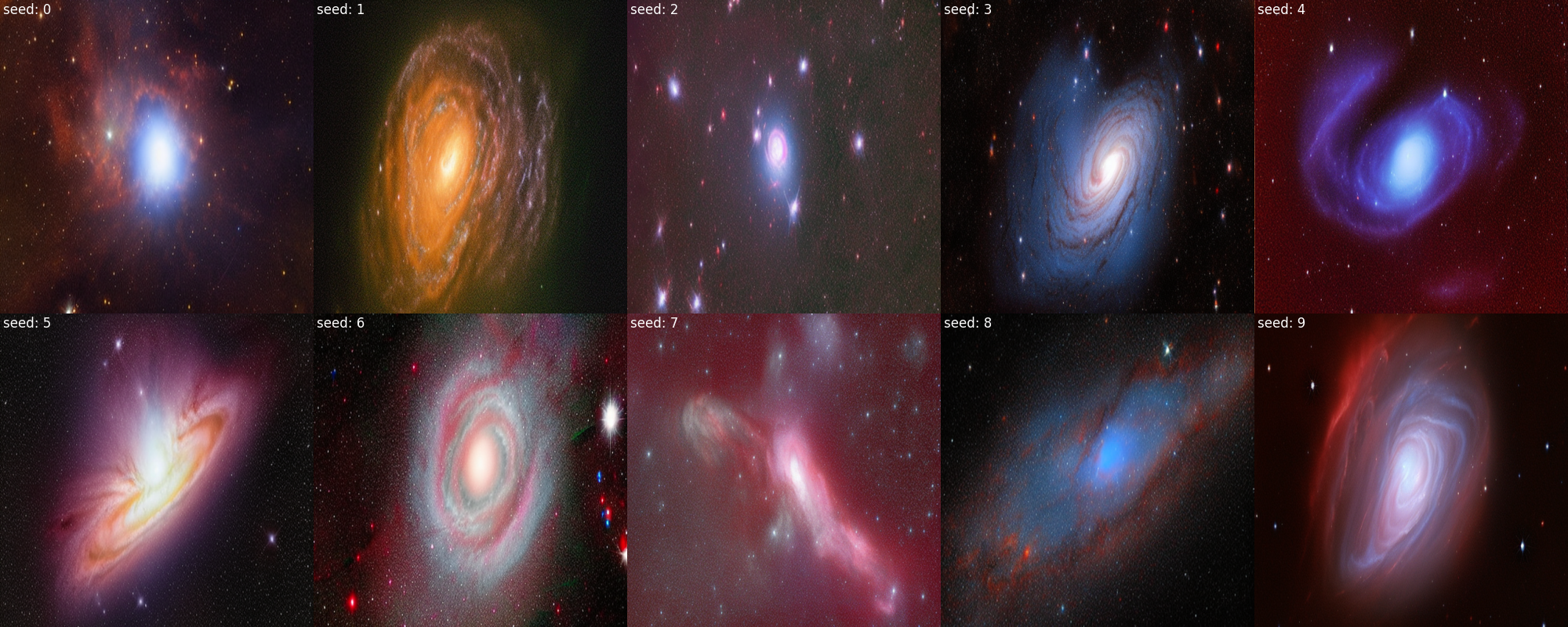}
            \includegraphics[width=.7\linewidth]{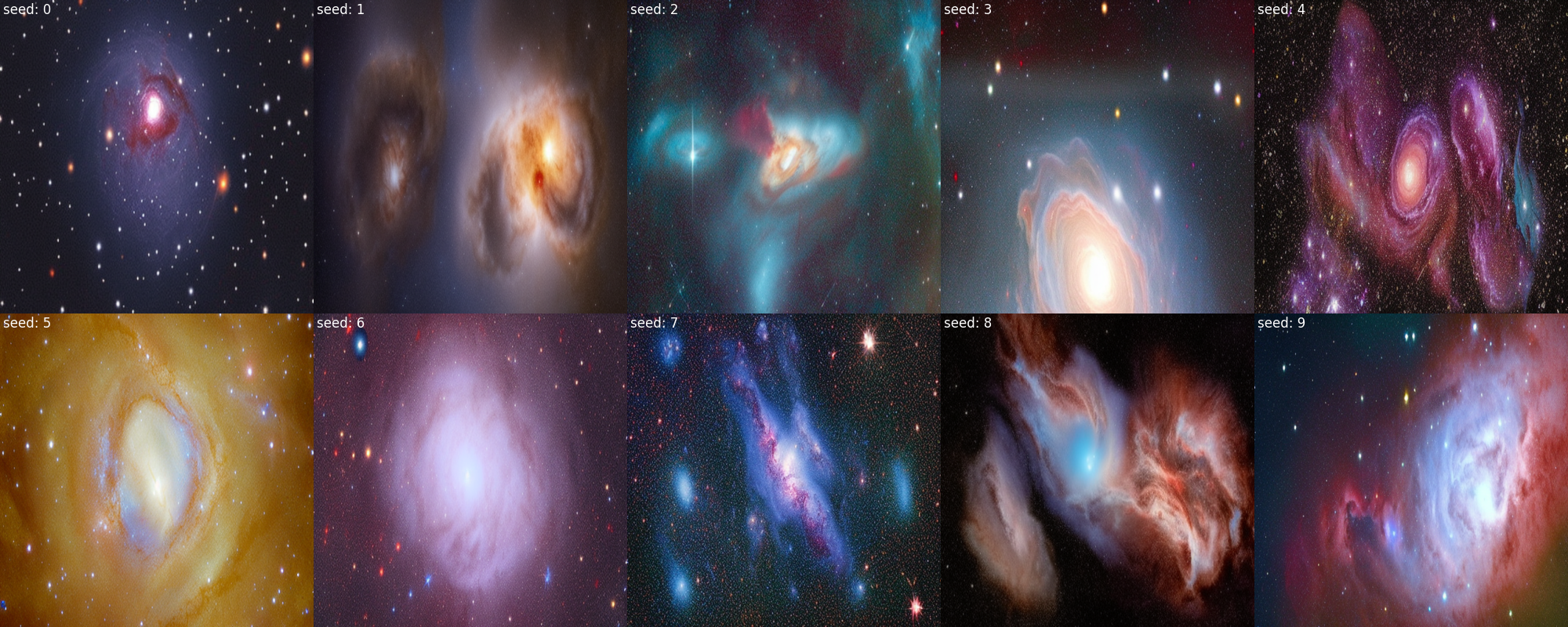} 
            \vspace{-.25cm}
            \caption{Images generated conditioned on the textual prompt \textit{Stunning, breathtaking view of a galaxy or nebula} using Standard CFG ($\omega=3$, top two rows) and Power-Law CFG ($\omega=10, \alpha=0.8$, bottom two rows).}
        \end{minipage}
        
    \end{figure}
\else
    \begin{figure}[ht]
        \centering
        \begin{minipage}{0.9\linewidth}
            \centering
            \includegraphics[width=.75\linewidth]{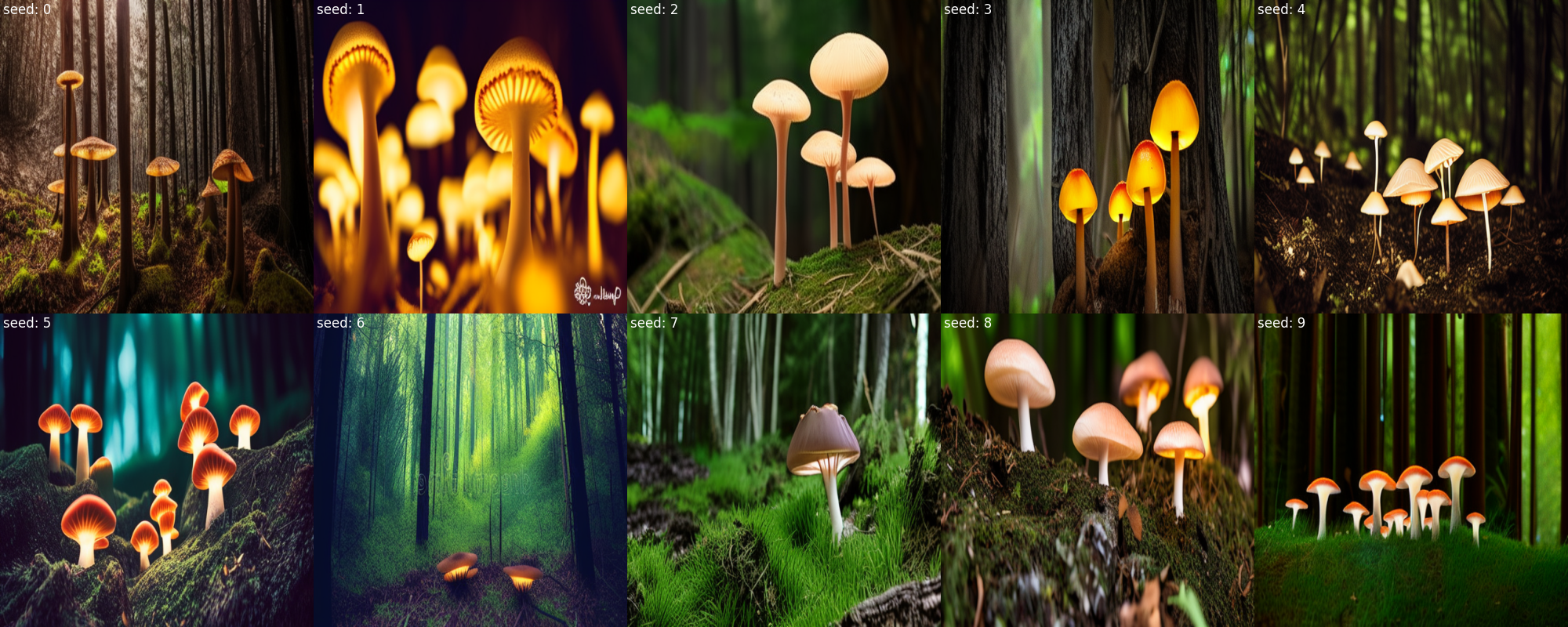}
            \includegraphics[width=.75\linewidth]{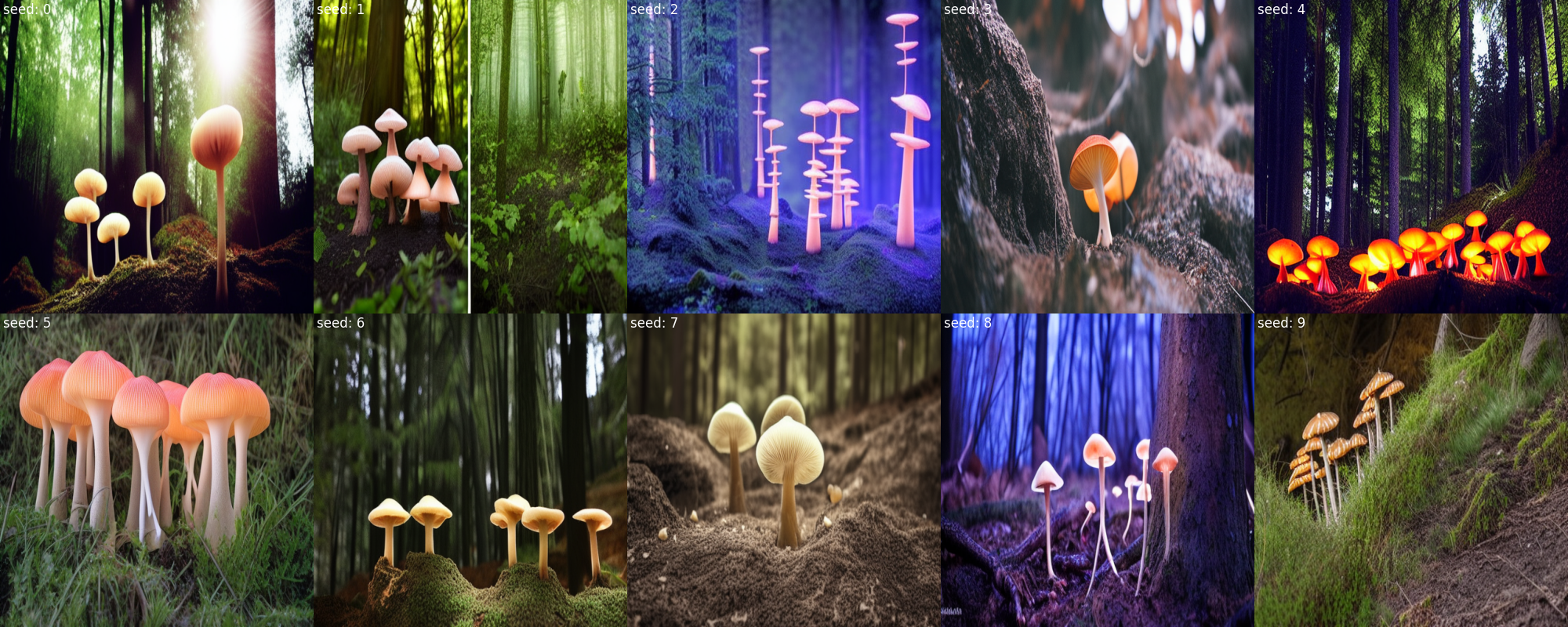} 
        \vspace{-.25cm}
            \caption{Images generated conditioned on the textual prompt \textit{Glowing mushrooms in a dark forest.} using Standard CFG with $\omega=3$ (top two rows) and Power-Law CFG with $\omega=10, \alpha=0.8$ (bottom two rows).}
            \vspace{0.5cm}
        \end{minipage}
    
        \begin{minipage}{\linewidth}
            \centering
            \includegraphics[width=.75\linewidth]{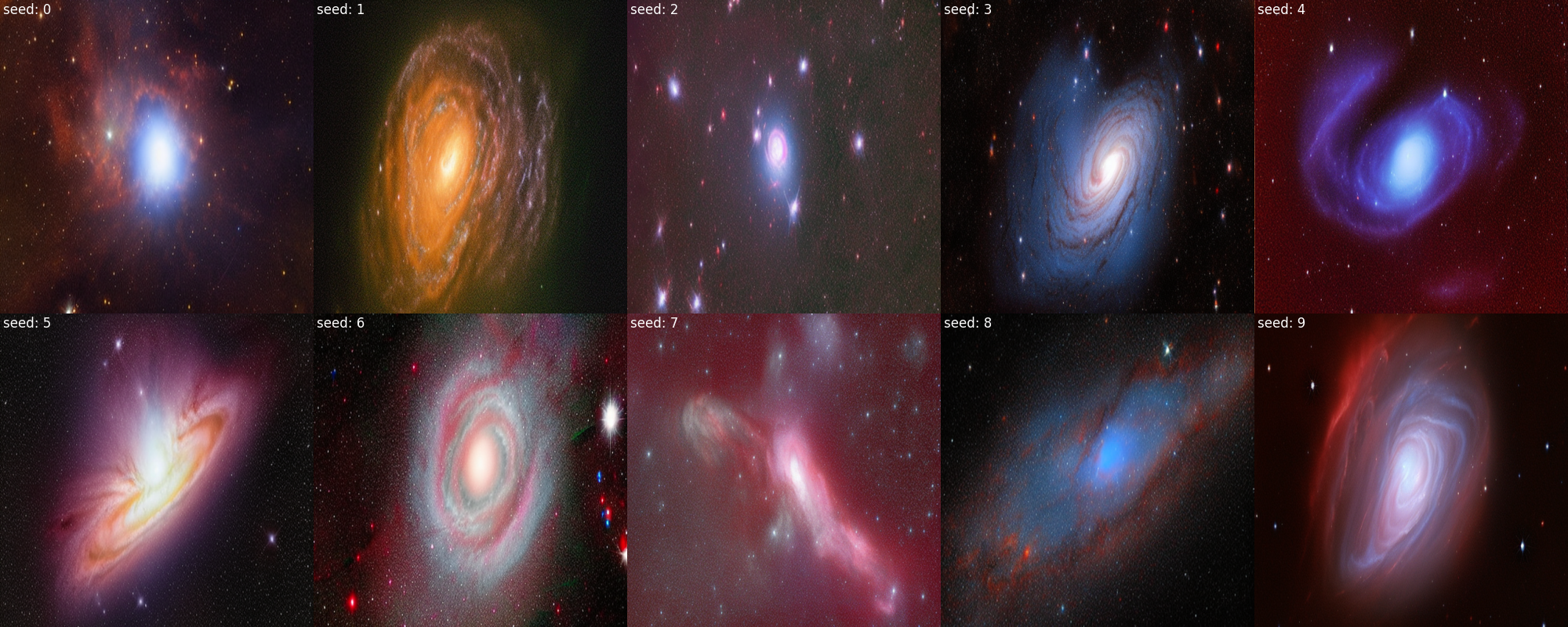}
            \includegraphics[width=.75\linewidth]{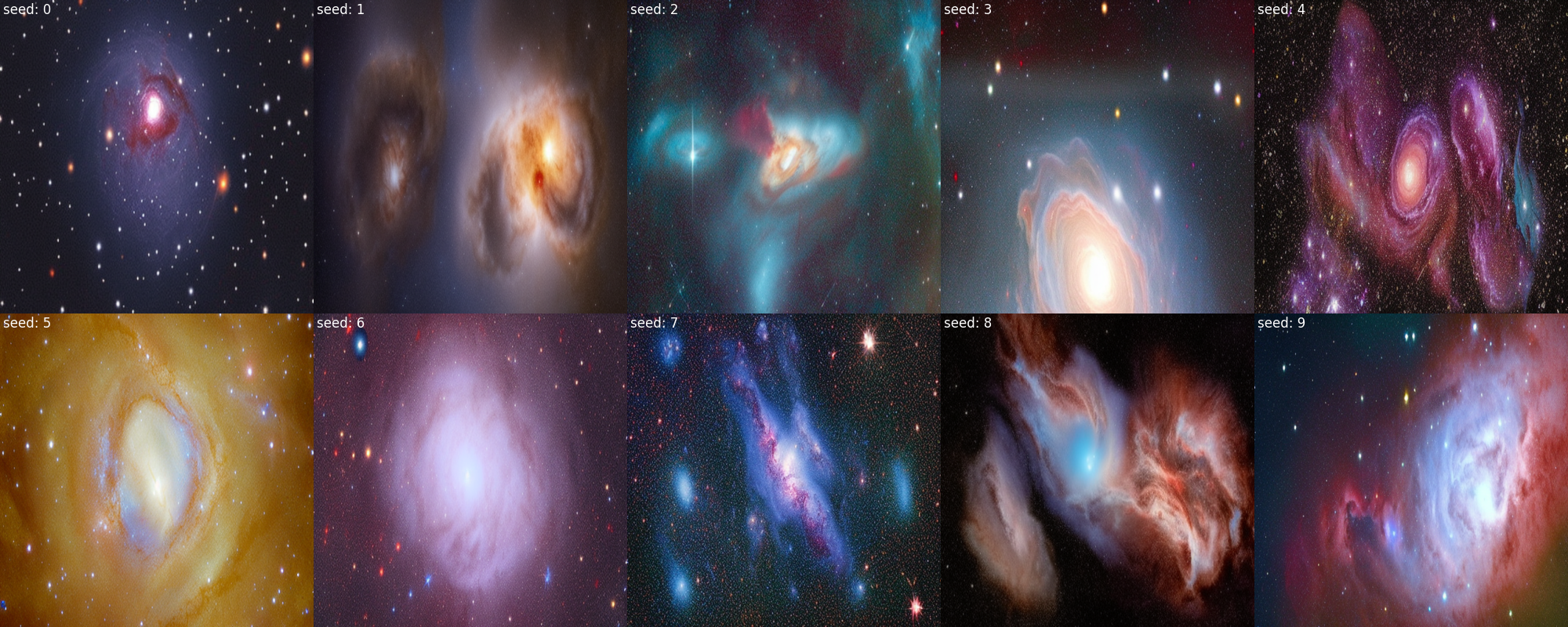} 
            \vspace{-.25cm}
            \caption{Images generated conditioned on the textual prompt \textit{Stunning, breathtaking view of a galaxy or nebula} using Standard CFG with $\omega=3$ (top two rows) and Power-Law CFG with $\omega=10, \alpha=0.8$ (bottom two rows).}
        \end{minipage}
        
    \end{figure}
\fi

\clearpage

\section{Further notes on non-linear CFG}
\label{sec:appx_G_nonlin_cfgs}

\ifarxiv
    \begin{figure}[ht]
        \centering
         \includegraphics[width=0.3\linewidth]
    {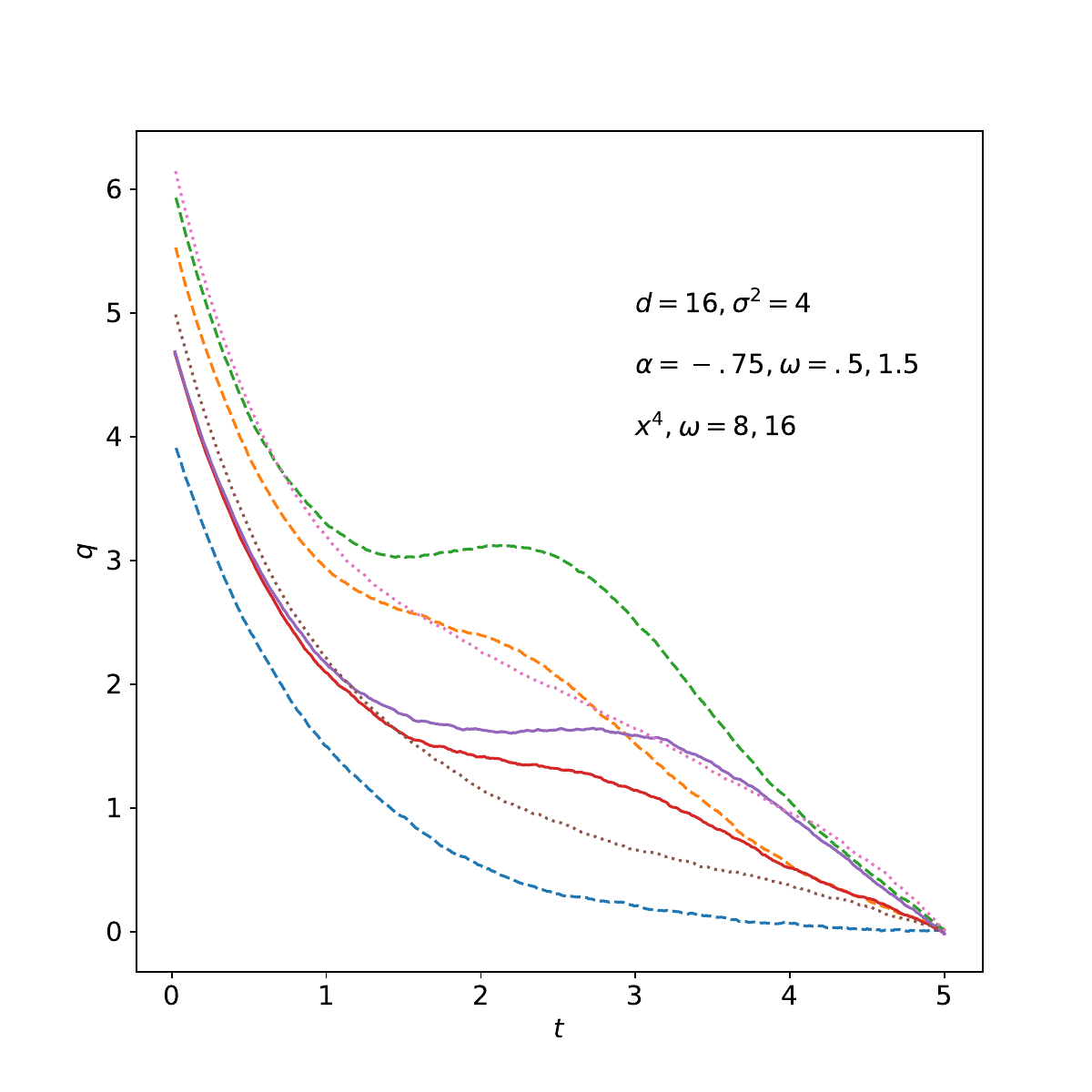}
        \caption{
        $\langle q\rangle $ versus time in Gaussian mixture with $d=16$ and $\sigma^2=4$. The dashed lines are obtained with standard CFG with $\omega=0,8,16$ from bottom to top. The dotted lines are obtained with the Power-Law scheme $f(x)=\omega x^{-.75}$ with $\omega=.5, 1.5 $ from bottom to top. The full lines are obtained with the non-linear guidance of Eq.(\ref{eqn:nonlin_cfg2}) with $\gamma=4$ and $\omega=8, 16$
        from bottom to top. The Rescaled Power-law non-linear scheme departs from $q=0$ at large time on a trajectory similar to the linear scheme and to the Power-Law non-linear scheme. But it gives a smaller bias at $t=0$.}
        \label{fig:newnonlin1}
    \end{figure}
\else
    \begin{figure}[ht]
        \centering
         \includegraphics[width=0.35\linewidth]
    {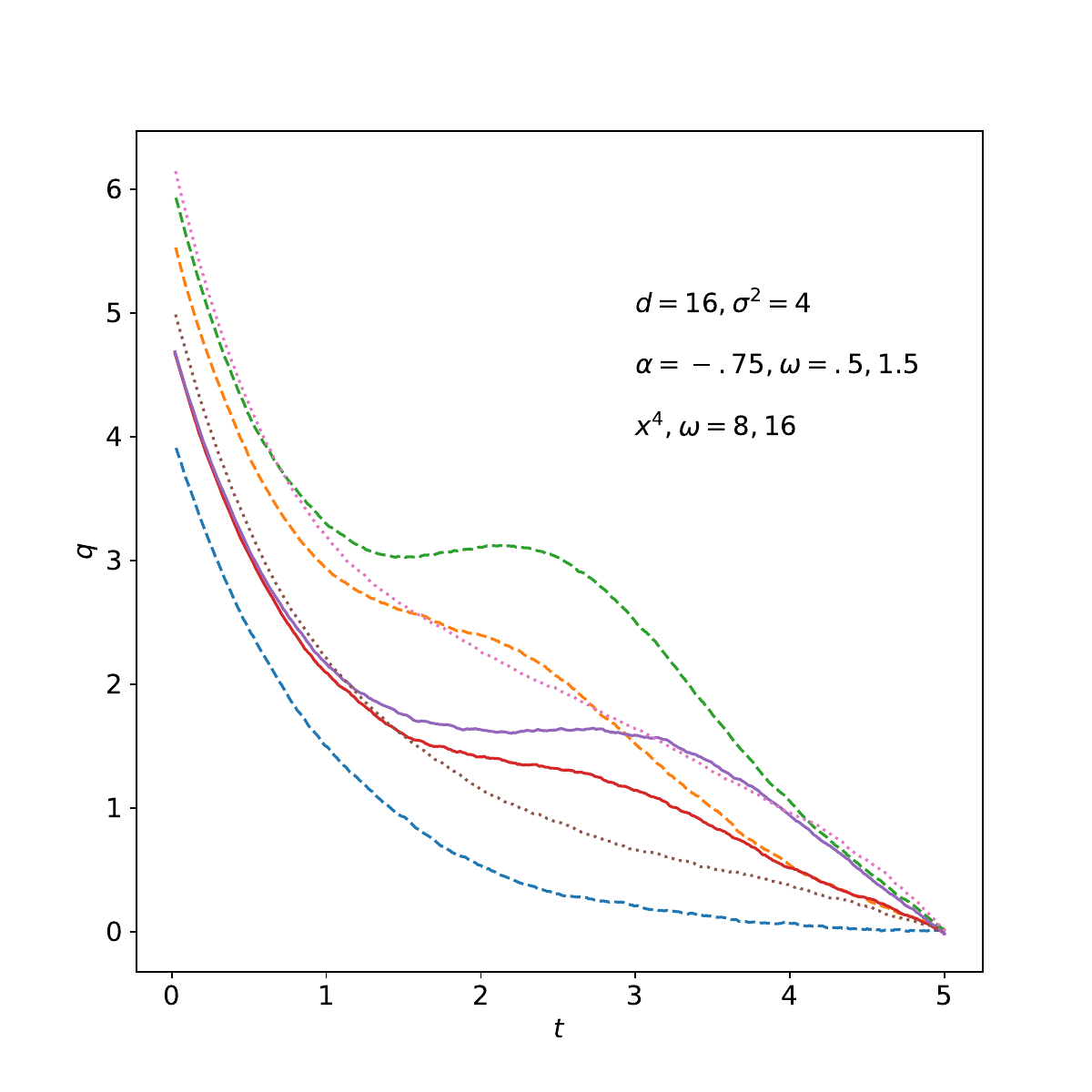}
        \caption{
        $\langle q\rangle $ versus time in the Gaussian binary mixture with $d=16$ and $\sigma^2=4$. The dashed line are obtained by the standard linear CFG with $\omega=0,8,16$ from bottom to top. The dotted line are obtained with the Power-Law non-linear scheme $f(x)=\omega x^{-.75}$ with $\omega=.5, 1.5 $ from bottom to top. The full lines are obtained with the non-linear guidance of Eq.(\ref{eqn:nonlin_cfg2}) with $\gamma=4$ and $\omega=8, 16$
        from bottom to top. The Rescaled Power-law non-linear scheme departs from $q=0$ at large time on a trajectory similar to the linear scheme and to the Power-Law non-linear scheme. But it gives a smaller bias at $t=0$.}
        \label{fig:newnonlin1}
    \end{figure}
\fi

The first non-linear CFG proposal, the power-law CFG with $\phi_t(s)=\omega s^{\alpha}$ and $\alpha>-1$ results in the following guidance scheme:
\begin{align}
    \vec S_{t}^\textrm{PL}(\vec{x},c)= S_t(\vec{x}, c)+ \omega \left[S_t(\vec{x}, c) -  S_t(\vec{x})\right] \left|\vec{S}_t(\vec{x}, c) -  \vec{S}_t(\vec{x})\right|^{\alpha}. 
\end{align}
As mentioned, the $\ell_2$ distance between scores $\delta S_t =|\vec{S}_t(\vec{x}, c) -  \vec{S}_t(\vec{x})|$ is exponentially small both at the beginning of the backward process (as both conditional and unconditional distributions are standard Gaussian clouds) and before exiting Regime \rom{1} (as shown in \Cref{sec:gauss_mixt}), after which it remains zero. Theis non-linear scheme automatically switches off in Regime \rom{2} and has the following properties: choosing $\alpha<0$ provides guidance which speeds up convergence to the target at early times, while $\alpha>0$ dampens the guidance for small $\delta S_t$ and strengthens it for large $\delta S_t$. In practice, we found positive values for $\alpha$ to perform best. In numerical experiments for finite dimension it biases the distribution obtained at $t=0$ (see Fig.\ref{fig:large}).

One would like to have different non-linearities applying to the regimes $t\gg t_s$ and $t<t_s$. One possibility is to use the following version, which extends to more general effective distributions $P_0(\vec{a}) e^{-\vec{a}^2 s(t) /\left(2 s(t)^2\sigma(t)^2\right)}$ with non-standard $s(t)$ and $\sigma(t)$.

\begin{figure}[ht]
    \centering
     \includegraphics[width=0.325\linewidth]
    {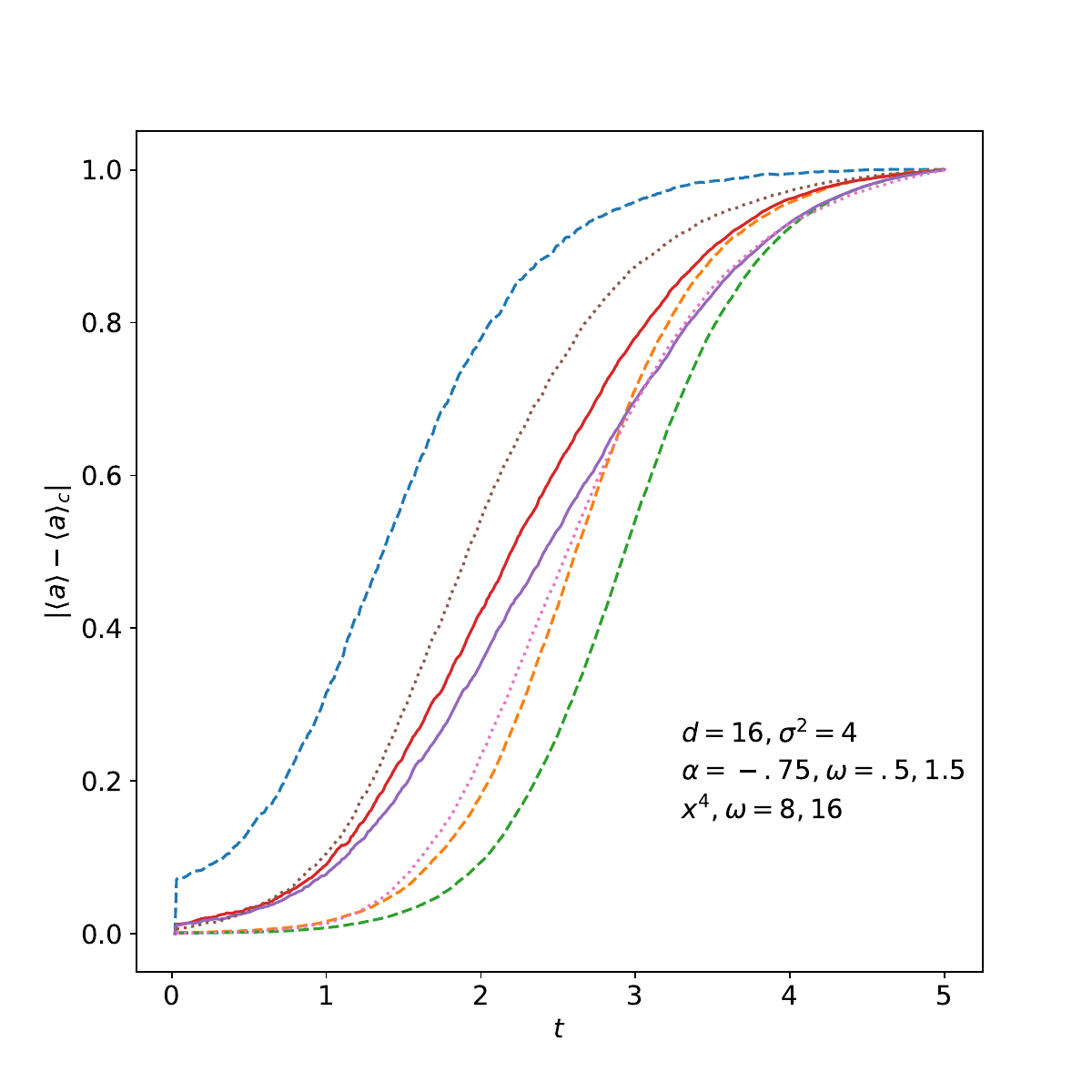}
    \includegraphics[width=0.325\linewidth]
    {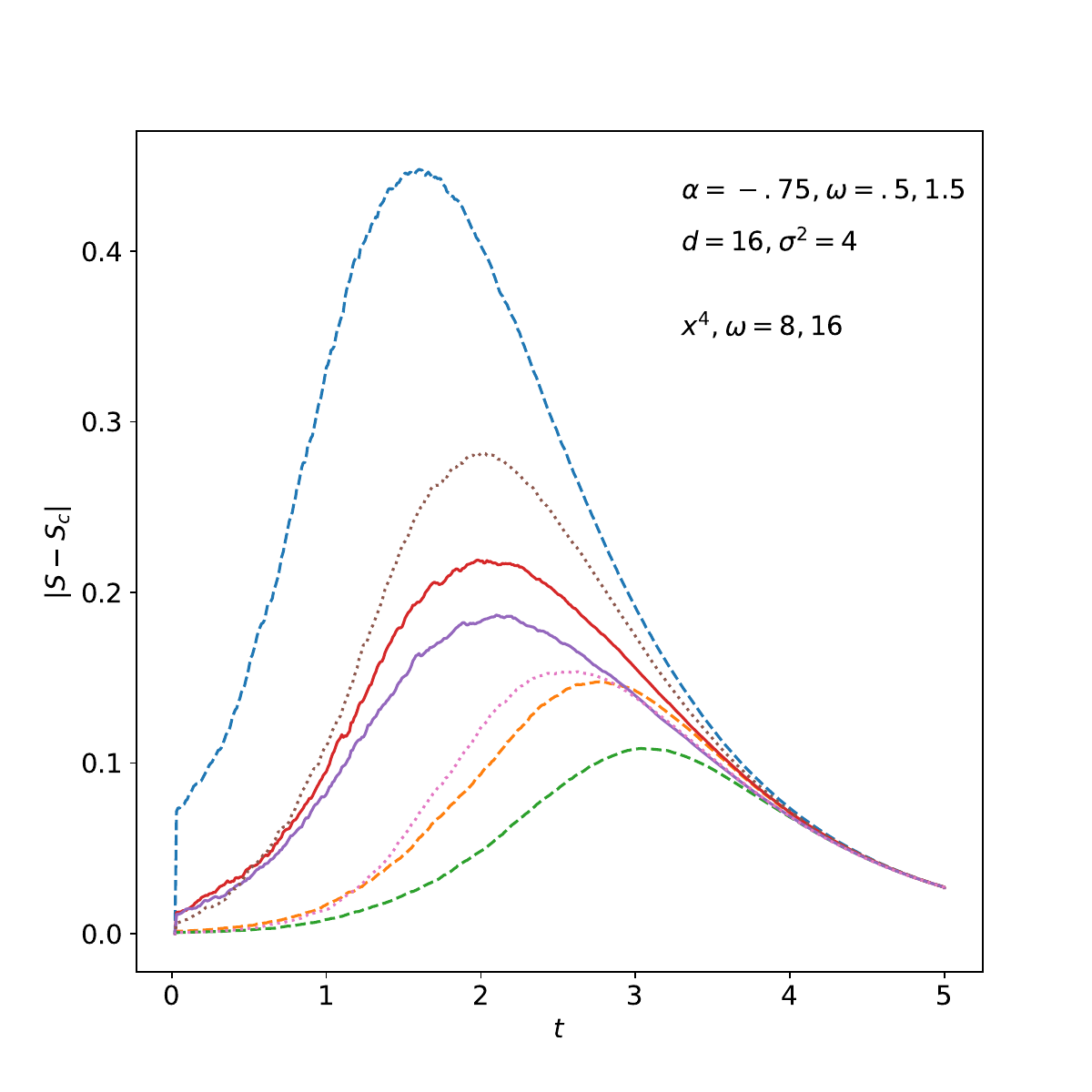}
    \caption{We perform the same experiment as in Fig. \ref{fig:newnonlin1}.
    Left: the value of $|\langle \vec a\rangle_{\vec x=\vec 0, c}-\langle \vec a\rangle_{\vec x=\vec 0} |$. Right: the value of $|S_t(\vec{x}, a) -  S_t(\vec{x})|$, with the same linestyle and color code as in Fig. \ref{fig:newnonlin1}.
    }
    \label{fig:newnonlin2}
\end{figure}

\paragraph{Rescaled Power-law CFG.}
Here, by denoting with $\langle \cdot \rangle$ the expectation \wrt the effective distribution $P_0(\vec{a}) e^{-\vec{a}^2 s(t) /\left(2 s(t)^2\sigma(t)^2\right)}$, the score difference can be expressed as $|\vec{S}_t(\vec{x}, c) -  \vec{S}_t(\vec{x})|=(1/(s(t)\sigma(t)^2)\;  |\langle \vec a\rangle_{\vec x, c}-\langle \vec a\rangle_{\vec x} |$, where $s(t)$ and $\sigma(t)$ are related to the functions $f(t)$ and $g(t)$ by 
$s(t)=\exp\int _0^t d\tau f(\tau)$ and $\sigma(t)=\int_0^t d\tau g(\tau)^2/s(\tau)^2$.
Therefore the non-linear function depends on the difference between the estimators of the initial value $\vec a$, given $\vec x(t)$, in the class and in the full distribution. This difference is typically a function that decreases with the time of the backward process. 
This suggests to use a non-linear CFG of the form 

\begin{align}
 \vec S_{t}^\textrm{RPL}(\vec{x},c)
 &= 
 \vec{S}_t(\vec{x}, c) + \omega \; \left[\vec{S}_t(\vec{x}, c) -  \vec{S}_t(\vec{x})\right] \; 
\left|\langle \vec a\rangle_{\vec x, c}-\langle \vec a\rangle_{\vec x} \right|^\gamma\nonumber\\
&= 
\vec{S}_t(\vec{x}, c) + \omega \; \left[\vec{S}_t(\vec{x}, c) -  \vec{S}_t(\vec{x})\right] \; 
\left|\vec{S}_t(\vec{x}, c) -  \vec{S}_t(\vec{x})\right|^{\gamma} s(t)^\gamma \sigma(t)^{2\gamma},
\label{eqn:nonlin_cfg2}
\end{align}
with positive $\gamma$. 
As we will show in Figures \ref{fig:newnonlin1}-\ref{fig:newnonlin2}, this non-linear guidance term has interesting performance in terms of combining a rapid drift toward the desired class $c$ at early stages of the backward process together with small bias in the finite distribution in finite dimensional problems. 

The behavior of both versions is portrayed in Figure \ref{fig:large}: both non-lin. versions yield smaller bias at $t=0$. Furthermore, Figure \ref{fig:large} also displays additional experiments highlighting the benefits of non-linear versions.

\newcommand{\labelfiglarge}{\label{fig:large}}

\ifarxiv
    \begin{figure*}[ht]
        \centering
        \begin{subfigure}[b]{\textwidth}
        \centering
        \includegraphics[width=.7\textwidth]{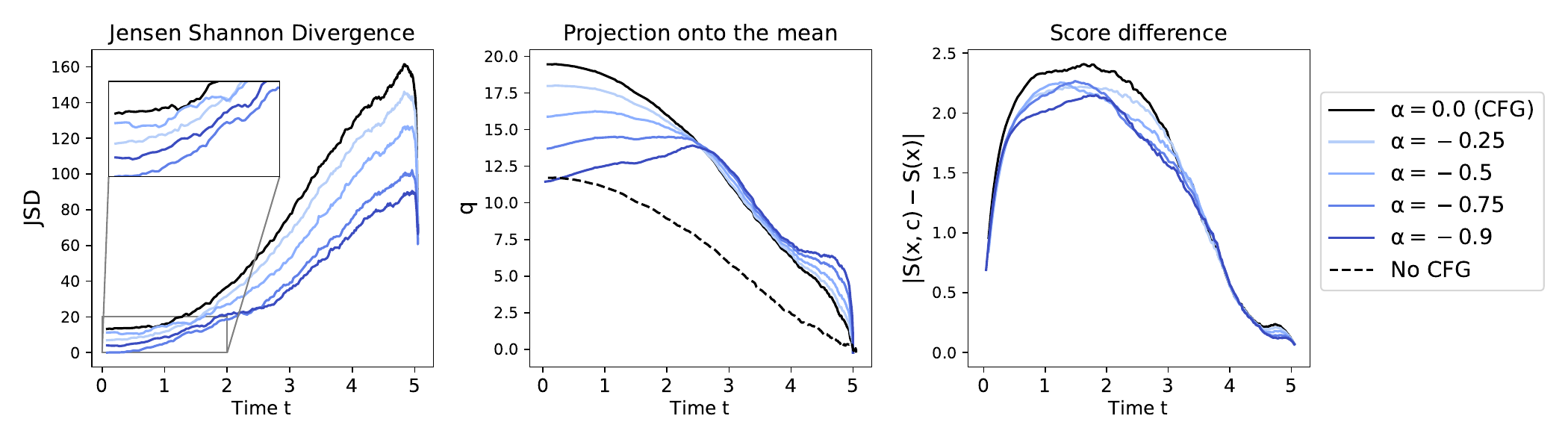}
        % \caption{First version of non-linear CFG.}
        % \label{fig:9_nonlin_v1}
        \end{subfigure}
        
        \begin{subfigure}[b]{\textwidth}
        \centering
        \includegraphics[width=.7\textwidth]{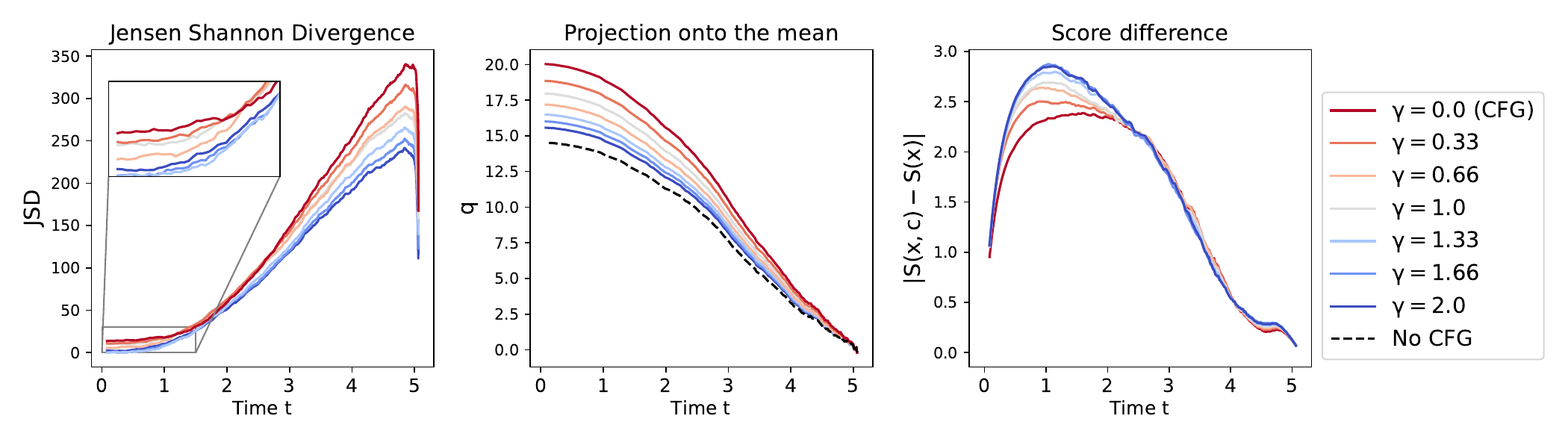}
        % \caption{Second version of non-linear CFG.}
        % \label{fig:10_nonlin_v2}
        \end{subfigure}
        \caption[Caption for LOF]{Real-world experiments using DiT/XL-2 \citep{peebles2023scalable} trained on ImageNet-1000 \citep{deng2009imagenet}: randomly selected class with $\omega=4$, using DDPM \citep{ho2020denoisingdiffusionprobabilisticmodels} with $250$ sampling steps, averaged over $25$ samples. 
        \textbf{First column:} Power-Law CFG. \textbf{Second column:} Rescaled Power-Law CFG (\ref{eqn:nonlin_cfg2}).
        \textbf{Left column:} Jensen-Shannon Divergence between the embedded data points corresponding to randomly selected class and the generated samples as a function of reverse time $\tau$.
        \textbf{Middle column:} mean dot product of the normalized class centroid and the diff. trajectories $\vec{x}\cdot\vec{c_i}/\|\vec{c_i}\|$ (in latent space) as a function of reverse time $\tau$. 
        \textbf{Right column:} Evolution of the distance between cond. and uncond. scores. From all three plots, we can see that using first (second) version of non-linear CFG with $\alpha<0$ ($\gamma>0$) results in paths that have smaller JSD, estimated as in \citet{wang2009divergence}, throughout the whole trajectory and smaller overshoot of the distribution's mean at $\tau=0$. 
        We can also see that the score difference $|S_\tau(x,c)-S_\tau(x)|$ has the same qualitative behavior as in numerical simulations of Gaussian mixtures.
        }
        \labelfiglarge
    \end{figure*}
\else
    \begin{figure*}[ht]
    \centering
    \begin{subfigure}
    \centering
    \includegraphics[width=.8\textwidth]{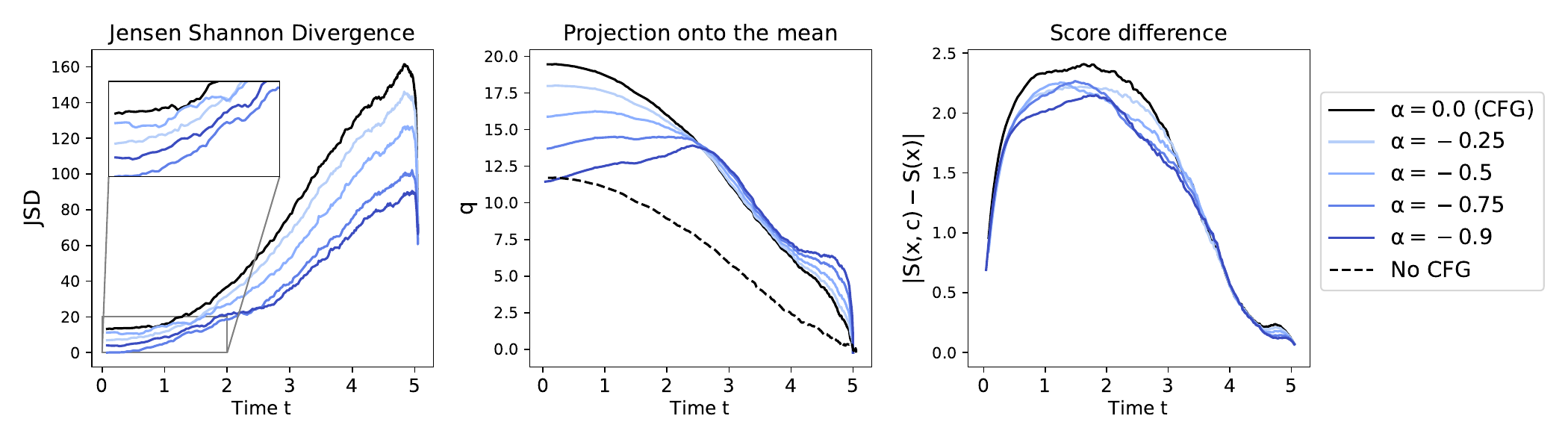}
    % \caption{First version of non-linear CFG.}
    % \label{fig:9_nonlin_v1}
    \end{subfigure}
    
    \begin{subfigure}
    \centering
    \includegraphics[width=.8\textwidth]{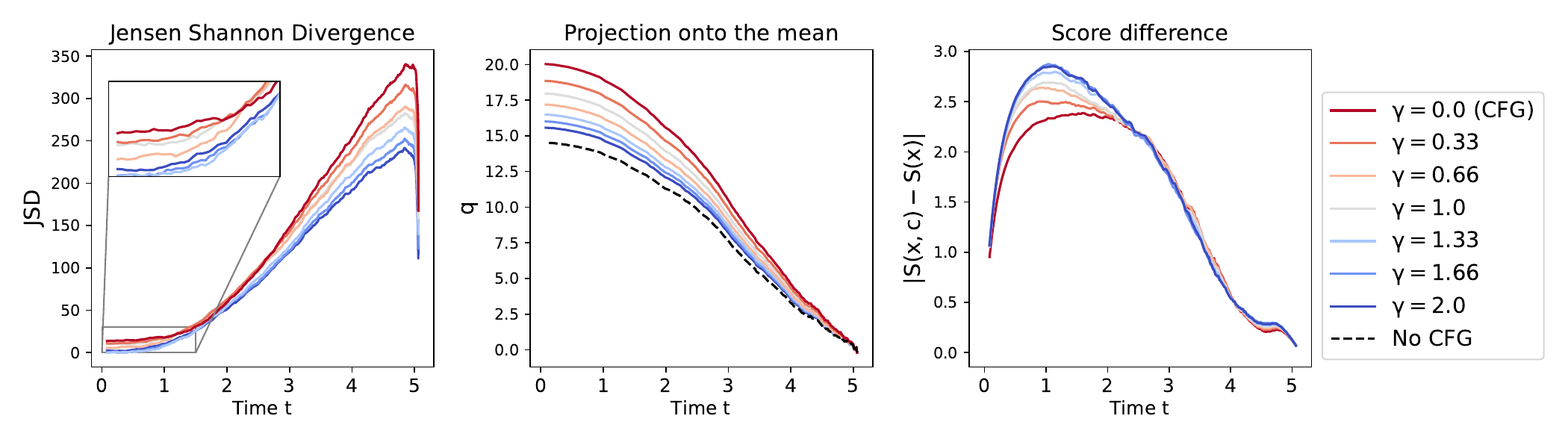}
    % \caption{Second version of non-linear CFG.}
    % \label{fig:10_nonlin_v2}
    \end{subfigure}
    \caption[Caption for LOF]{Real-world experiments using DiT/XL-2 \citep{peebles2023scalable} trained on ImageNet-1000 \citep{deng2009imagenet}: randomly selected class with $\omega=4$, using DDPM \citep{ho2020denoisingdiffusionprobabilisticmodels} with $250$ sampling steps, averaged over $25$ samples. 
    \textbf{First column:} Power-Law CFG. \textbf{Second column:} Rescaled Power-Law CFG (\ref{eqn:nonlin_cfg2}).
    \textbf{Left column:} Jensen-Shannon Divergence between the embedded data points corresponding to randomly selected class and the generated samples as a function of reverse time $\tau$.
    \textbf{Middle column:} mean dot product of the normalized class centroid and the diffusion trajectories $\vec{x}\cdot\vec{c_i}/\|\vec{c_i}\|$ (both in the latent space) as a function of reverse time $\tau$. 
    \textbf{Right column:} Evolution of the distance between conditional and unconditional scores. From all three plots, we can see that using first (second) version of non-linear CFG with $\alpha<0$ ($\gamma>0$) results in paths that have smaller JSD, estimated as in \citet{wang2009divergence}, throughout the whole trajectory and smaller overshoot of the distribution's mean at $\tau=0$. 
    We can also see that the score difference $|S_\tau(x,c)-S_\tau(x)|$ has the same qualitative behavior as in numerical simulations of Gaussian mixtures.
    }
    \labelfiglarge
    \end{figure*}
\fi

\clearpage

\section{Impact Statement}  % does not count to 8 page limit
\label{sec:appx_H_impact}

This study contributes to the growing body of research aimed at deepening our theoretical understanding of diffusion models and their broader implications for generative modeling. By bridging the gap between theory and practice, we strive to improve the performance and efficiency of these models, which have far-reaching applications in various fields.

However, as with any powerful technology, there are also potential risks associated with development and deployment of advanced generative models. The increasing sophistication of deepfakes raises concerns about misinformation, propaganda, and the erosion of trust in digital media. Moreover, the misuse of generative models for malicious purposes, such as creating fake identities or spreading disinformation, poses significant threats to individuals, communities, and society as a whole.

In light of these challenges, we hope that our paper, along with many others that aim to improve understanding of the models, will contribute to a deeper understanding of their strengths and limitations. We believe it is essential for developing effective strategies to mitigate the risks associated with generative models, and we hope that our work will be a step toward achieving this goal.

%%%%%%%%%%%%%%%%%%%%%%%%%%%%%%%%%%%%%%%%%%%%%%%%%%%%%%%%%%%%

\end{document}